\documentclass[twocolumn,switch,times]{article}
\usepackage{preprint}
\usepackage[authoryear,round]{natbib}
\usepackage{hyperref}

\usepackage[table,dvipsnames]{xcolor}
\definecolor{linkblue}{RGB}{48, 118, 219}
\hypersetup{
    colorlinks=true,
    linkcolor=linkblue,
    urlcolor=linkblue,
    citecolor=linkblue,
    anchorcolor=black,
}
\usepackage[utf8]{inputenc}	
\usepackage[T1]{fontenc}
\usepackage{lineno}
\usepackage{tikz}
\usepackage{fancyhdr}
\usepackage{geometry}
\usepackage{lipsum}
\usepackage{scalerel}
\usepackage{sidecap}
\usepackage{times}
\usepackage{doi}
\usepackage{amssymb}
\usepackage{amsmath}
\usepackage{bbm}
\usepackage{bm}
\usepackage{array,booktabs}
\usepackage{upgreek}
\usepackage{algorithm}
\usepackage{algpseudocodex}
\usepackage{appendix}
\usepackage{authblk}
\usepackage{placeins}
\usepackage[nameinlink]{cleveref}

\usepackage{newfloat}
\DeclareFloatingEnvironment[name={Supplementary Figure}]{suppfigure}
\usepackage{sidecap}
\sidecaptionvpos{figure}{c}
\usepackage{etoolbox}
\usepackage{microtype}
\usepackage{tablefootnote}
\usepackage{siunitx}
\usepackage{multirow}
\usepackage{threeparttable}
\usepackage{CJKutf8}

\usepackage{titlesec}

\definecolor{lime}{HTML}{A6CE39}
\definecolor{lightgray}{gray}{0.9} 

\newcommand{\mathbfcal}[1]{\bm{\mathcal{#1}}}
\newcommand{\real}{\mathbbm{R}}
\newcommand{\integer}{\mathbbm{Z}}
\newcolumntype{P}[1]{>{\centering\arraybackslash}p{#1}}
\newcolumntype{R}[1]{>{\raggedleft\arraybackslash}p{#1}}
\newcolumntype{L}[1]{>{\raggedright\arraybackslash}p{#1}}

\crefname{equation}{equation}{equations}
\Crefname{equation}{Equation}{Equations}
\creflabelformat{equation}{#2#1#3}

\newcommand*{\belowrulesepcolor}[1]{%
  \noalign{%
    \kern-\belowrulesep 
    \begingroup 
      \color{#1}%
      \hrule height\belowrulesep 
    \endgroup 
  }%
} 
\newcommand*{\aboverulesepcolor}[1]{%
  \noalign{%
    \begingroup 
      \color{#1}%
      \hrule height\aboverulesep 
    \endgroup 
    \kern-\aboverulesep 
  }%
}
\title{SCATR-ITS: a scalable transformer for large-area individual tree segmentation from LiDAR point clouds}

\usepackage{authblk}

\author[1,2,\small\ensuremath{\ast}]{Lassi Ruoppa}
\author[1]{Verneri Seppänen}
\author[1]{Antero Kukko}
\author[1]{Juha Hyyppä}

\affil[1]{Department of Remote Sensing and Photogrammetry, Finnish Geospatial Research Institute FGI, The National Land Survey of Finland, Vuorimiehentie 5, Espoo, FI-02150, Finland}
\affil[2]{Department of Computer Science, School of Science, Aalto University, P.O. Box 15400, Aalto, FI-00076, Finland}

\begin{document}

\twocolumn[\begin{@twocolumnfalse}

\maketitle

\vspace{-3.0ex}
\noindent\rule{\textwidth}{\heavyrulewidth}\par\vspace{1ex}
\begin{abstract}
Individual tree segmentation (ITS) from LiDAR point clouds is one of the most widely studied segmentation tasks in forestry. Although deep learning methods define the state of the art on the majority of ITS benchmarks, their high computational complexity and long runtime currently prevent their adoption in large-scale operational workflows. In this work, we present SCATR-ITS, a scalable transformer model specifically designed for efficient ITS over large geographical areas. Based on the ForestFormer3D framework, SCATR-ITS transitions to superpoint-level processing to accelerate inference, with the superpoints constructed from learned embeddings. The architecture incorporates size-aware superpoint-level postprocessing, a dynamic query count module to handle high variation in forest structure, and a maximal input stride designed to minimize prediction redundancy. SCATR-ITS achieves accuracy on par with the state of the art on the FGI-EMIT and FOR-InstanceV2 datasets with an up to $8.5\times$ higher inference speed. In addition, SCATR-ITS has a significantly lower GPU VRAM footprint during both training and inference, and exhibits excellent zero-shot generalization across multiple datasets. To demonstrate large-area ITS capabilities, SCATR-ITS is applied to a 142 km$^2$ region of Finnish national laser scanning data where it achieves an average throughput of 13.8 min/km$^2$, $\sim\!18.6\times$ faster than ForestFormer3D and approaching the speed of the unsupervised watershed algorithm, which it vastly outperforms in accuracy and segment quality. Overall, SCATR-ITS delivers state-of-the-art accuracy at speeds where operational-scale adoption becomes viable, paving the way toward countrywide deployment. The source code for SCATR-ITS and pretrained weights are available at \url{https://github.com/ruoppa/scatr_its}.
\end{abstract}

\keywords{LiDAR, Individual tree segmentation, Airborne laser scanning, Deep learning, Forest, Scalable 3D segmentation, Computational efficiency}
\noindent\rule{\textwidth}{\heavyrulewidth}

\vspace{0.5cm}

\end{@twocolumnfalse}]

\renewcommand{\thefootnote}{\small\ensuremath{\ast}}
\footnotetext[1]{\footnotesize Corresponding author: Lassi Ruoppa (lassi.ruoppa@nls.fi)}
\renewcommand{\thefootnote}{\arabic{footnote}}
\setcounter{footnote}{0}

\section{Introduction}

Light detection and ranging (LiDAR) has been widely adopted for remote sensing of forests in the past two decades \citep{hyyppa2008review,kaartinen2012international}. The three-dimensional (3D) point clouds acquired by LiDAR can be utilized in a wide range of downstream tasks, for example, automated forest inventories \citep{xiang2024automated}, estimation of aboveground biomass (AGB) for forest carbon accounting \citep{seely2025forest}, and identification of trees with high biodiversity value \citep{eysn2015benchmark}. Many of these applications, such as tree species classification \citep{hakula2023individual,puliti2025benchmarking,taher2026multispectral} and growth and change monitoring \citep{yu2004automatic, yu2006change,duncanson2018monitoring,tavi2026framework}, benefit from or require partitioning the point cloud into non-overlapping subsets representing individual trees, a process known as individual tree segmentation (ITS) \citep{hyyppa1999detecting}.

Since ITS is an essential preprocessing step for such a large number of applications, it has historically been one of the most widely studied point cloud segmentation tasks in forestry \citep{kaartinen2012international,cao2023benchmarking}. Traditionally, ITS has been performed using unsupervised algorithms that rely on point cloud geometry and heuristics designed to capture the characteristic structure of individual trees \citep{hyyppa1999detecting,koch2006detection,li2012new,ferraz2016lidar,xi20223d}. Due to their computational efficiency, many unsupervised ITS algorithms remain in use to this day, especially in large-scale applications \citep{ma2023individual,hyyppa2024concepts,kostensalo2026large}. Following the successful application of deep learning (DL) to general-purpose 3D instance segmentation \citep{jiang2020pointgroup,vu2022softgroup,sun2023superpoint,kolodiazhnyi2024oneformer3d} and the recent introduction of multiple manually annotated forest point cloud datasets \citep{puliti2023forinstance,xiang2025forestformer3d,li2026itsnet,ruoppa2026benchmarking,sterenczak2026central}, ITS research has increasingly shifted toward 3D DL-based models \citep[see e.g.][]{xiang2024automated,wielgosz2024segmentanytree,henrich2024treelearn,xiang2025forestformer3d,wielgosz2026segmentanytreev2}. At present, DL approaches define the state of the art on several ITS benchmarks and outperform conventional unsupervised approaches by a significant margin across the majority of forest data types.

In operational forestry, area-based approaches \citep[ABAs;][]{naesset2002predicting} that estimate stand-level forest attributes are currently the predominant approach for deriving forest variables from point cloud data \citep{kotivuori2021prediction,imangholiloo2022adding}. Although ABAs are computationally efficient, stand-level attributes lack the spatial detail of ITS, such as tree location and spacing \citep{duncanson2014efficient} and size distribution, which can be advantageous, especially for ecology-oriented applications and autonomous harvesting \citep{perry2026towards}. The scalability of ITS methods to large forested areas covering hundreds or even thousands of km$^2$ is essential for their operational deployment \citep{hamraz2017scalable}. The increasing availability of national laser scanning data \citep{kakoulaki2021noncommercial} further creates opportunities for extending tree-level mapping to nationwide scales. However, research on large-area ITS from LiDAR point clouds remains limited, with existing studies relying exclusively on unsupervised algorithms \citep[see e.g.][]{chubey2009semi,leckie2017production,krzystek2020large,ma2023individual,hyyppa2024concepts}. Moreover, due to their computational efficiency, many of these studies employ 2D-based ITS methods, such as watershed segmentation \citep{yu2011predicting}, which has been successfully applied even at the national scale \citep{hyyppa2024concepts}. Unsupervised 2D algorithms, while fast, are inherently incapable of segmenting understory trees because they operate on $xy$-plane projections of the original 3D point cloud. At the same time, a recent benchmark of ITS methods found that widely used unsupervised 3D algorithms are either less accurate than watershed or their performance deteriorates on sparse point clouds at densities typical of large-scale airborne laser scanning (ALS) inventories \citep{ruoppa2026benchmarking}.

The computational efficiency of 2D ITS methods extends to DL-based approaches, which have proven viable for nationwide tree mapping from aerial imagery \citep{mugabowindekwe2023nation,li2023deep}, are straightforward to adapt to LiDAR data \citep[see e.g.][]{windrim2019forest,sun2022individual,straker2023instance}, and generally achieve slightly better segmentation accuracy than their unsupervised counterparts \citep{ruoppa2026benchmarking}. However, much like unsupervised 2D methods, they are limited to segmenting large overstory trees. When validated against 3D ground truth, individual tree segments derived from 2D data using DL often suffer from limited structural accuracy  \citep{steier2024is,allen2025manual}.

Given the limitations of both unsupervised algorithms and 2D DL approaches, 3D DL-based ITS models seem like the natural choice for large-area segmentation, especially since they have consistently outperformed unsupervised methods by a significant margin \citep{xiang2024automated,xi2025new,ruoppa2026benchmarking}, and retain this performance advantage even at point densities as low as 5 points/m$^2$ \citep{ruoppa2026benchmarking}. However, comparatively long runtimes and high computational requirements currently limit the adoption of 3D DL models in large-scale applications. In fact, several studies have noted that utilizing existing DL-based ITS approaches for segmenting extensive geographical areas is infeasible in practice due to their high computational complexity \citep{itakura2026large,kostensalo2026large}.

At present, research on individual tree segmentation has yet to address the gap between the accuracy offered by 3D DL-based methods and the computational efficiency required for large-area applications. In this work, we propose a scalable and fast transformer-based ITS model optimized for segmenting forest point clouds across broad geographical areas. The model is validated against existing DL-based methods on multiple individual tree segmentation benchmark datasets. In addition, we assess the model's suitability for large-area segmentation using an extensive dataset of Finnish national laser scanning data (20 points/m$^2$), comparing its performance directly to an established unsupervised ITS algorithm previously deployed in large-scale tasks. Our main contributions are as follows:
\begin{enumerate}
    \item We present \textbf{SCATR-ITS}, a \textbf{Sca}lable \textbf{Tr}ansformer for Large-area \textbf{I}ndividual \textbf{T}ree \textbf{S}egmentation. To our knowledge, this is the first 3D DL ITS model specifically designed for segmenting point clouds spanning extensive geographical areas. Evaluation on the FGI-EMIT \citep{ruoppa2026benchmarking} and FOR-InstanceV2 \citep{xiang2025forestformer3d} datasets shows that SCATR-ITS achieves segmentation accuracy comparable to the state of the art while being significantly faster and less resource-intensive, with inference speeds comparable to a 2D-based approach when preprocessing is accounted for. SCATR-ITS also demonstrates strong zero-shot generalization on the LAUTx \citep{tockner2022automatic} and Wytham Woods \citep{calders2022laser} datasets.
    \item Based on the ForestFormer3D framework \citep{xiang2025forestformer3d}, SCATR-ITS introduces several adaptations designed to increase inference speed and reduce resource usage while retaining accuracy, including superpoint-level instance decoding, superpoint-size-aware postprocessing, dynamic query count, and maximal input stride. An ablation study on FGI-EMIT and FOR-InstanceV2 demonstrates that, when combined, these modifications accelerate inference and improve segmentation accuracy while simultaneously reducing GPU VRAM usage.
    \item We demonstrate the large-area inference performance of SCATR-ITS on Finnish national laser scanning data covering 142 km$^2$. The model achieves a runtime approximately $2.5\times$ that of watershed, the unsupervised algorithm typically employed for ITS at such scales, while significantly improving segmentation accuracy. We further estimate that SCATR-ITS is more than $18\times$ faster than ForestFormer3D for this dataset. To the best of our knowledge, this is the first application of DL-based ITS to 3D point clouds over such an extensive area, demonstrating the viability of SCATR-ITS for large-scale operational forest inventories.
\end{enumerate}

\section{Related work}

\subsection{Individual tree segmentation from point clouds}

Individual tree segmentation from point clouds is one of the most extensively studied segmentation tasks in forestry. Early ITS research focused primarily on unsupervised heuristic algorithms that relied on geometric characteristics of trees, occasionally supplemented by radiometric or RGB features \citep{ruoppa2026benchmarking}. In recent years, the focus has shifted to supervised deep-learning-based approaches, which typically outperform conventional unsupervised methods by a significant margin \citep{xiang2024automated,xi2025new,ruoppa2026benchmarking}, although exceptions exist, particularly for dense TLS data \citep{cherlet2026benchmarking}.

Unsupervised ITS algorithms can be broadly divided into two categories according to whether they operate on the data in two or three dimensions. 2D methods generally project the point clouds onto the $xy$-plane to construct a canopy height model (CHM), detect local maxima, and subsequently segment individual trees using, for example, region growing \citep{hyyppa1999detecting,hyyppa2001segmentation,dalponte2016tree}, marker-controlled watershed \citep{koch2006detection,yu2010comparison,yu2011predicting}, or centroidal Voronoi tessellation \citep{silva2016imputation}. Template-matching-based approaches have also shown some success in 2D ITS \citep{pirotti2010assessing,huo2020individual}.

For unsupervised ITS methods operating directly on 3D point clouds, the range of methodologies is significantly broader. Common approaches include clustering \citep{lee2010adaptive,ferraz20123d,ferraz2016lidar}, graph representations \citep{strimbu2015graph,wang2020unsupervised,xi20223d}, and hybrid methods that process the point clouds as several 2D layers \citep{wang2008lidar,duncanson2014efficient,ayrey2017layer}. Additionally, a variety of 3D-based algorithms combine multiple point cloud processing paradigms, making explicit categorization challenging \citep[see e.g.][]{li2012new,mongus2015efficient,hamraz2016robust}. 

Similarly to unsupervised algorithms, research on DL-based point cloud ITS can be categorized into three primary methodological directions: 2D models, hybrid models employing a combination of DL and conventional algorithms, and fully 3D DL models. Much like their unsupervised counterparts, 2D DL methods first transform the point cloud into an image representation by projecting it onto the $xy$-plane. ITS is then usually performed using image segmentation models, which are occasionally specifically adapted for the task. Examples include Faster R-CNN \citep{windrim2019forest,windrim2020detection} and variants of YOLO \citep{chang2022twostage,sun2022individual,straker2023instance}. On the other hand, the pipeline of hybrid methods generally consists of semantic segmentation of the point cloud data using either a 2D or 3D model, followed by a conventional ITS algorithm that takes advantage of the resulting classification. The two most common approaches are binary classification into wood and foliage points \citep{krisanski2021forest,wielgosz2023point2tree,zhu2024synergizing}, and tree and non-tree points \citep{chen2021individual,yang2024improved}.

The remaining methodological direction, fully 3D DL models, generally produces the highest segmentation accuracy. However, the methods are typically also the most computationally demanding \citep{ruoppa2026benchmarking}. Early works were based on the general-purpose 3D segmentation model PointGroup \citep{jiang2020pointgroup}, which employs a combination of learned center offsets and clustering. \citet{xiang2023towards} adapted PointGroup for ITS by introducing an additional feature embedding branch trained with a contrastive loss to improve the separation of closely grouped instances. The framework was later utilized by both the forest panoptic segmentation model ForAINet \citep{xiang2024automated} and the sensor-agnostic SegmentAnyTree \citep{wielgosz2024segmentanytree}. Several other studies have since proposed similar deep-clustering-based frameworks for ITS, with modifications to, for example, backbone feature extractors and clustering algorithms. Many have also introduced additional adaptations specifically designed for forest data, including 2D offset \citep{henrich2024treelearn}, stem extraction \citep{xi2025new}, and heatmap \citep{du2026satree} branches, as well as positional embedding \citep{li2026itsnet}.

The current overall state of the art in ITS is set by 3D mask transformer models. First adapted for forest data by \citet{xiang2025forestformer3d}, who introduced ForestFormer3D, mask transformers directly predict a binary mask across the entire input for each tree instance in an end-to-end fashion. The same general framework was also used in SegmentAnyTreeV2 \citep{wielgosz2026segmentanytreev2} and ForestMamba \citep{nguyen2026forestmamba}. The former shifted to a transformer-based backbone feature extractor, while the latter opted for a Mamba-based network for both the backbone and instance decoder. Outside of research focused strictly on improving the state of the art, some recent studies have explored alternative methodological directions for DL-based ITS. Examples include user-promptable ITS methods \citep{xiu2025individual,nguyen2026selectanytree} and a weak-supervision-based rating module for refining the outputs of ITS models \citep{destouches2026weakly}.

Despite their accuracy, DL-based ITS methods have two primary limitations. First, nearly all existing approaches are fully supervised and therefore require large amounts of manually annotated 3D forest point cloud data for training. Second, DL models typically require more computational resources and have a longer runtimes \citep{ruoppa2026benchmarking}, which can make them infeasible for large-area ITS \citep{itakura2026large,kostensalo2026large}. Regarding the former, the lack of public large-scale datasets was previously considered a major factor limiting advances in forest point cloud segmentation research \citep{lines2022ai,hamedianfar2022deep}. However, the recent introduction of datasets such as LAUTx \citep{tockner2022automatic,tockner2022lautx}, Wytham Woods \citep{calders2022terrestrial,calders2022laser}, FOR-Instance \citep{puliti2023forinstance,puliti2023forinstancedata}, FOR-InstanceV2 \citep{xiang2025forestformer3d,xiang2025forinstancev2}, FGI-EMIT \citep{ruoppa2026benchmarking,ruoppa2026fgiemit}, AerialTrees \citep{li2025aerialtrees,li2026itsnet}, TLS benchmark \citep{cherlet2025tls,cherlet2026benchmarking}, and TreeScanPL10K \citep{sterenczak2026central,sterenczak2026treescanpl10k} has largely alleviated this problem, particularly for European forest environments. Moreover, some recent studies have introduced large amounts of synthetic training data \citep{she2026scaling,liu2026synthetic,jiang2026cross} and self-supervised pretraining \citep{rizaldy2026label,yue2026litept} to further mitigate the issue. By contrast, reducing the computational complexity of accurate DL-based ITS models remains largely unexplored.

\subsection{Advantages of ITS over area-based approaches}

Alongside ITS, area-based approaches that predict stand-level attributes, such as average volume and basal area, represent another primary approach for deriving forest variables from LiDAR data \citep{vastaranta2012combination}. ABAs divide the point cloud into grid cells (e.g. $16\times16$ m) for which stand-level attributes are predicted using statistical or machine learning regression models based on LiDAR-derived features. These models are typically fitted to ground truth data from field inventories \citep{packalen2019resolution}. 

At present, ABAs are the predominant approach for processing LiDAR point clouds over large geographical areas in operational forestry \citep{vastaranta2012combination,packalen2019resolution,kotivuori2021prediction,imangholiloo2022adding}. In fact, several studies comparing ABAs with conventional unsupervised ITS algorithms have found that stand-level attributes derived from automatically segmented individual trees are comparable to those obtained using ABAs \citep{vastaranta2009comparison,yu2010comparison,peuhkurinen2011comparing}, with the two approaches often yielding highly correlated results \citep{latifi2015forest}. Consequently, when only stand-level attributes are of interest, ABAs are generally preferred because of their substantially lower computational requirements.

The low detection rate of understory trees has been cited as a primary limitation of deriving stand-level attributes from ITS results \citep{latifi2015forest,dalponte2016tree,jeronimo2018applying}. Given that previous performance comparisons have focused on conventional unsupervised algorithms, DL-based ITS has the potential to mitigate this limitation: modern DL models, such as ForestFormer3D, are significantly more effective at detecting understory trees \citep{ruoppa2026benchmarking}. However, whether stand-level attributes derived from modern DL-based ITS can outperform those obtained using ABAs remains unclear, since, to the best of our knowledge, no studies to date have addressed this question.
    
Regardless of whether ITS-based methods can outperform ABAs in stand-level attribute prediction, large-scale tree-level information is required for several applications. Stand-level attributes lack the spatial detail provided by individual trees, including their location, spacing, and size distribution, which can provide a more ecologically meaningful description of forest structure \citep{duncanson2014efficient}. A recent review of tree-level management planning and decision-making in forestry identified several advantages over stand-level strategies \citep{perry2026towards}. For example, tree-level information enables the consideration of factors such as dead trees, which are important for local biodiversity, and stem defects, which affect economic value. Furthermore, tree-level information also has potential applications in autonomous harvesting \citep{perry2026towards}.

While many downstream tasks are compatible with both the stand-level and tree-level paradigms, processing data as individual tree segments is often advantageous, particularly for ecology-oriented applications. \citet{khatrichhetri2024enhancing} proposed that mapping and studying tree mortality at the tree-level improves both ecological understanding and the capacity to mitigate loss. Similarly, \citet{dalponte2016tree} argued that individual tree segments provide a more appropriate representation for forest carbon stock mapping than ABAs, since the they more closely resemble manual forest inventories, enable the use of species-specific allometry, and extend applicability to different spatial scales. ITS also enables tree-level species classification and age prediction. The former can be particularly useful for certain rare keystone tree species, such as the European aspen (\emph{Populus tremula} L.) \citep{toivonen2024mapping,taher2026multispectral}, while the latter is important for understanding forest dynamics, tree growth, and site productivity \citep{puliti2026forage}.

Finally, there are multiple applications for which individual tree segments are a strict requirement. LiDAR-derived individual tree segments can be used for computing competition indices that improve the accuracy of growth models \citep{pedersen2012deriving,ma2018quantifying}. Furthermore, if the data are multi-temporal, ITS enables growth monitoring and detection of harvested trees \citep{yu2004automatic,duncanson2018monitoring,tavi2026framework}. In addition, individual tree maps derived from LiDAR point clouds can be used for developing optimal thinning plans \citep{lee2025development,platt2026precision}, supporting harvester operator decisions, and potentially real-time harvester positioning when combined with onboard MLS \citep{seppala2026lidar}.

\subsection{Large-area individual tree segmentation}

Scalability to large geographic extents is a prerequisite for the operational deployment of individual tree segmentation in forest inventories (here defined as $\gtrsim100$ km$^2$) \citep{hamraz2017scalable}. However, research on applying ITS at large scales remains relatively scarce, particularly for LiDAR point cloud data, where area-based approaches remain the predominant operational paradigm \citep{vastaranta2012combination,packalen2019resolution,kotivuori2021prediction,imangholiloo2022adding}. By contrast, large-area ITS from remote sensing (RS) imagery is considerably more developed, primarily due to the lower computational cost of processing 2D data.

Earlier studies on large-area ITS from imagery typically relied on conventional unsupervised approaches. The valley-following algorithm \citet{gougeon1995crown}, for example, has been used in several works. \citet{chubey2009semi} used the method to assess the viability of ITS for large-scale forest inventories, while \citet{frescino2007investigation} and \citet{leckie2017production} applied it to multispectral satellite and aerial imagery, respectively. Other conventional approaches have also been used. For example, \citet{karlson2014tree} combined spectral attributes and region growing to segment a 100 km$^2$ area of WorldView-2 satellite imagery.

Transferring conventional ITS methods to extensive geographical areas can be challenging due to their sensitivity to image quality \citep{freudenberg2022individual}. Consequently, DL-based methods have emerged as a promising approach for large-area ITS from RS imagery. For example, \citet{weinstein2021remote} presented a dataset consisting of 100 million trees segmented from aerial imagery using DeepForest \citep{weinstein2020deepforest}, while \citet{brandt2020unexpectedly} applied DL for ITS from satellite imagery covering 1.3 million km$^2$. Country-level ITS from aerial imagery has also been demonstrated in Rwanda \citep{mugabowindekwe2023nation} and Denmark \citep{li2023deep}.

To the best of our knowledge, no previous studies have demonstrated DL-based segmentation directly from 3D point clouds over large geographic areas. Existing methods have predominantly been evaluated at plot or site level, typically covering only a few hectares. By contrast, several studies have explored conventional unsupervised methods for large-area ITS from point clouds, with computationally efficient 2D CHM-based approaches being particularly common. \citet{ma2023individual} and \citet{itakura2026large} both applied standard watershed segmentation to broad geographic extents, with the latter using a high performance computing system. Similarly, \citet{kostensalo2026large} employed the CHM-based method of \citet{silva2016imputation} with no modifications. Notably, \citet{kostensalo2026large} proposed using a partial least squares model to predict missed and incorrect detections and improve stand-level statistics derived from the segmentation results. Large-area-optimized 2D approaches have also been proposed, with \citet{hyyppa2024concepts} employing a parallelized version of the watershed algorithm.

Unsupervised 3D ITS methods have also been applied to large-area segmentation. \citet{qin2014individual} first divided LiDAR data into small subsections by automatically delineating forest stands from aerial imagery, and then performed ITS using AMS3D \citep{ferraz20123d}. \citet{hamraz2017scalable} proposed a distributed pipeline for processing large LiDAR point clouds in parallel, including the merging of tree segments near plot boundaries. Their approach used a hybrid ITS algorithm based on sequentially segmenting the largest tree using global maxima and vertical profiles \citep{hamraz2016robust}. Finally, \citet{krzystek2020large} employed the graph-based algorithm of \citet{reitberger20093d}, with additional postprocessing steps to improve broadleaf tree segments and parallelization to increase processing speed at large scales.

\begin{table*}[!t]
    \centering
    \caption{Overview of the publicly available LiDAR-based individual tree segmentation datasets with point-level annotations used in this study.} \bigskip
    \begin{threeparttable}
    \scriptsize{
    \begin{tabular}{
        L{2.0cm}
        *{1}{S[
            table-format=5.0,
            group-separator={,},
            group-minimum-digits=4
        ]}
        P{1cm}
        L{3.1cm}
        L{3.1cm}
        L{3.1cm}
        P{0.8cm}
    }
        \toprule
        \textbf{Name} & \multicolumn{1}{P{1cm}}{\textbf{Number of trees}} & \textbf{Scanner type} & \textbf{Forest type} & \textbf{Dataset} & \textbf{Associated publications(s)} & \textbf{Test only} \\ \midrule \midrule
        FGI-EMIT & 1561 & ALS & Coniferous-dominated, deciduous-dominated and mixed boreal forest, planted trees & \citet{ruoppa2026fgiemit} & \citet{ruoppa2026benchmarking} & \\ \midrule
        FOR-InstanceV2 & 11035 & ULS, MLS, TLS & Coniferous-dominated boreal and temperate forest, deciduous-dominated boreal and temperate forest, subtropical forest, tropical forest & \citet{xiang2025forinstancev2} & \citet{xiang2025forestformer3d} & \\ \midrule
        LAUTx\tnote{$\dagger$} & 514 & PLS & Mixed temperate forest & \citet{henrich2024extended}  & \citet{tockner2022automatic,tockner2022lautx} & $\checkmark$ \\ \midrule
        Wytham Woods\tnote{$\dagger$} & 877 & TLS & Deciduous-dominated temperate forest & \citet{henrich2024extended} & \citet{calders2022terrestrial,calders2022laser} & $\checkmark$ \\
        \bottomrule
    \end{tabular}}
    \scriptsize{
    \begin{tablenotes}
      \item[$\dagger$] Extended version of \citet{henrich2024extended}. Number of tree instances differs from the original dataset.
    \end{tablenotes}}
    \end{threeparttable}
    \label{table:dataset_overview}
\end{table*}

At present, several countries around the world collect and publish full or partial countrywide airborne laser scanning data, with point densities ranging from 0.5 to 50 points/m$^2$ \citep{kakoulaki2021noncommercial}. National laser scanning campaigns are particularly common in Europe, where a recent report found that the majority of countries have at least partial LiDAR coverage. The report further identified 18 European countries with nationwide coverage, of which at least 13 have made the corresponding point cloud data fully or partially  available to the public: Denmark, Estonia, Finland, France, Latvia, Luxembourg, Malta, Netherlands, Norway, Poland, Slovenia, Spain, and Switzerland \citep{kakoulaki2021noncommercial}. Outside Europe, examples of countries with ongoing national laser scanning programs include New Zealand \citep{linz2026elevation}, the United States of America \citep{usgs2024lidar}, and Japan \citep{gsi2026pointcloud}. The latter two also provide the corresponding point cloud data publicly.

The growing availability of nationwide point cloud data opens up the possibility of scaling LiDAR-based large-area individual tree segmentation to the level of entire countries. In fact, recent studies have demonstrated that nationwide automatic forest inventory at the individual tree level is already feasible using unsupervised 2D approaches. \citet{schaller2023improving} proposed a CHM-based tree detection method that applies a hybrid filter with parameters determined by a regression model. They demonstrated the efficiency of their method by performing tree detection across the entire country of Switzerland using the national laser scanning data. \citet{hyyppa2024concepts} presented a software framework for nationwide individual tree-level forest data using Finnish national laser scanning data \citep[5 points/m$^2$,][]{nls2026laser}. ITS was performed using a parallelized watershed algorithm on 1 km $\times$ 1 km tiles covering a total area of 3896 km$^2$, with an average runtime of 1.5 min per tile. The proposed framework was later used to create the Metsäkanta website\footnote{\url{https://metsakanta.fi/en}}, which contains records of more than 8.5 billion individual trees segmented from Finnish national laser scanning data, with coverage currently spanning most of Finland.

\section{Materials}

This section provides an overview of the publicly available individual tree segmentation point cloud datasets used for training and evaluating the SCATR-ITS model, followed by a description of the data used to demonstrate its large-area ITS capabilities.

\subsection{ITS datasets for model training and performance evaluation} \label{section:its_datasets_training}

We used two established and publicly available individual tree segmentation benchmark datasets, FGI-EMIT \citep{ruoppa2026benchmarking,ruoppa2026fgiemit} and FOR-InstanceV2 \citep{xiang2025forestformer3d,xiang2025forinstancev2} for model training and performance evaluation. Both datasets consist of several forest-scene point clouds with manually generated point-level annotations for all tree instances. Furthermore, both have been used in multiple previous studies, enabling direct comparison with existing results.

FOR-InstanceV2 can be considered the current gold standard among point cloud ITS datasets. It is one of the largest publicly available datasets in terms of the number of annotated trees (11,035) and is highly diverse, with data from multiple continents and scanner types \citep{xiang2025forestformer3d}. By contrast, FGI-EMIT contains significantly fewer trees (1,561), but its focus on small understory trees makes it particularly useful for evaluating segmentation performance in multi-layered forest environments. Furthermore, despite being limited to a single geographic area, FGI-EMIT exhibits considerable diversity in species composition and forest structure and notably also includes built environments. Although FGI-EMIT also contains multispectral (MS) reflectance information, we only utilize the $xyz$-coordinates. The dataset characteristics, including forest types and sensor modalities, are summarized in \Cref{table:dataset_overview}. Both datasets are ML-ready and provide predefined training, validation, and test splits, which we followed in all experiments.

Additional preprocessing steps were performed for FOR-InstanceV2 to standardize the format of the training data. First, to reduce processing times and balance the number of TLS-based forest scenes, the large BlueCat region plot in the training set was divided into six plots of approximately equal size. Instances located on the plot boundaries were assigned to the plot containing the majority of their points. Second, to enable assessment of segmentation accuracy at different forest levels, crown categories equivalent to those in FGI-EMIT (see \Cref{table:crown_categories}) were assigned to each instance following the procedure presented in \citet{ruoppa2026benchmarking}. A Python script for reproducing these preprocessing steps is provided in the SCATR-ITS GitHub repository.

\begin{table}[!b]
    \centering
    \caption{Specifications of the four tree crown categories. Two trees were considered neighbors if the distance between them was less than 3 m. Table reproduced from \citet{ruoppa2026benchmarking}.}
    \bigskip
    \scriptsize{
    \begin{tabular}{cp{2.2cm}p{4.5cm}}
        \toprule
        \textbf{Category} & \textbf{Description} & \textbf{Definition} \\
        \midrule \midrule
        A & Isolated or dominant trees & Tree has no neighboring trees or is $\geq2$ m higher than all neighbors. \\ \midrule
        B & Group of similar trees & Tree has at least one neighboring tree and is $<2$ m higher than all neighbors. \\ \midrule
        C & Tree alongside a dominant tree & Tree has at least one neighboring tree which is $\geq2$ m higher and at a distance of $\geq1.5$ m. \\ \midrule
        D & Tree under a dominant tree & Tree has at least one neighboring tree which is $\geq2$ m higher and at a distance of $<1.5$ m.\\ \bottomrule
    \end{tabular}}
    \label{table:crown_categories}
\end{table}

\subsection{ITS datasets for assessing model generalization} \label{section:its_datasets_generalization}

To assess ITS model generalization, the trained models were evaluated on two previously unseen forest datasets. Following \citet{wielgosz2024segmentanytree,xiang2025forestformer3d}, we opted for LAUTx \citep{tockner2022automatic,tockner2022lautx} and Wytham Woods \citep{calders2022terrestrial,calders2022laser}. Both differ from the training data in terms of scanner modality and forest type, rendering them particularly useful for evaluating model generalization. The relevant dataset characteristics are listed in \Cref{table:dataset_overview}.

\begin{figure*}[!t]
\centering
\includegraphics[width=\textwidth]{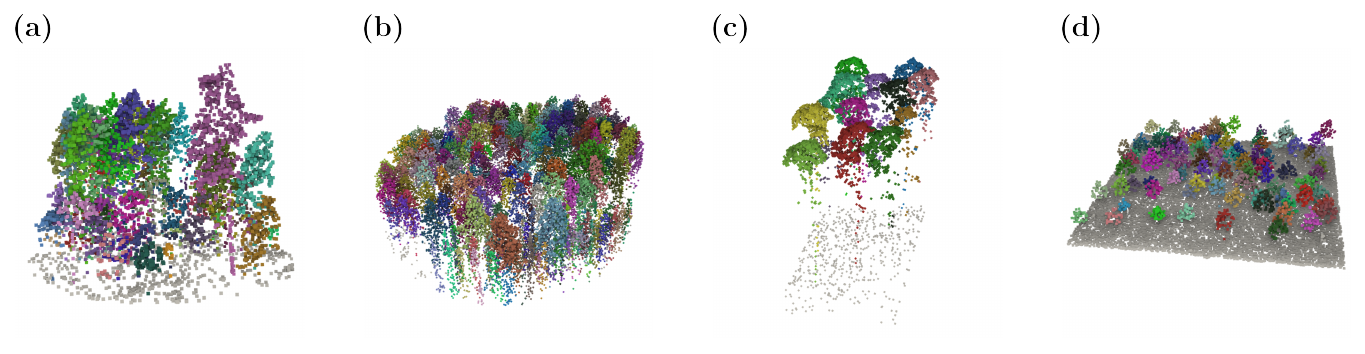}
\caption{Examples of sparsified (20 points/m$^2$) ground-truth-annotated point clouds from the ALS-XL-20-pts dataset. Each tree instance is shown in a distinct color, while non-tree points are shown in gray. (a) Plot 45 from the NIBIO2 region of FOR-InstanceV2. (b) Plot 2 from EvoMS. (c) Plot 4 from the Gaoefeng region in AerialTrees. (d) Plot 15 from the Mulan region in AerialTrees.}
\label{figure:als_xl_20_pts_examples}
\end{figure*}

Since the original  LAUTx and Wytham Woods datasets consist exclusively of individual tree segments, and omit components such as ground and understory vegetation, we instead employed the versions introduced by \citet{henrich2024extended}, which extend the annotations to cover complete forest scenes.

\subsection{Datasets for large-area ITS} \label{section:its_datasets_large_area}

The primary purpose of the proposed SCATR-ITS model is to provide a more efficient alternative for individual tree segmentation, while achieving performance comparable to that of more computationally demanding state-of-the-art DL-based approaches. The overarching goal is to enable accurate ITS over large-scale datasets spanning hundreds or even hundreds of thousands of square kilometers within a reasonable time frame. This section presents the Finnish national laser scanning data (20 points/m$^2$) used to demonstrate the capabilities of SCATR-ITS for large-area ITS, as well as the ALS-XL-20-pts composite dataset used to train the model.

\subsubsection{ALS-XL-20-pts}

Laser scanning campaigns covering extensive geographical areas are typically conducted from aircraft flown at relatively high altitudes to achieve efficient coverage. Consequently, the resulting data generally have lower point densities than most publicly available annotated forest datasets. For example, the newest ongoing Finnish national laser scanning program has a target density of 20 points/m$^2$ \citep{nls2026laserkeilaus}. To the best of our knowledge, no publicly available manually annotated datasets for individual tree segmentation exist at such low point densities. To this end, we constructed the ALS-XL-20-pts composite dataset, intended for training ITS models for large-area segmentation of sparse ALS data. The dataset is a combination of several publicly available ALS and ULS forest datasets with manually generated instance labels, which have been sparsified to an approximate density of 20 points/m$^2$ through random sampling. It should be noted that the acquisition geometry of ALS data depends on factors such as flight altitude and scanning angle \citep{keranen2016effect}, which are not accounted for by simple random downsampling. Consequently, the downsampled high-density ALS data provide only an approximation of actual low-density data. Nevertheless, although this approximation is not exact, the resulting geometry was experimentally determined to be sufficiently similar to support training ITS models for application to sparse ALS data.

The specific datasets used to construct ALS-XL-20-pts are FGI-EMIT \citep{ruoppa2026benchmarking,ruoppa2026fgiemit}, EvoMS \citep{ruoppa2025evoms,ruoppa2025unsupervised}, all ALS and ULS plots from FOR-InstanceV2 (CULS, NIBIO, RMIT, SCION, TUWIEN, and Yuchen) \citep{xiang2025forestformer3d,xiang2025forinstancev2}, and the publicly available regions of AerialTrees (Gaofeng and Mulan) \citep{li2025aerialtrees,li2026itsnet}. For FGI-EMIT and FOR-InstanceV2 data, ALS-XL-20-pts follows the official training, validation and test splits. No official data splits were available for the remaining datasets. Consequently, due to its small size, EvoMS was assigned entirely to the training set. By contrast, the AerialTrees plots were randomly assigned to training, validation, and test sets using a 70\%/10\%/20\% ratio, respectively.

Sparsification to an approximate density of 20 points/m$^2$ was achieved by computing the area of each plot and randomly subsampling the number of points required to obtain the target density. Following \citet{ruoppa2026benchmarking}, all ground truth tree instances containing fewer than five points after sparsification were reclassified as background, as reliable detection of such small instances is generally infeasible. The final ALS-XL-20-pts dataset contains 7,815 tree instances across 164 forest plots. Examples of ground-truth-annotated point clouds are shown in \Cref{figure:als_xl_20_pts_examples}. Further details, including the exact data splits, can be found in \Cref{appendix:als_xl_20pts}.

\subsubsection{Finnish national laser scanning data}

\begin{figure*}[!t]
\centering
\includegraphics[width=\textwidth]{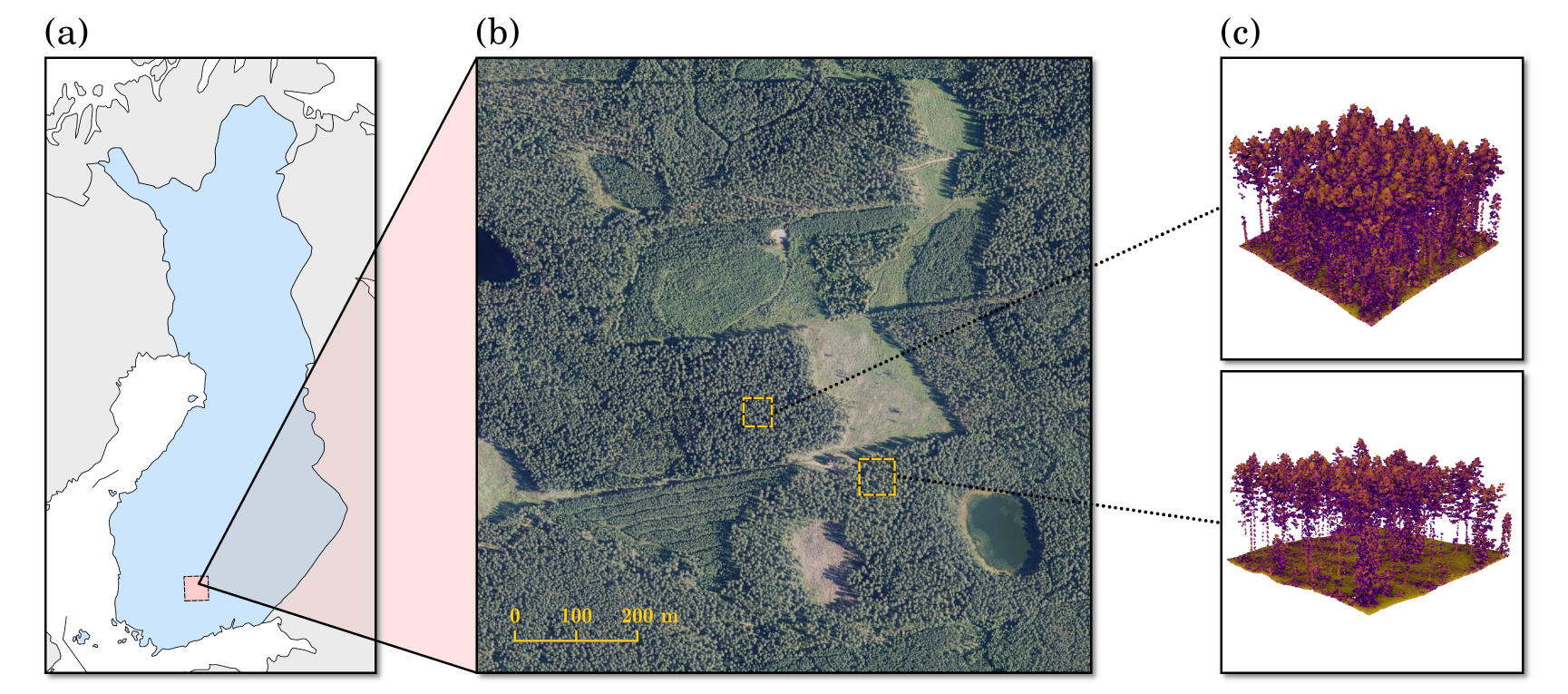}
\caption{Overview of the study area for the Finnish national laser scanning data (20 points/m$^2$), located in Padasjoki, Finland. (a) Map of Finland with the Padasjoki production area highlighted in light red. (b) Example orthophoto (0.5 m pixel size, obtained from \citet{nls2025orthophotos}) corresponding to a single 1 km $\times$ 1 km tile with LiDAR coverage. The boundaries of two test sites with reference tree measurements located within the tile are highlighted in yellow. (c) Subsections of the LiDAR point cloud corresponding to the two test sites highlighted in (b). Colors represent intensity values, with higher values shown as brighter colors.}
\label{figure:padasjoki_20_pts_overview}
\end{figure*}

The first nationwide laser scanning campaign in Finland was conducted from 2008 to 2019, with a target density of 0.5 points/m$^2$. Starting in 2019, the KALLIO program aims to acquire LiDAR data covering the entire country on a nine-year cycle. Under the program, Finland has been divided into several production areas, with approximately 37,000 km$^2$ covered annually \citep{nls2026laserkeilaus}. The first iteration of KALLIO, which ran from 2019 to 2025, acquired LiDAR at an approximate density of 5 points/m$^2$, and the resulting data have been made publicly available \citep{nls2026laser}. A new iteration of the program began in 2026, with an increased target density of 20 points/m$^2$ for most of the country \citep{nls2026laserkeilaus}.

\begin{table}[!b]
    \centering
    \caption{Technical specifications of the airborne laser scanning system used in the Padasjoki production area. AGL denotes above ground level.} \bigskip
    \begin{threeparttable}
    \scriptsize{
    \begin{tabular}{ll}
        \toprule
        \textbf{Parameter} & \textbf{Value} \\ \midrule \midrule
        Scanner model & RIEGL\tnote{$\dagger$}\; VQ-1560II-S \\
        Wavelength (nm) & 1,064 \\
        Field of view ($^\circ$) & 40 \\
        Pulse repetition rate (kHz) & 4,000 \\
        Scan rate (Hz) & 540 \\
        Approximate point density (points/m$^\mathbf{2}$) & 23 \\
        Approximate flight altitude AGL (m) & 1,300 \\
        Approximate ground speed (km/h) & 150 \\
        \bottomrule
    \end{tabular}}
    \scriptsize{
    \begin{tablenotes}
      \item[$\dagger$] RIEGL Laser Measurement Systems GmbH, Austria
    \end{tablenotes}}
    \end{threeparttable}
    \label{table:specifications_20_pts}
\end{table}

To demonstrate the capabilities of SCATR-ITS for large-area ITS and assess its potential for processing country-level LiDAR point clouds, we tested the model on the new Finnish national laser scanning data (20 points/m$^2$). Specifically, we employed data collected during the summer 2025 test flights conducted in the Padasjoki production area in Southern Finland. Centered at approximately 61.2772$^\circ$N , 24.9282$^\circ$E (WGS84, EPSG:4326), the Padasjoki production area covers 3183 km$^2$, of which $\sim\!85\%$ was surveyed during the test flights. The geometry of the point clouds acquired during the ongoing KALLIO program will be equivalent to that of the 2025 test flight data. An overview of the study area is shown in \Cref{figure:padasjoki_20_pts_overview}, and the technical specifications of the ALS system used to collect the data are provided in \Cref{table:specifications_20_pts}.

The point cloud data are provided in 1 km $\times$ 1 km tiles, which we segmented in full. A statistical outlier filter was applied to each tile prior to segmentation, with the number of neighbors and standard deviation threshold ratio set to 3 and 10, respectively.

While point-level ground truth labels similar to those found in, e.g., FOR-InstanceV2 and FGI-EMIT are not available for the national laser scanning data, field-measured reference data collected by the Finnish Forest Centre between 2024 and 2025 are available for a subset of the data. Specifically, several smaller test sites of varying sizes (see \autoref{figure:padasjoki_20_pts_overview} (b)--(c)) are located within the Padasjoki production area, 254 of which fall within LiDAR data coverage. The field measurements contain manually recorded locations and heights for all trees within the test sites and can therefore be used to assess ITS performance. The reference data contain a total of 22,694 living trees, each of which was assigned a crown category according to the definitions in \Cref{table:crown_categories}. In this work, we limited the segmentation tests to complete 1 km$^2$ tiles within the Padasjoki production area that contained one or more test sites with reference data. A total of 142 tiles met this criterion.

\section{Methods}

\begin{figure*}[!t]
\centering
\includegraphics[width=\textwidth]{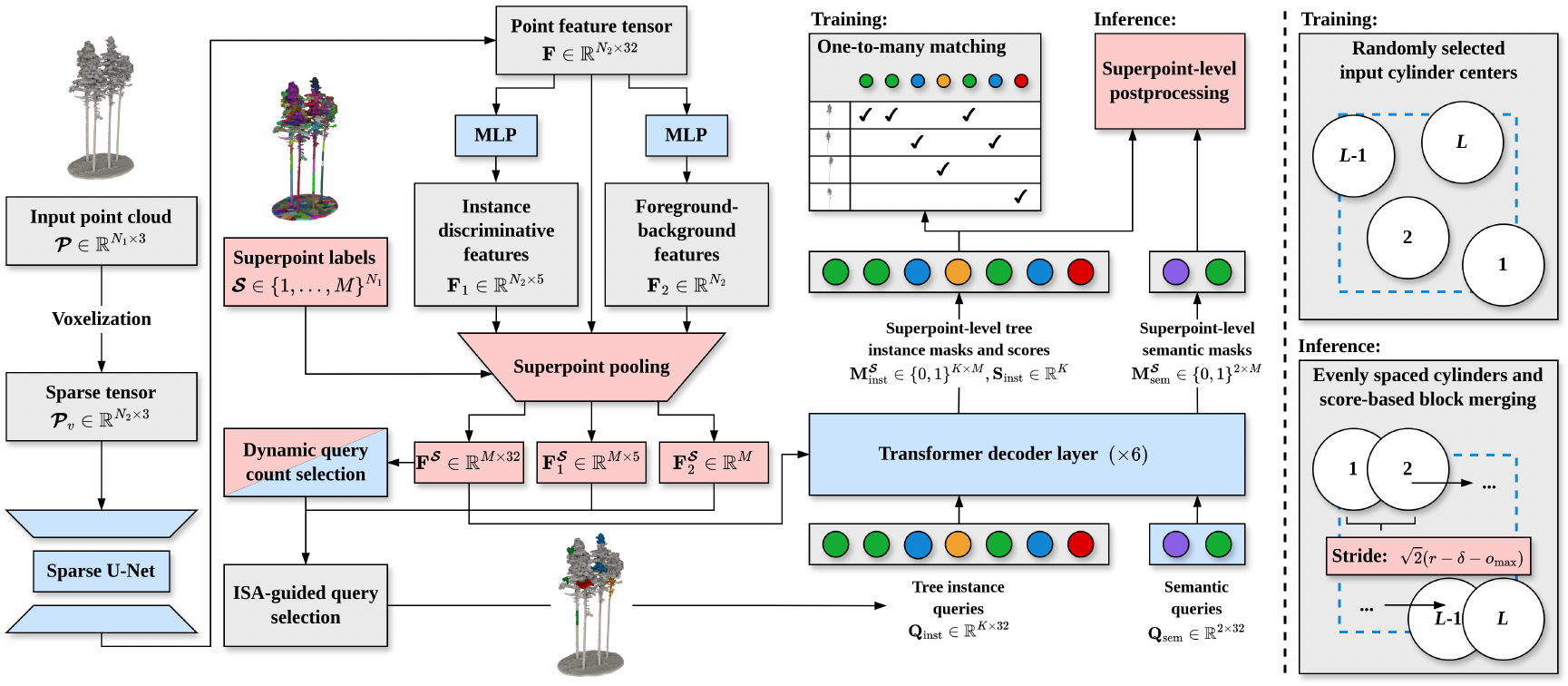}
\caption{Overview of the SCATR-ITS framework for large-area individual tree segmentation from LiDAR point clouds. Key modifications to the underlying ForestFormer3D model are highlighted in light red, while components with learnable weights are shown in light blue.}
\label{figure:scatr_its_architecture}
\end{figure*}

In this section, we provide a detailed description of SCATR-ITS, our transformer-based model for large-area individual tree segmentation from LiDAR point clouds. Based on the ForestFormer3D \citep{xiang2025forestformer3d} framework, SCATR-ITS incorporates several modifications specifically designed to increase inference speed while simultaneously reducing usage of computational resources and retaining segmentation accuracy. We first provide a brief overview of the full SCATR-ITS pipeline, followed by a more detailed discussion of each submodule. An overview of the SCATR-ITS model architecture, with the key modifications highlighted, is shown in \Cref{figure:scatr_its_architecture}.

Given a forest point cloud $\mathbfcal{P}\in\real^{N_1\times 3}$ with $N_1\in\integer_+$ points, SCATR-ITS first assigns each point a superpoint label from $\mathbfcal{S}\in\{1,\ldots,M\}^{N_1}$  using a DL-based superpoint generator (\mbox{\Cref{section:superpoint_generator}}). Next, following the ForestFormer3D pipeline, SCATR-ITS voxelizes the point cloud and extracts 32D point-level features $\mathbf{F}\in\real^{N_2\times 32}$ using a sparse U-Net backbone (\Cref{section:scatr_its_backbone_network}), where $N_2\leq N_1$ is the number of active voxels. These features are subsequently fed into two multilayer perceptron (MLP) branches, which transform them into 5D instance-discriminative features $\mathbf{F}_1\in\real^{N_2\times 5}$ and foreground-background logits $\mathbf{F}_2\in\real^{N_2}$.

The extracted feature tensors and generated superpoint labels are passed to a superpoint pooling module (\Cref{section:instance_decoder}), which obtains superpoint-level feature representations $\mathbf{F}^{\mathbfcal{S}}\in\real^{M\times 32}$, $\mathbf{F}^{\mathbfcal{S}}_1\in\real^{M\times 5}$, and $\mathbf{F}^{\mathbfcal{S}}_2\in\real^{M}$ by computing the superpoint-wise mean values. A dynamic query count module then predicts the scene density and selects the number of queries $K$ accordingly. Queries are obtained using the ISA-guided strategy of \citet{xiang2025forestformer3d} at the superpoint level: predicted foreground superpoints are first identified, after which query positions are selected by applying farthest point sampling (FPS) to the corresponding instance discriminative features. The query values $\mathbf{Q}_{\text{inst}}\in\real^{K\times 32}$ are then set to those of the 32D features at the selected positions and fed into an instance decoder consisting of six transformer layers. The decoder outputs binary masks $\mathbf{M}^{\mathbfcal{S}}_{\text{inst}}\in\{0,1\}^{K\times M}$ and scores $\mathbf{S}_{\text{inst}}\in\real^K$ for each query, representing the superpoint level predicted tree instances and their confidences. Simultaneously, a learnable semantic query tensor $\mathbf{Q}_{\text{sem}}\in\real^{2\times 32}$ is also passed to the decoder to produce semantic masks $\mathbf{M}^{\mathbfcal{S}}_{\text{sem}}\in\{0,1\}^{2\times M}$ corresponding to tree and non-tree superpoints.

The training setup (\Cref{section:scatr_its_training}) of SCATR-ITS largely follows that of ForestFormer3D. Since forest point clouds are typically too large to fit into GPU memory in their entirety, the data are processed in smaller cylinders. During training, cylinder center positions are sampled at random from each scene, and the predicted instance masks are matched to the ground truth instances using a one-to-many approach. During inference (\Cref{section:scatr_its_inference}), the inputs are processed using a cylindrical sliding window and then combined using score-based block merging, as in ForestFormer3D. However, SCATR-ITS adopts a maximal input stride designed to minimize the overlap between consecutive window positions and thereby reduce the amount of data processed. All postprocessing of the predicted instances, including non-maximum suppression (NMS), score normalization, erroneous-instance filtering, and score-based block merging, is performed at the superpoint level before the segmentation results are cast onto the original point cloud. To account for the high variance in superpoint size, the postprocessing is further made superpoint-size-aware such that the contribution of each superpoint is weighted by the number of points it contains.

Superpoint-level processing has previously been utilized quite extensively in general-purpose point cloud segmentation \citep[see e.g.][]{landrieu2018large,robert2023efficient,sun2023superpoint,geist2026ezsp}, including the original OneFormer3D model \citep{kolodiazhnyi2024oneformer3d}, and has also been successfully applied to semantic segmentation of forest data \citep{shao2023semantic,ruoppa2025unsupervised,takhtkeshha20263d}. In this work, we adapt the same principle to individual tree segmentation to reduce computational requirements. While the point clouds considered in most 3D segmentation studies are small enough to be processed in a single pass, forest point clouds are typically divided into several overlapping chunks due to their large size. This introduces an additional challenge for superpoint-based processing, which we address through the size-aware superpoint-level postprocessing.

\subsection{Superpoint generator} \label{section:superpoint_generator}

\begin{figure*}[!t]
\centering
\includegraphics[width=\textwidth]{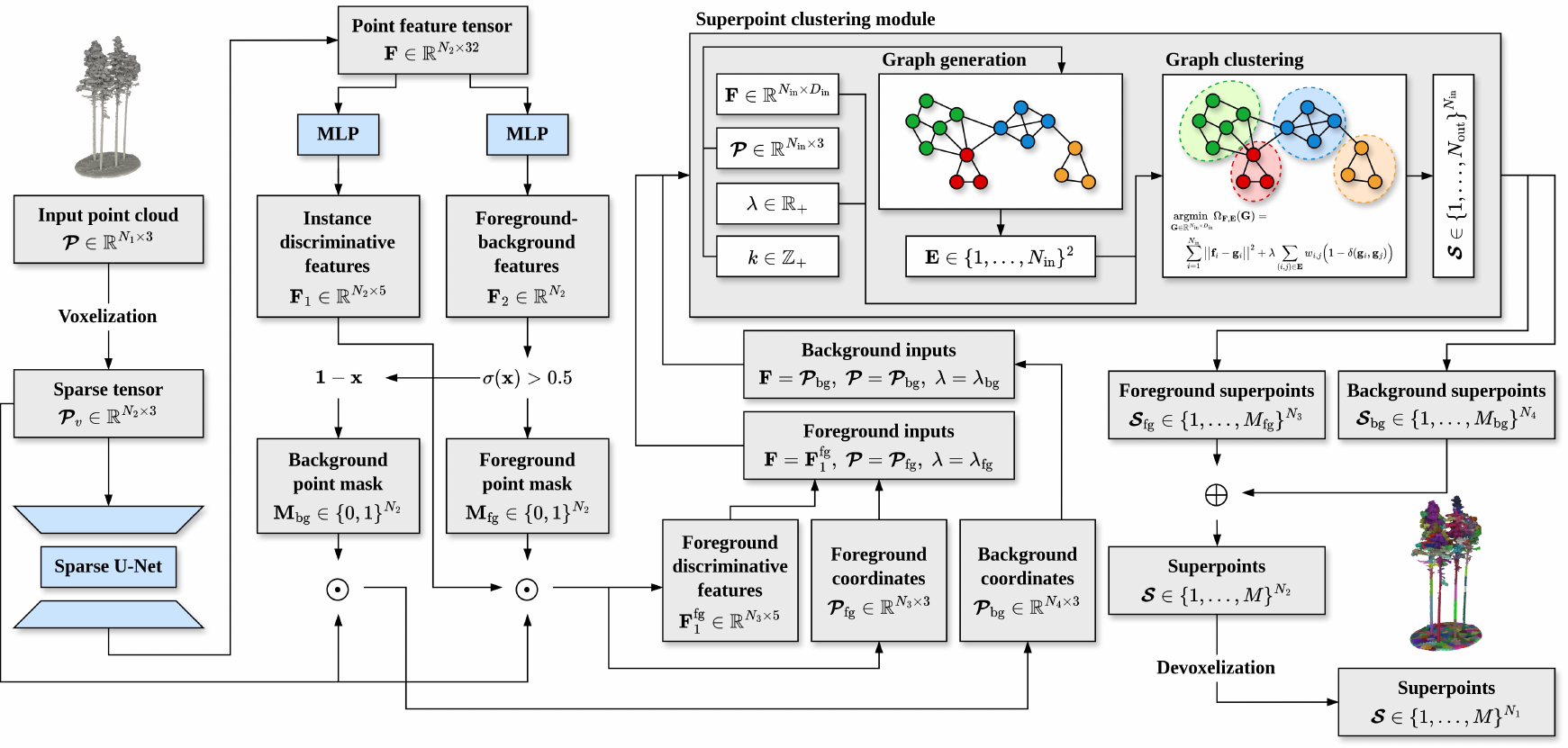}
\caption{Overview of the SCATR-ITS superpoint generator. Components with learnable weights are shown in light blue.}
\label{figure:superpoint_architecture}
\end{figure*}

SCATR-ITS employs a combination of learned feature embeddings and efficient graph partitioning for generating superpoints, essentially adapting the framework used by \citep{landrieu2019point} and \citep{geist2026ezsp} to forest data. As in the ITS stage, the superpoint generator uses a sparse U-Net backbone followed by two MLP branches to extract 5D instance-discriminative features $\mathbf{F}_1\in\real^{N_2\times 5}$ and foreground-background logits $\mathbf{F}_2\in\real^{N_2}$ from the input point cloud (\Cref{section:superpoint_backbone_network}). Next, a graph representation is constructed for the point cloud and extracted feature embeddings, and an efficient graph clustering algorithm is used to generate the superpoint partition (\Cref{section:superpoint_partitioning}). Notably, since $\mathbf{F}_1$ is a meaningful representation only for tree instances, predicted foreground and background points are handled separately.

Similarly to the actual segmentation phase, the full forest scenes are too large to process in a single pass, so smaller cylindrical subsets are randomly sampled during training (\Cref{section:superpoint_training}). Furthermore, a sliding cylindrical window is used during inference (\Cref{section:superpoint_inference}), and the individual results are combined into a scene-level superpoint partition using a greedy earliest-first merging strategy. An overview of the superpoint generation pipeline is shown in \Cref{figure:superpoint_architecture}.

\subsubsection{Superpoint generator backbone network} \label{section:superpoint_backbone_network}

The superpoint generator utilizes the exact same backbone as the main SCATR-ITS segmentation model, consisting of a sparse-convolution-based 3D U-Net \citep{choy20194d} followed by two MLP branches (see \Cref{section:scatr_its_backbone_network}). Equivalent to the full SCATR-ITS model, the input point clouds are voxelized on a uniform grid with a voxel size of 0.2 m. The lack of transformer-based decoding stage in the superpoint generator enables the use of significantly larger radii for the cylindrical input point clouds compared to those permitted in the ITS stage. The increased input extent, in turn, substantially improves inference speed when processing the full forest scenes. Consequently, we opted to train a separate backbone network with different input size hyperparameters for the superpoint generator. In addition, slight modifications were made to the loss function.

The predicted foreground-background logits $\mathbf{F}_2$ output by the other MLP branch are used to construct two binary masks $\mathbf{M}_{\text{fg}},\mathbf{M}_{\text{bg}}\in\{0,1\}^{N_2}$ corresponding to foreground and background points:
\begin{align}
    &\mathbf{M}_{\text{fg}}=\sigma\left(\mathbf{F}_2\right)>0.5,\\
    &\mathbf{M}_{\text{bg}}=1-\mathbf{M}_{\text{fg}},
\end{align}
where $\sigma(\cdot)$ is the sigmoid function. These masks are subsequently applied to the voxelized point cloud $\mathbfcal{P}_v\in\real^{N_2\times 3}$ and the instance-discriminative features $\mathbf{F}_1$ to extract the foreground instance-discriminative features, foreground coordinates, and background coordinates, denoted by $\mathbf{F}_1^{\text{fg}}\in\real^{N_3\times 5}$, $\mathbfcal{P}_{\text{fg}}\in\real^{N_3\times 3}$, and $\mathbfcal{P}_{\text{bg}}\in\real^{N_4\times 3}$ respectively. $N_3$ and $N_4$ denote the number of predicted foreground and background points, respectively, such that $N_3+N_4=N_2$. The resulting feature and coordinate tensors are then used as inputs for graph construction and partitioning in the subsequent stage of the pipeline.

\subsubsection{Superpoint partitioning} \label{section:superpoint_partitioning}

Following feature extraction, we employ a strategy similar to \citep{landrieu2019point} and \citep{geist2026ezsp} for generating the superpoints: the point cloud is transformed into a graph representation, with extracted features serving as node values, after which graph clustering is employed to create the superpoints. Unlike previous works on superpoint partitioning from graph representations with DL-based features, which have focused on semantic segmentation, our generator is designed to produce superpoints that respect instance boundaries. Consequently, since the extracted instance-discriminative features generally provide a reasonable representation only for tree instances, the superpoint partitioning stage is performed separately for predicted foreground and background points.

Given an input point cloud $\mathbfcal{P}\in\real^{N\times 3}$ and the corresponding features $\mathbf{F}\in\real^{N\times D}$, where $D$ is the feature dimensionality, the superpoint generator first constructs a directed $k$-nearest neighbors ($k$-NN) graph with edge set $\mathbf{E}^{\prime}$ and edge weights $w^{\prime}_{i,j}>0$, where $(i,j)\in\mathbf{E}^{\prime}$. The graph is then symmetrized by treating edges $(i,j)$ and $(j,i)$, when both are present, as a single undirected edge and combining their weights such that $w_{i,j}=w^{\prime}_{i,j}+w^{\prime}_{j,i}$. We denote the resulting undirected edge set by $\mathbf{E}$ and the corresponding weights by $w_{i,j}$. The task of finding an optimal graph partition can then be expressed as the following energy minimization problem \citep{landrieu2017cut}:
\begin{equation}
    \label{equation:energy_minimization}
    \underset{\mathbf{G\in\real^{\emph{N}\times \emph{D}}}}{\text{argmin }}\Omega_{\mathbf{F},\mathbf{E}}(\mathbf{G})=\sum_{i=1}^N \big|\big|\mathbf{f}_i-\mathbf{g}_i\big|\big|_2^2+\lambda\sum_{(i,j)\in\mathbf{E}}w_{i,j} \Big(1-\delta(\mathbf{g}_i,\mathbf{g}_j)\Big),
\end{equation}
where $||\cdot||_2$ is the Euclidean norm, $\delta(\cdot,\cdot)$ is the Kronecker delta, and $\lambda>0$ is the regularization strength. The solution to \Cref{equation:energy_minimization}, $\mathbf{G}^\ast\in\real^{N\times D}$, is a piecewise-constant approximation of $\mathbf{F}$. Retrieving the connected components of $\mathbf{G}^\ast$ then yields the optimal superpoint partition \citep{geist2026ezsp}.

Let $\mathbfcal{P}^{\mathbfcal{S}}$ denote a partition of $\mathbfcal{P}$ into superpoints. Formally, the superpoints $\mathbf{s}_i\in\mathbfcal{P}^{\mathbfcal{S}}$ are defined as the connected components of the $k$-NN graph such that:
\begin{align}
    \bigcup_{\mathbf{s}_i\in\mathbfcal{P}^{\mathbfcal{S}}}\mathbf{s}_i&=\mathbfcal{P} \\
    \bigcap_{\mathbf{s}_i\in\mathbfcal{P}^{\mathbfcal{S}}}\mathbf{s}_i&=\emptyset.
\end{align} Two superpoints $\mathbf{s}_i,\mathbf{s}_j\in\mathbfcal{P}^{\mathbfcal{S}}$ are connected by an undirected edge, if any of their underlying points are also connected by an edge in $\mathbf{E}$. We denote this superpoint-induced graph by $\mathbfcal{E}$, defined formally as:
\begin{equation}
    \mathbfcal{E}=\big\{(i,j)\;|\;\exists(k,l)\in\mathbf{E}\; :\;\mathbf{p}_k\in\mathbf{s}_i\land \mathbf{p}_l\in\mathbf{s}_j\big\}
\end{equation}

For each superpoint $\mathbf{s}_i$, we compute a feature representation $\mathbf{g}_{\mathbf{s}_i}$ by averaging over the feature embeddings of its points:
\begin{equation}
    \mathbf{g}_{\mathbf{s}_i}=\frac{1}{|\mathbf{s}_i|}\sum_{\mathbf{p}_j\in\mathbf{s}_i}\mathbf{f}_j,
\end{equation}
where $\mathbf{p}_j\in\mathbfcal{P}$ denotes the $j$th point in the point cloud, $\mathbf{f}_j\in\mathbf{F}$ the corresponding feature embedding, and $|\cdot|$ the cardinality of the set. We then define a feature embedding matrix $\mathbf{G}^{\mathbfcal{S}}\in\real^{N\times D}$ that associates each point with the feature representation of its corresponding superpoint:
\begin{equation}
    \forall i,\forall j:\mathbf{p}_j\in\mathbf{s}_i\implies\mathbf{G}^{\mathbfcal{S}}_{j}=\mathbf{g}_{\mathbf{s}_i}
\end{equation}
This enables reformulating \Cref{equation:energy_minimization} as a combinatorial optimization problem of minimizing $\Omega(\mathbfcal{P}^{\mathbfcal{S}})=\Omega_{\mathbf{F},\mathbf{E}}(\mathbf{G}^{\mathbfcal{S}})$ with respect to the partition $\mathbfcal{P}^{\mathbfcal{S}}$. As shown by \citet{geist2026ezsp}, merging adjacent superpoints $\mathbf{s}_i$ and $\mathbf{s}_j$ with $(i,j)\in\mathbfcal{E}$ decreases $\Omega(\mathbfcal{P}^{\mathbfcal{S}})$ by:
\begin{equation}
    \Delta(\mathbf{s}_i,\mathbf{s}_j)=-\frac{|\mathbf{s}_i||\mathbf{s}_j|}{|\mathbf{s}_i|+|\mathbf{s}_j|}\big|\big|\mathbf{g}_{\mathbf{s}_i}-\mathbf{g}_{\mathbf{s}_j}\big|\big|_2+\lambda\sum_{(k,l)\in\mathbf{E}_{i,j}}w_{k,l},
\end{equation}
where $\mathbf{E}_{i,j}\subseteq\mathbf{E}$ is the subset of edges connecting points in $\mathbf{s}_i$ to points in $\mathbf{s}_j$:
\begin{equation}
    \mathbf{E}_{i,j}=\big\{(k,l)\in\mathbf{E}\;|\;\mathbf{p}_k\in\mathbf{s}_i\land\mathbf{p}_l\in\mathbf{s}_j\big\}.
\end{equation}

We approximate the solution to \Cref{equation:energy_minimization} using the greedy and parallelized bottom-up algorithm of \citet{geist2026ezsp}. The algorithm takes as input the point cloud $\mathbfcal{P}$, the corresponding point-level features $\mathbf{F}$, which serve as the node values, and the initial $k$-NN graph edges $\mathbf{E}$. The superpoint partition and adjacency graph are initialized to $\mathbfcal{P}^{\mathbfcal{S}}=\big\{\{\mathbf{p}_i\}\;|\;\mathbf{p}_i\in\mathbfcal{P}\big\}$ and $\mathbfcal{E}=\mathbf{E}$, respectively. While possible merges remain, the algorithm repeats the following steps:
\begin{enumerate}
    \item \textbf{Construct candidate merges:} Construct $\mathbfcal{E}_{\text{merge}}$ by converting each edge $(i,j)\in\mathbfcal{E}$ that satisfies $\Delta(\mathbf{s}_i,\mathbf{s}_j)>0$ or $|\mathbf{s}_i|<\tau_{\text{sp}}$ into a directed edge $(i\rightarrow j)$. If $\mathbfcal{E}_{\text{merge}}=\emptyset$ or the maximum number of iterations has been reached, return $\mathbfcal{P}^{\mathbfcal{S}}$.
    \item \textbf{Remove conflicts:} To ensure there are no conflicting merges, each superpoint may only occur once as a source in $\mathbfcal{E}_{\text{merge}}$. For each superpoint $\mathbf{s}_i$, the edge $(i\rightarrow\cdot)$ with the highest gain $\Delta(\mathbf{s}_i,\cdot)$ is retained.
    \item \textbf{Merging:} The remaining merges $\mathbfcal{E}_{\text{merge}}$ define a directed merge graph $(\mathbfcal{P}^{\mathbfcal{S}},\mathbfcal{E}_{\text{merge}})$. The weakly connected components of this merge graph determine the superpoints to be merged, and $\mathbfcal{P}^{\mathbfcal{S}}$ is updated accordingly.
    \item \textbf{Recalibration:} The superpoint feature embeddings $\mathbf{G}^{\mathbfcal{S}}$ and the adjacency graph are updated to reflect the newly merged superpoints. The algorithm then returns to Step 1.
\end{enumerate}
Once the algorithm terminates, the final superpoint partition $\mathbfcal{P}^{\mathbfcal{S}}$ corresponds to the connected components of the approximated optimal solution $\mathbf{G}^\ast\approx\mathbf{G}^{\mathbfcal{S}}$. The superpoint labels $\mathbfcal{S}\in\{1,\ldots,M\}^N$, where $M=\left|\mathbfcal{P}^{\mathbfcal{S}}\right|$, can then be generated by assigning each of the $M$ components a unique integer identifier (ID) from the range $[1,M]$.

Superpoint generation is handled separately for predicted foreground and background points. For the foreground points, the foreground coordinates are used for graph generation, $\mathbfcal{P}=\mathbfcal{P}_{\text{fg}}$, with the corresponding instance-discriminative features serving as the node values, $\mathbf{F}=\mathbf{F}_1^{\text{fg}}$. On the other hand, since the extracted features $\mathbf{F}_1$ do not provide a meaningful representation for non-instance points, the $xyz$-coordinate values are used both for graph generation and as node values for the background points, $\mathbfcal{P}=\mathbf{F}=\mathbfcal{P_{\text{bg}}}$. Following \citet{geist2026ezsp}, we first construct the $k$-NN graph $\mathbf{E}^{\prime}$ of the $xyz$-coordinates with $k=8$ and assign $w^{\prime}_{i,j}=1$ for all $(i,j)\in\mathbf{E}^{\prime}$. The graph is subsequently symmetrized to obtain the initial adjacency graph $\mathbf{E}$, and the minimum superpoint size is set to $\tau_{\text{sp}}=1$. The foreground and background regularization strengths are set to $\lambda_{\text{fg}}=0.003$ and $\lambda_{\text{bg}}=0.3$, respectively. The value of $\lambda_{\text{fg}}$ was determined empirically with the aim of maximizing superpoint quality (see \Cref{section:superpoint_quality_metrics}) while simultaneously minimizing the total number of superpoints within each scene. By contrast, $\lambda_{\text{bg}}$ was chosen afterward by experimentally identifying a value that yielded background superpoints of similar size. Finally, the resulting foreground and background superpoint labels, denoted by $\mathbfcal{S}_{\text{fg}}\in\{1,\ldots,M_{\text{fg}}\}^{N_3}$ and $\mathbfcal{S}_{\text{bg}}\in\{1,\ldots,M_{\text{bg}}\}^{N_4}$, respectively, are combined into a single scene-level superpoint label vector $\mathbfcal{S}$ by concatenation:
\begin{equation}
    \mathbfcal{S}=\mathbfcal{S}_{\text{fg}}\oplus\left(\mathbfcal{S}_{\text{bg}}+M_{\text{fg}}\right)
    \in\{1,\ldots,M\}^{N_2},
\end{equation}
where $\oplus$ denotes concatenation and $M=M_{\text{fg}}+M_{\text{bg}}$.

\subsubsection{Superpoint generator training} \label{section:superpoint_training}

The superpoint generator training pipeline largely mirrors that of the full SCATR-ITS model (see \Cref{section:scatr_its_training}). Inputs are formed by randomly selecting a center point from the full point cloud and extracting a cylindrical region of a predetermined radius. For the foreground-background semantic head, the original binary cross-entropy loss is replaced with the Lovász hinge loss, since preliminary experiments indicated it provides a slight improvement in superpoint quality. The default instance-discriminative loss is retained for the 5D feature head. Additional details of the training procedure are provided in \Cref{appendix:scatr_its_training_details}.

\subsubsection{Superpoint generator inference} \label{section:superpoint_inference}

Similar to the full SCATR-ITS model, the superpoint generator processes point clouds as evenly spaced, overlapping cylinders. To accelerate inference, it employs the maximal input stride presented in \Cref{section:maximal_input_stride}, with maximum object size $o_{\max}=4$ m. The input cylinders are processed sequentially. First, a cylindrical region is extracted from the full scene and voxelized, after which superpoint labels are generated and the output is devoxelized by assigning all points within a voxel the corresponding superpoint label. To ensure scene-level IDs remain unique, generated superpoints within each cylinder are assigned consecutive integer labels, continuing from the maximum label across previous cylinders.

Inspired by ForestFormer3D, all superpoints containing one or more points with a margin of $\delta_s\in\real_0$ from the cylinder border are marked invalid. This ensures that only superpoints corresponding to objects fully contained within the cylinder are retained, mitigating edge effects caused by partitioning the full scene into cylindrical subsets. The margin $\delta_s$ is defined as the Euclidean distance in the $xy$-plane and set to twice the voxel size (0.4 m). Finally, the generated superpoint labels and corresponding global point indices for each cylinder are stored for merging.

Subsequent to generating the superpoint labels for each input cylinder, the results are merged into a single scene-level labeling. For instance segmentation, this is achieved with score-based merging (see \Cref{section:superpoint_level_postprocessing}), where each point receives the label of the corresponding instance with the highest confidence score. Because the superpoints lack confidence scores, we instead employ an earliest-first approach, assigning each point the first valid label based on the cylinder processing order. Although size-based ordering, where each point is assigned the label of the largest corresponding superpoint, was also tested, initial experiments showed no significant improvement in superpoint quality. Consequently, the computationally lighter earliest-first approach is used instead. The merged superpoint labels are mapped to a continuous range, yielding a scene-level labeling $\mathbfcal{S}\in\{1,\ldots,M\}^N$.

\subsection{Backbone network} \label{section:scatr_its_backbone_network}

Following \citet{sun2023superpoint,kolodiazhnyi2024oneformer3d,xiang2025forestformer3d}, SCATR-ITS employs a 3D sparse convolutional U-Net \citep{choy20194d,spconv2022spconv} backbone. The input point cloud $\mathbfcal{P}\in\real^{N_1\times3}$ is voxelized on a uniform grid of size 0.2 m, yielding $\mathbfcal{P}_v\in\real^{N_2\times 3}$. The voxelized point cloud is then transformed into a sparse tensor format and fed into the U-Net. An input layer consisting of a single submanifold 3D sparse convolution first maps the number of input channels from 3 to 32. The features are subsequently processed through a four-layer encoder. Each encoder layer contains two blocks consisting of two submanifold 3D sparse convolutional layers with ReLU activation, batch normalization and a residual connection, followed by a strided 3D sparse convolution that halves the input resolution and doubles the channel count. At the encoder bottleneck, the data have been downsampled to 1/16th of the original resolution with 160 channels. The 160D features then enter a four-layer decoder. Each layer consists of 3D sparse inverse convolution that doubles the resolution and halves the number of channels, followed by two blocks equivalent to those in the encoder. Skip connections link equivalent-resolution layers between the encoder and decoder, and the first block in each decoder layer restores the original channel count. The final output of the sparse U-Net is $\mathbf{F}\in\real^{N_2\times 32}$, a sparse tensor containing a 32D feature representation for each point in the voxelized point cloud.

Following ForestFormer3D, the extracted features are separated into two branches: a semantic branch that predicts foreground-background logits $\mathbf{F}_2\in\real^{N_2}$, and an instance-discriminative branch that extracts 5D feature embeddings $\mathbf{F}_1\in\real^{N_2\times5}$ designed to discern between individual tree instances. These embeddings are trained such that points belonging to the same instance have similar representations, while points from distinct instances are pushed apart. Both branches are implemented as MLPs consisting of two linear layers with LeakyReLU activations ($\alpha=0.2$).

\subsection{Instance decoder} \label{section:instance_decoder}

The extracted features $\mathbf{F}$, $\mathbf{F}_1$, and $\mathbf{F}_2$ are aggregated into superpoint-level representations $\mathbf{F}^{\mathbfcal{S}}\in\real^{M\times 32}$, $\mathbf{F}^{\mathbfcal{S}}_1\in\real^{M\times 5}$, and $\mathbf{F}^{\mathbfcal{S}}_2\in\real^{M}$ using on the labels $\mathbfcal{S}\in\{1,\ldots,M\}^{N_1}$ produced by the superpoint generator (\Cref{section:superpoint_pooling}). A dynamic query count module then determines the number of queries $K$ based on $\mathbf{F}^{\mathbfcal{S}}$, (\Cref{section:dynamic_query_count}) after which the ISA-guided query selection method of \citet{xiang2025forestformer3d} obtains the instance queries $\mathbf{Q}_{\text{inst}}\in\real^{K\times 32}$ (\Cref{section:query_selection}). These instance queries, alongside $\mathbf{F}^{\mathbfcal{S}}$ and two learnable semantic queries $\mathbf{Q}_{\text{sem}}\in\real^{2\times 32}$ are fed into a six-layer transformer decoder (\Cref{section:transformer_decoder}), which outputs the predicted superpoint-level tree instance masks $\mathbf{M}^{\mathbfcal{S}}_{\text{inst}}\in\{0,1\}^{K\times M}$, the corresponding confidence scores $\mathbf{S}_{\text{inst}}\in\real^K$, and tree--non-tree semantic masks $\mathbf{M}^{\mathbfcal{S}}_{\text{sem}}\in\{0,1\}^{2\times M}$. Notably, since the generated superpoints do not consistently align with the wood--foliage semantic boundaries, the SCATR-ITS decoder simplifies semantic classification into a binary tree/non-tree task, omitting the more intricate semantic classes of ForestFormer3D in favor of efficient superpoint-level processing.

\subsubsection{Superpoint pooling} \label{section:superpoint_pooling}

Prior to superpoint-level feature aggregation, the feature tensors are devoxelized by assigning all points within the same voxel the corresponding feature value. This ensures that variations in local point density are accounted for during aggregation. Following prior work \citep{sun2023superpoint,kolodiazhnyi2024oneformer3d}, the point-level features are aggregated using mean pooling. Specifically, let $\mathbf{i}_j$ denote the index set of points with the superpoint label $j\in\{1,\ldots,M\}$. The corresponding superpoint-level feature $\mathbf{f}_{\mathbf{i}_j}$ is obtained by averaging the feature representation of all points within $\mathbf{i}_j$:
\begin{equation}
    \mathbf{f}_{\mathbf{i}_j}=\frac{1}{|\mathbf{i}_j|}\sum_{k\in\mathbf{i}_j}\mathbf{f}_k,
\end{equation}
where $\mathbf{f}_k$ is the feature vector of the $k$th point. The feature representations of all $M$ superpoints are stacked to construct the final superpoint-level feature tensors $\mathbf{F}^{\mathbfcal{S}}$, $\mathbf{F}^{\mathbfcal{S}}_1$, and $\mathbf{F}^{\mathbfcal{S}}_2$.

    \subsubsection{Dynamic query count} \label{section:dynamic_query_count}

Point cloud mask transformers typically utilize a fixed number of queries, with ForestFormer3D employing $K=300$ non-parametric queries by default. The query count is set substantially higher than the expected maximum number of trees in each input to ensure most instances are assigned a query even in high-density forests. By contrast, sparse forests require far fewer queries, and a query count optimized for dense scenes results in high levels of redundancy, as instances are often assigned multiple queries. Consequently, due to the substantial structural variability in forest environments, a fixed query count is usually suboptimal, being either insufficient for dense scenes, or excessively large for sparse ones. Inspired by DQ-DETR \citep{huang2024dqdetr}, which proposed a categorical counting module for tiny object detection in aerial images, we introduce a dynamic query count module that adaptively selects the number of queries based on estimated forest density.

\begin{figure}[!t]
\centering
\includegraphics[width=0.45\textwidth]{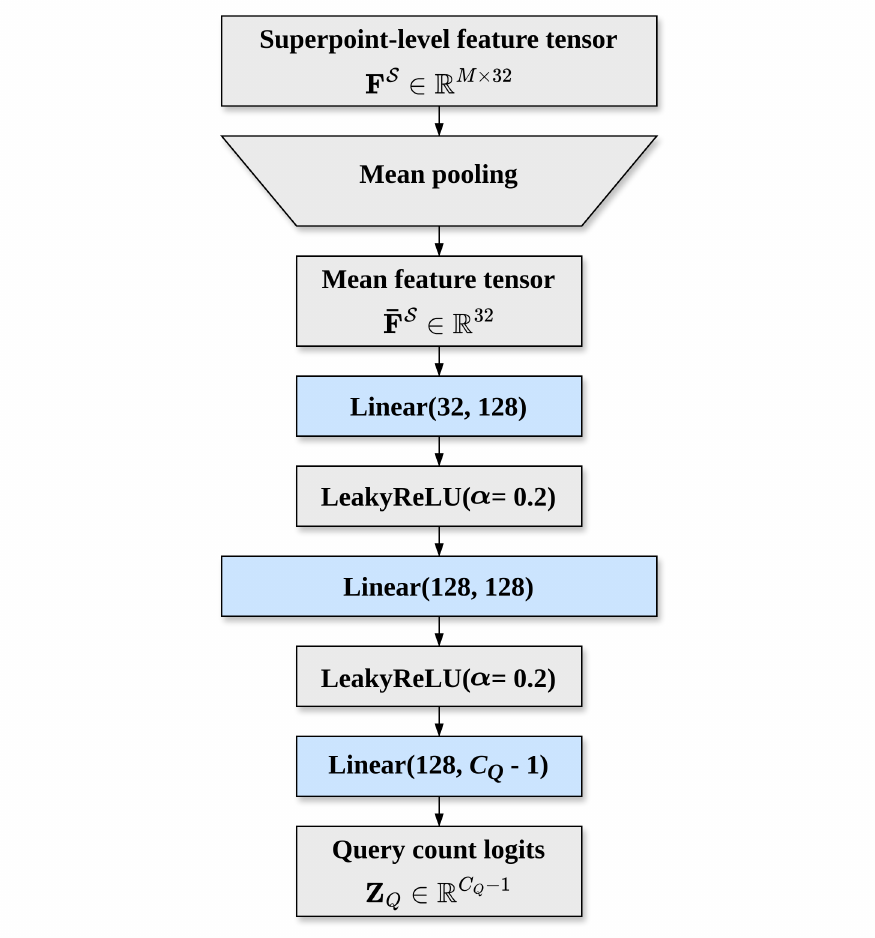}
\caption{Overview of the proposed dynamic query count module. Components with learnable weights are shown in light blue.}
\label{figure:dynamic_query_count_module}
\end{figure}

An overview of the dynamic query count module is shown in \Cref{figure:dynamic_query_count_module}. Given the superpoint-level features $\mathbf{F}^{\mathbfcal{S}}$, the module aggregates them into a single scene-level descriptor $\bar{\mathbf{F}}^{\mathbfcal{S}}\in\real^{32}$ via mean pooling. The feature vector is then fed into an MLP, similar to the semantic and instance-discriminative heads, consisting of linear layers and LeakyReLU activations with $\alpha=0.2$. Following \citet{huang2024dqdetr}, the module is trained to predict an instance density category rather than directly regressing the query count, as precise regression is challenging under high variance in scene density. The MLP outputs logits that assign each scene into one of $C_Q\in\integer_+$ query bins. Each bin $c_Q\in\{1,\ldots,C_Q\}$ is defined by an instance count interval $[a,b)$, where $a<b$, and maps to a corresponding query count $K\in\integer_0$. Furthermore, the bins are non-overlapping and form a contiguous sequence covering all non-negative integers starting from 0, with the upper bound of the final bin set to $\infty$.

Because the query bins are inherently ordered by their lower bounds, we adopt ordinal classification over standard multiclass classification to more effectively exploit the ordering information \citep{gutierrez2016ordinal,vargas2020cumulative}. Specifically, the dynamic query count module outputs logits $\mathbf{Z}_Q\in\real^{C_Q-1}$, where the $i$th element $z_{Q,i}$ represents the probability that the scene-level instance count surpasses the upper bound of the $i$th bin. The predicted query bin $c_Q$ is then obtained as the sum of ordinal predictions whose probability exceeds 0.5:
\begin{equation}
    c_Q=1+\sum_{i=1}^{C_Q-1}\big[\sigma\left(z_{Q,i}\big)>0.5\right].
\end{equation}
Finally, the predicted query bin is mapped to the corresponding query count $K$.

Using training and validation data across all three datasets, we configured the bin intervals and corresponding query counts. We opted for a total of five bins to account for the high variance in forest density. Bin intervals were determined experimentally by analyzing ground truth instance counts in sampled input cylinders under various bin configurations. Our findings indicate that intervals should be assigned such that approximately 30\% of inputs fall into the centermost bin (3), with remaining bins accounting for 10--25\% each, gradually decreasing toward the sparsest and densest extremes. For FGI-EMIT and FOR-InstanceV2, we set the query count for bin 3 to the default value of $K=300$. The remaining query counts were determined empirically based on average instance counts within each bin. For ALS-XL-20-pts, which utilizes an input radius twice that of FGI-EMIT, we found that tripling the query counts from the corresponding FGI-EMIT bins yielded optimal performance. The final query bins employed to train the models are presented in \Cref{table:query_bins}.

\begin{table}[!t]
    \centering
    \caption{Query bins used for training SCATR-ITS on the FGI-EMIT, FOR-InstanceV2, and ALS-XL-20-pts datasets.}
    \bigskip
    \scriptsize{
    \begin{tabular}{lclr}
        \toprule
        \textbf{Dataset} & \textbf{Bin} & \textbf{Interval} & \textbf{K} \\ \midrule \midrule
        \multirow{5}{*}{FGI-EMIT} & 1 & $[0,5)$ & 50 \\
        & 2 & $[5,10)$ & 100 \\
        & 3 & $[10,25)$ & 300 \\
        & 4 & $[25,50)$ & 450 \\
        & 5 & $[50,\infty)$ & 600 \\ \midrule
        \multirow{5}{*}{FOR-InstanceV2} & 1 & $[0,10)$ & 100 \\
        & 2 & $[10,20)$ & 250 \\
        & 3 & $[20,40)$ & 300 \\
        & 4 & $[40,80)$ & 400 \\
        & 5 & $[80,\infty)$ & 500 \\ \midrule
        \multirow{5}{*}{ALS-XL-20-pts} & 1 & $[0,15)$ & 150 \\
        & 2 & $[15,25)$ & 300 \\
        & 3 & $[25,45)$ & 900 \\
        & 4 & $[45,75)$ & 1350 \\
        & 5 & $[75,\infty)$ & 1800 \\ \bottomrule
    \end{tabular}
    }
    \label{table:query_bins}
\end{table}

\subsubsection{Query selection} \label{section:query_selection}

Query selection follows the ISA-guided strategy of \citet{xiang2025forestformer3d} applied at the superpoint level. First, the 5D instance-discriminative features corresponding to predicted foreground superpoints are extracted from $\mathbf{F}_1^{\mathbfcal{S}}$ using $\mathbf{F}_2^{\mathbfcal{S}}$. FPS is then applied to the resulting superpoint features to select $K$ query positions. The corresponding feature values are extracted from $\mathbf{F}^{\mathbfcal{S}}$ and used as the instance queries $\mathbf{Q}_{\text{inst}}\in\real^{K\times 32}$.

\subsubsection{Transformer decoder} \label{section:transformer_decoder}

SCATR-ITS adopts the transformer-based query decoder from previous work \citep{sun2023superpoint,kolodiazhnyi2024oneformer3d,xiang2025forestformer3d}. Given the features $\mathbf{F}^{\mathbfcal{S}}$ and the instance and semantic queries $\mathbf{Q}_{\text{inst}},\mathbf{Q}_{\text{sem}}$, the decoder first projects them into 256D representations using two MLPs. An additional MLP transforms $\mathbf{F}^{\mathbfcal{S}}$ into separate 256D mask-aware features. These mask-aware features, along with the projected queries, are fed to a prediction head that outputs predicted mask logits and corresponding scores for each query. Furthermore, the head produces attention masks, which are used to constrain cross-attention to the predicted instance superpoints for each query.

The attention masks, projected superpoint features, and queries are processed through six transformer decoder layers. Each layer $l\in\{1,\ldots,6\}$ consists of masked cross-attention for the features and queries, followed by self-attention and an MLP block. The prediction head is applied to the final output $\mathbf{Z}_l$, along with the mask features, to generate the predicted mask logits and corresponding scores for the $l$th layer. Finally, $\mathbf{Z}_l$ and the updated attention masks serve as the input for the subsequent decoder layer.

The final layer outputs the superpoint-level instance mask logits $\mathbf{Z}_{\text{inst}}^{\mathbfcal{S}}\in\real^{K\times M}$, the corresponding confidence scores $\mathbf{S}_{\text{inst}}\in\real^K$, and the semantic mask logits $\mathbf{Z}_{\text{sem}}^{\mathbfcal{S}}\in\real^{2\times M}$. The logits are transformed into binary masks $\mathbf{M}_{\text{inst}}^{\mathbfcal{S}}\in\{0,1\}^{K\times M}$ and $\mathbf{M}_{\text{sem}}^{\mathbfcal{S}}\in\{0,1\}^{2\times M}$ through sigmoid activation and thresholding.

\subsection{Training} \label{section:scatr_its_training}

Given that SCATR-ITS is based on the same general framework as ForestFormer3D, we adopt its training procedure. During training, input point clouds are generated by extracting fixed radius cylinders with randomly sampled center points from the full scene. The one-to-many matching strategy of \citet{xiang2025forestformer3d} pairs predicted instances with ground truth during training: each ground truth tree is matched to the set of predictions whose queries land within it, and the predictions are then supervised using the ground truth mask of the match. Queries corresponding to non-instance points are supervised with an empty mask. Because queries are selected from the 5D feature representations through FPS, their positions in the original point cloud are explicitly known, rendering the matching trivial. 

The loss function of SCATR-ITS is equivalent to that of ForestFormer3D, with the exception of the transformer decoder losses being computed at the superpoint level rather than voxel level. Furthermore, a BCE loss is added to supervise the dynamic query count module. Additional details of the SCATR-ITS training pipeline are provided in \Cref{appendix:scatr_its_training_details}.

\subsection{Inference} \label{section:scatr_its_inference}

Similar to the training setup, the inference pipeline is largely adapted from ForestFormer3D. During inference, SCATR-ITS processes the full point cloud sequentially in overlapping cylindrical regions. To accelerate processing, a maximal input stride is employed (\Cref{section:maximal_input_stride}). For each input cylinder, the model extracts features, retrieves queries, and predicts instance and semantic masks. Identical to the superpoint generator, instances within $\delta_s=0.4$ m of the cylinder boundary are discarded to reduce edge effects from the cylinder-level processing. Score normalization and non-maximum suppression are then applied to the predicted instance masks, and any instances with confidence scores below 0.3 or containing fewer than 10 points are filtered out. For a given cylinder, the highest-confidence instance mask for each superpoint is compared across all previously processed cylinders. Wherever scores improve or predictions are missing, the corresponding instances are stored as prospective trees and the global scores are updated.

Final scene-level predictions are constructed using the score-based block merging algorithm of \citet{xiang2025forestformer3d}. The prospective instances are processed in ascending order by confidence score. If an instance prediction shares more than 30\% of its superpoints with previously accepted instances, it is discarded. Otherwise, any superpoints with no prior prediction are assigned the corresponding instance label. The final semantic masks are obtained via a majority vote across all cylinder-level predictions. 

To conserve computational resources, all postprocessing is conducted at the superpoint level (\Cref{section:superpoint_level_postprocessing}), and the results are projected back to the original point cloud only as a final step.

\subsubsection{Maximal input stride} \label{section:maximal_input_stride}

\begin{figure*}[!t]
\centering
\includegraphics[width=\textwidth]{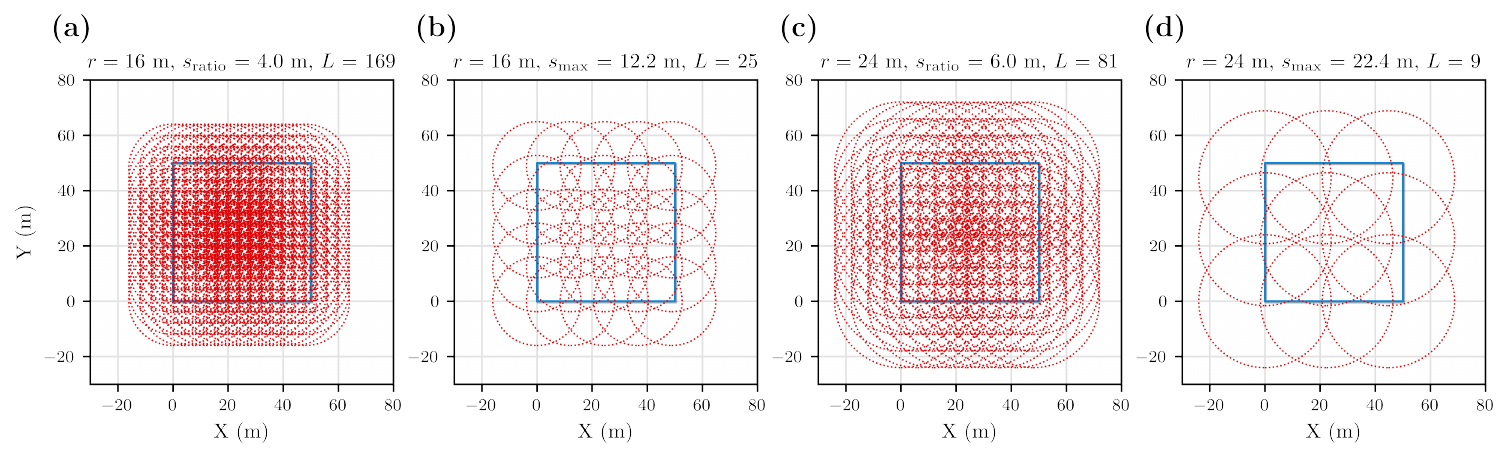}
\caption{Comparison of input cylinder positions produced by the original fixed-ratio stride ($s_{\text{ratio}}$) and the proposed maximal  stride ($s_{\max}$) on $50\times 50$ m plot. Results are shown for input cylinder radii $r=16$ m and $r=24$ m. The plot boundaries are shown in blue, with the cylinder locations superimposed in red. $L$ denotes the total number of input cylinders corresponding to the given stride. The original stride formulation $s_{\text{ratio}}$ uses 25\% of the cylinder radius as the step size. The maximal stride $s_{\max}$ is computed using $\delta=0.2$ m, $o_{\max}=6$ m, and $\alpha=0.9$. (a) Original stride formulation, $r=16$ m. (b) Proposed maximal stride, $r=16$ m. (c) Original stride formulation, $r=24$ m. (d) Proposed maximal stride, $r=24$ m.}
\label{figure:stride_comparison}
\end{figure*}

For an input radius of $r$, ForestFormer3D uses a fixed ratio of $0.25r$ as the step between the input cylinder centers during inference. This results in substantial overlap between the cylinders, thus introducing significant computational redundancy, as most points are fed to the inference pipeline multiple times. Reducing this overlap offers a straightforward method for decreasing data volume and accelerating inference.

The theoretical maximum stride that still covers all points within the scene is $\sqrt{2}r$. However, because instances within $\delta_s=0.4$ m of the cylinder boundary are discarded during inference, the effective radius with valid predictions is smaller. In the worst-case scenario, where a large instance is located within $\delta_s$ of the boundary, the effective radius is $r-\delta_s-o_{\max}$, where $o_{\max}\in\real_+$ denotes the maximum instance width in the dataset. The maximum admissible stride ensuring complete scene-level coverage is given by:
\begin{equation}
    s_{\max}=\sqrt{2}\left(r-\delta_s-o_{\max}\right).
\end{equation}
Empirically, we find that $o_{\max}=6$ m yields the best balance between accuracy and speed on the evaluated datasets. In practice, scaling the maximal stride by a slight safety margin $\alpha_s\in(0,1]$ is preferable to account for instances that exceed the estimated maximum object width. We set $\alpha_s=0.9$ and use $\alpha_s s_{\max}$ as the input stride. A qualitative comparison of input cylinder positions produced by the original fixed-ratio stride and the maximal stride is shown in \Cref{figure:stride_comparison}.

While our experiments indicate that high levels of redundancy and overlap in the instance predictions can be beneficial for model accuracy, and employing the proposed maximal stride generally results in a slight accuracy drop compared to the fixed-ratio approach, the computational efficiency gains far outweigh the minor accuracy reduction.

\subsubsection{Superpoint-level postprocessing} \label{section:superpoint_level_postprocessing}

To accelerate inference and minimize computational overhead, all postprocessing is performed at the superpoint level. This includes every step following the instance decoder: the highest-confidence instance masks from each input cylinder are stored solely at the superpoint level, and score normalization, non-maximum suppression, and block merging are all applied directly to superpoint representations.

Because the number of points within each superpoint varies, certain postprocessing metrics, such as IoU and instance overlap percentage, can differ significantly from true point-level values if computed directly at the superpoint level. To this end, we introduce a size-aware strategy where the contribution of each superpoint is weighted by its point count. As an example, the size-weighted superpoint-level IoU is given by:
\begin{equation}
    \text{IoU}(\mathbf{A}^{\mathbfcal{S}},\mathbf{B}^{\mathbfcal{S}})=\frac{\sum_{i=1}^M m_i a_ib_i}{\sum_{i=1}^M m_i a_i+\sum_{i=1}^M m_i b_i-\sum_{i=1}^M m_i a_ib_i},
\end{equation}
where $\mathbf{A}^{\mathbfcal{S}},\mathbf{B}^{\mathbfcal{S}}\in\{0,1\}^{M}$ are superpoint-level binary masks from the same scene, $a_i$ and $b_i$ denote their $i$th elements, respectively, and $m_i$ is the size of the $i$th superpoint.

The size-weighted formulations are mathematically equivalent to the original point-level values for all statistics required during postprocessing. Consequently, our superpoint-level postprocessing approach yields performance identical to that achieved by processing data at the point level.

\subsection{Experimental setup}

\subsubsection{Evaluation metrics for individual tree segmentation quality} \label{section:evaluation_metrics}

We adopt the FGI-EMIT evaluation protocol for assessing the accuracy of individual tree segmentation models \citep{ruoppa2026benchmarking}. Apart from the inclusion of crown-category-level metrics and Average Precision (AP), the protocol is equivalent to those used in most modern 3D DL-based ITS studies \citep[see e.g.][]{henrich2024treelearn,xiang2024automated,xiang2025forestformer3d,wielgosz2026segmentanytreev2}.

As a first step in evaluating ITS accuracy, one-to-one correspondences between predicted and ground truth tree instances are established using the intersection over union (IoU) metric. Given the set of ground truth instances $\mathbfcal{\tilde{P}}^{\text{gt}}$ and the set of predicted individual trees $\mathbfcal{\tilde{P}}^{\text{pred}}$, the IoU between the $i$th ground truth instance and the $j$th predicted segment, denoted by $\mathbfcal{P}^{\text{gt}}_i\in\mathbfcal{\tilde{P}}^{\text{gt}}$ and $\mathbfcal{P}^{\text{pred}}_j\in\mathbfcal{\tilde{P}}^{\text{pred}}$, respectively, is defined as:
\begin{equation}
    \text{IoU}\left(\mathbfcal{P}^\text{gt}_i,\mathbfcal{P}^\text{pred}_j\right)=\frac{N^\text{int}_{i,j}}{N_i^\text{gt}+N_j^{\text{pred}}-N_{i,j}^{\text{int}}},
\end{equation}
where $N_i^\text{gt}$ and $N_j^{\text{pred}}$ denote the number of points in $\mathcal{P}^\text{gt}_i$ and $\mathcal{P}^\text{pred}_j$, respectively, while $N^\text{int}_{i,j}$ denotes the number of points in their intersection. The maximum IoU of $\mathbfcal{P}_i^{\text{gt}}$ with respect to the set of all predicted instances is then:
\begin{equation}
    \text{maxIoU}\left(\mathbfcal{P}^{\text{gt}}_i\right)=\max_{\mathbfcal{P}^{\text{pred}}_j\in\tilde{\mathbfcal{P}}^{\text{pred}}}\left\{\text{IoU}\left(\mathbfcal{P}^{\text{gt}}_i,\mathbfcal{P}^{\text{pred}}_ j\right)\right\}.
\end{equation}
Each ground truth instance $\mathbfcal{P}_i^{\text{gt}}$ is matched to the predicted instance $\mathbfcal{P}^{\text{pred}}_j$ corresponding to $\text{maxIoU}\left(\mathbfcal{P}^{\text{gt}}_i\right)$. A threshold of $\text{IoU}_{\text{th}}=50\%$ is then applied, and all matches below it are discarded. As shown by \citet{kirillov2019panoptic}, setting the IoU threshold to $\geq50\%$ also ensures that the retained matches are unique.

For the Finnish national laser scanning data, for which only reference tree locations and heights are available, we employ the Hausdorff-distance-inspired matching approach of \citet{yu2006change}. A ground truth instance $\mathbfcal{P}^{\text{gt}}_i$ and predicted segment $\mathbfcal{P}^{\text{pred}}_j$ are matched if each is the nearest counterpart to the other among all ground truth instances $\tilde{\mathbfcal{P}}^{\text{gt}}$ and predicted segments $\tilde{\mathbfcal{P}}^{\text{pred}}$, respectively. Proximity is determined using the 3D Euclidean distance between tree locations and heights. Matches with a distance greater than a threshold $d_{\text{th}}$ are rejected to avoid matching segments that are too far apart to reasonably represent the same tree. Typical values for $d_{\text{th}}$ include 3 m \citep{hyyppa2024concepts} and 5 m \citep{yu2017single,hakula2023individual}. Locations and heights of predicted tree segments are computed as described in \citet{ruoppa2026benchmarking}.

All predictions with a matching ground truth instance are counted as true positives (TPs), while unmatched predictions are classified as false positives (FPs). Finally, ground truth instances with no matching predicted segment are considered false negatives (FNs). ITS performance is then evaluated using the standard Precision, Recall, and F$_1$-score:
\begin{align}
    &\text{Precision}=\frac{\text{TP}}{\text{TP}+\text{FP}}\\
    &\text{Recall}=\frac{\text{TP}}{\text{TP}+\text{FN}}\\
    &\text{F$_1$}=2\cdot\frac{\text{Precision}\cdot\text{Recall}}{\text{Precision}+\text{Recall}}
\end{align}
Crown-category-level performance for category $\text{X}\in\{\text{A},\text{B},\text{C},\text{D}\}$ is assessed using category-level Recall:
\begin{equation}
    \text{Recall}_{\text{X}}=\frac{\text{TP}_{\text{X}}}{\text{TP}_{\text{X}}+\text{FN}_{\text{X}}}.
\end{equation}

In addition to the conventional ITS accuracy metrics, we compute the Average Precision and Coverage (Cov). Defined as the area below the Precision--Recall curve, AP provides an effective instance segmentation metric that simultaneously accounts for overall segmentation accuracy and model confidence. Coverage, on the other hand, describes the overall alignment between the set of predicted tree segments and ground truth instances:  
\begin{equation}
    \text{Cov}=\frac{1}{\left|\tilde{\mathbfcal{P}}^{\text{gt}}\right|}\sum_{\mathbfcal{P}^{\text{gt}}_i \in\tilde{\mathbfcal{P}}^{\text{gt}}}\text{maxIoU}\left(\mathbfcal{P}^{\text{gt}}_i\right).
\end{equation}
$\text{maxIoU}\left(\cdot\right)$ values below the threshold $\text{IoU}_{\text{th}}$ are also included when computing the Coverage, which effectively serves as a proxy for the average segment quality.

Following \citet{ruoppa2026benchmarking}, all metrics are computed for the entire data split rather than being averaged across scene-level metrics. Any unlabeled tree points (semantic class 5 in FGI-EMIT) are filtered out prior to evaluation. Similarly, detected objects that are too small to represent actual trees are reclassified as background before computing the accuracy metrics. The minimum height and point-count thresholds are set to the standard values of 1.5 m and 40 points, respectively. For ALS-XL-20-pts and the Finnish national laser scanning data, the minimum point count is reduced to 5 points due to the significantly lower point densities. Object filtering was disabled on FOR-InstanceV2 for consistency with previous works.

\subsubsection{Evaluation metrics for superpoint quality} \label{section:superpoint_quality_metrics}

The quality of the superpoints has a significant influence on model performance, since the accuracy of any part of the pipeline that uses superpoint representations is bounded by the accuracy of said superpoints. Ideally, the superpoint generator would produce superpoints that perfectly align with the ground truth instances. However, this is rarely the case in practice, thus, evaluating the quality of the generated superpoints before incorporating them into the full segmentation pipeline is crucial.

Following \citet{landrieu2019point,lu2025serialization,geist2026ezsp}, we employ so-called \emph{Oracle} metrics to assess the quality of the superpoints. The Oracle metrics reflect model performance under the assumption of perfect predictions and are computed by assigning each superpoint the majority instance label among the points it contains, which is then treated as its predicted label. In practice, each Oracle metric represents the upper bound of the corresponding accuracy metric that can be achieved using the given superpoints. We assess superpoint accuracy using Oracle Precision, Recall, F$_1$-score, and Coverage, denoted by the ``O'' prefix, e.g. OPrecision for Oracle Precision.

To provide a robust quantitative assessment of superpoint quality, we additionally employ Boundary Precision (BP) and Boundary Recall (BR) \citep{landrieu2019point}:
\begin{align}
    &\text{BP}=\frac{\left|\mathbf{E}_{\text{inter}}^{\text{pred}}\cap \mathbf{E}_{\text{inter}}^{(1)}\right|}{\left|\mathbf{E}_{\text{inter}}^{\text{pred}}\right|}\\
    &\text{BR}=\frac{\left|\mathbf{E}_{\text{inter}}^{\text{pred}}\cap \mathbf{E}_{\text{inter}}^{(1)}\right|}{\left|\mathbf{E}_{\text{inter}}^{\text{\phantom{pred}}}\right|},
\end{align}
where $\mathbf{E}$ is the set of edges in the graph representation of the point cloud, $\mathbf{E}_{\text{inter}}\subseteq \mathbf{E}$ is the set of transitions between distinct ground truth instances, and $\mathbf{E}_{\text{inter}}^{\text{pred}}\subseteq \mathbf{E}$ is the set of transitions between generated superpoints. All edges between points corresponding to different instances are considered transitions. For simplicity, all non-instance points are treated as a single instance when computing BP and BR. Following \citet{landrieu2019point}, both metrics are computed with a tolerance of one edge by relaxing $\mathbf{E}_{\text{inter}}$ as follows:
\begin{equation}
    \mathbf{E}_{\text{inter}}^{(1)}=\big\{(i,j)\in \mathbf{E}\;|\;\exists k\; :\; (i,k)\in \mathbf{E}_{\text{inter}}\lor(j,k)\in \mathbf{E}_{\text{inter}}\big\}.
\end{equation}
In contrast to the Oracle metrics, BR and BP consider alignment with actual instance boundaries rather than just majority class membership, rendering them useful auxiliary metrics for assessing superpoint quality.

In addition to being aligned with the ground truth instances, a useful superpoint representation should contain substantially fewer elements than the original point cloud. In other words, the number of superpoints $M$ should be much smaller than the number of points $N$, formally $M\ll N$. Simply assigning each point a distinct superpoint label would yield perfect Oracle metrics, but the resulting representation would not compress the input in any way, rendering it useless for improving computational efficiency. Consequently, we assess the compactness of the superpoint representation using the average number of superpoints per scene, denoted by $\overline{M}$. In addition, we report the reduction ratio (RR):
\begin{equation}
    \text{RR}=\frac{M_{\text{all}}}{N_{\text{all}}},
\end{equation}
where $N_{\text{all}}$ and $M_{\text{all}}$ denote the total number of points and superpoints in a data split, respectively. A well-performing superpoint generator will produce superpoints that are both accurate and sufficiently compact to significantly reduce the number of inputs.

\subsubsection{Evaluation metrics for computational cost and speed}

Since improved computational efficiency is a central contribution of SCATR-ITS, we extensively evaluate its speed and resource consumption against all performance baselines. Model runtime is measured by separately recording the wall-clock time required for preprocessing and segmentation. Here, preprocessing is defined as all steps performed prior to segmentation, such as superpoint generation, image projection, and transforming the point cloud data into the required input format, whereas segmentation includes all subsequent operations, specifically segmentation itself and any necessary postprocessing. For each model, preprocessing and segmentation are executed five times, and the final runtime is reported as the mean across these runs, averaged across the forest plots in the corresponding data split. Each full point cloud is counted as a distinct forest plot. We report the average preprocessing (Pre.) and segmentation (Seg.) runtimes separately, as well as their sum, referred to as the total (Tot.) runtime. Each quantity is rounded separately when reported.

In addition to the average plot-level runtimes, we report the model throughput to facilitate comparisons across datasets with varying point densities and plot sizes. Throughput is defined as follows:
\begin{equation}
    \text{Throughput}=\frac{N_{\text{all}}}{t_{\text{tot}}},
\end{equation}
where $t_{\text{tot}}$ denotes the total model runtime for the entire data split (averaged across five runs). While forest structure, plot size, and point density still heavily influence model runtime, throughput provides a more objective measure of processing speed by expressing the number of points processed per unit of time.

Besides runtime, the computational cost of the models is also of high interest. Unlike conventional unsupervised ITS algorithms, modern DL-based forest segmentation models perform most of their computation on the GPU. Consequently, GPU VRAM consumption is the most appropriate indicator of computational cost for the tested models. Specifically, we utilize peak GPU VRAM usage as the evaluation metric. Although metrics such as average VRAM usage can also be of interest, for example, when assessing the energy consumption of models, we focus on peak usage since it represents a hard memory constraint that dictates whether a model can be executed on a given system. We report the peak GPU VRAM consumption in gibibytes (GiB) during both training and inference. VRAM usage statistics are recorded using the \verb|nvidia-smi|\footnote{\url{https://docs.nvidia.com/deploy/nvidia-smi}} command-line tool.

\subsubsection{Performance baselines}

The performance of SCATR-ITS was compared with four existing DL-based ITS methods on the FGI-EMIT and FOR-InstanceV2 datasets. Specifically, we selected the same models as \citet{ruoppa2026benchmarking}: YOLOv12 \citep{tian2025yolov12}, SegmentAnyTree \citep{wielgosz2024segmentanytree}, TreeLearn \citep{henrich2024extended}, and ForestFormer3D \citep{xiang2023review}. These models encompass a wide range of methodological approaches and have also been employed as performance baselines in several recent works \citep{nguyen2026forestmamba,wielgosz2026segmentanytreev2,yue2026toward}.

The first baseline, YOLOv12 \citep{tian2025yolov12}, has been adapted for ITS following the strategy of \citet{straker2023instance}. The point cloud is processed with a cloth simulation filter (CSF) \citep{zhang2016easy} to detect ground points, and subsequently normalized and transformed into a 2D CHM image. Tree crowns are then detected from the image using YOLOv12, and the resulting segments are cast back onto the original 3D point cloud. While such 2D-based ITS methods are typically significantly less accurate than their 3D counterparts, YOLOv12 is currently one of the most computationally efficient approaches for DL-based ITS \citep{ruoppa2026benchmarking}, and thus an interesting point of comparison. 

SegmentAnyTree \citep{wielgosz2024segmentanytree} and TreeLearn \citep{henrich2024treelearn} both employ deep clustering, a methodological approach that has become well established in DL-based ITS. The general framework consists of learning instance-discriminative and semantic features, which are subsequently clustered to construct tree instances. SegmentAnyTree aims to make the model sensor-agnostic through a data augmentation strategy that artificially sparsifies the inputs. In contrast, TreeLearn is designed for ground-based LiDAR and leverages the verticality of tree stems.

Finally, ForestFormer3D \citep{xiang2025forestformer3d} is a modern 3D mask transformer adapted for forest data. It is particularly relevant as a baseline since it represents the current state of the art in ITS and employs the same general framework as SCATR-ITS.

Since existing DL-based ITS methods are too computationally demanding for segmenting large areas within a reasonable timeframe, we instead use the unsupervised watershed ITS algorithm \citep{yu2011predicting} as the performance baseline for the Finnish national laser scanning data. Similar to YOLOv12, the algorithm first detects ground points using CSF, and then transforms the point cloud into a CHM. Tree segments are subsequently generated using a combination of local maxima detection and the watershed transform. Watershed was selected because it appears to be the most common approach for segmenting LiDAR data covering extensive geographical areas and has been employed in several previous large-area ITS studies \citep{ma2023individual,hyyppa2024concepts,itakura2026large}. 

SCATR-ITS and all performance baselines, except watershed, were trained separately on FGI-EMIT and FOR-InstanceV2. For evaluation on Finnish national laser scanning data, SCATR-ITS was trained on ALS-XL-20-pts, while the watershed hyperparameters were optimized on the same dataset. Where official pretrained weights were available, they were used. Otherwise, the models were trained from scratch. Details of the model training and hyperparameter optimization procedures are provided in \Cref{appendix:baseline_training_details}.

All experiments were performed on the same system running Ubuntu 22.04.5 LTS. The hardware specifications are listed in \Cref{table:computational_hardware}. Publicly available official source code was used for all performance baselines except ForestFormer3D and watershed, for which we used FGI's faster internal implementations.

\begin{table}[!t]
    \centering
    \caption{Specifications of the computing hardware used in the experiments.} \bigskip
    \scriptsize{
    \begin{tabular}{ll}
        \toprule
        \textbf{Device} & \textbf{Specifications} \\ \midrule \midrule
        CPU & Intel\textsuperscript{{\textregistered}} Xeon\textsuperscript{{\textregistered}} w5-3425 \\
        GPU & NVIDIA\textsuperscript{{\textregistered}} RTX\textsuperscript{\texttrademark} A6000 48 GB GDDR6 \\
        Memory & $8\times64$ GB DDR5 4800 MHz \\ 
        \bottomrule
    \end{tabular}}
    \label{table:computational_hardware}
\end{table}

\section{Results}

\subsection{Performance of the superpoint generator}

\begin{figure*}[!t]
\centering
\includegraphics[width=\textwidth]{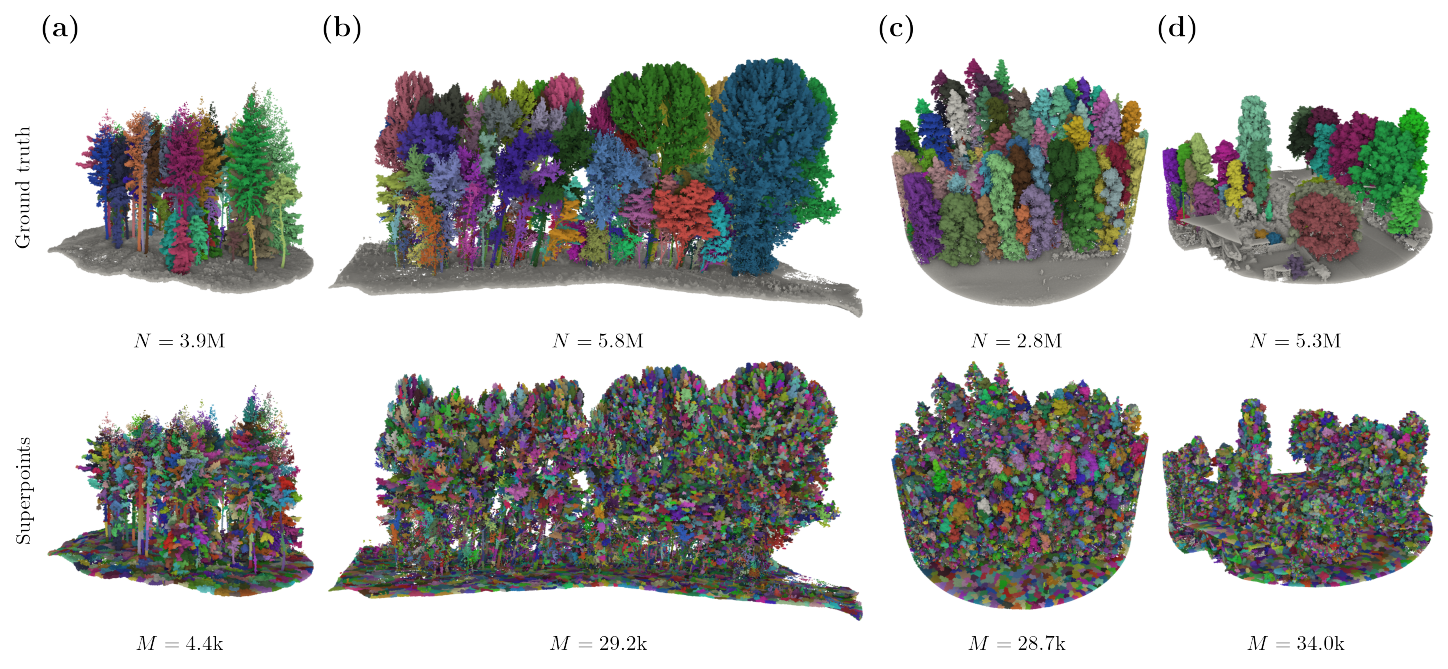}
\caption{Examples of results produced by the proposed superpoint generator, compared with the ground truth instance annotations on the FOR-InstanceV2 and FGI-EMIT datasets. $N$ denotes the number of points in the original point cloud, and $M$ denotes the corresponding number of superpoints. For ground truth annotations, each tree instance is shown in a distinct color and non-tree points are in gray, whereas superpoint results depict each predicted superpoint in a distinct color. (a) Plot 57 from the NIBIO\_MLS region in the FOR-InstanceV2 training set. (b) TUWIEN region plot from the FOR-InstanceV2 training set. (c) Plot 1018 from the FGI-EMIT test set. (d) Plot 1023 from the FGI-EMIT training set.}
\label{figure:superpoint_results}
\end{figure*}

\begin{table*}[!t]
    \centering
    \caption{Quantitative performance of the superpoint generator on the FGI-EMIT, FOR-InstanceV2, and ALS-XL-20-pts datasets.}
    \bigskip
    \scriptsize{
    \begin{tabular}{
        ll|
        *{1}{S[table-format=2.1]}
        *{7}{S[table-format=2.1]}}
        \toprule
        \textbf{Dataset} & \textbf{Split} & $\overline{\mathbf{M}}$ & \textbf{RR (\%)} & \textbf{OPrecision (\%)} & \textbf{ORecall (\%)} & \textbf{OF$_{\mathbf{1}}$ (\%)} & \textbf{OCov (\%)} & \textbf{BP (\%)} & \textbf{BR (\%)} \\
        \midrule \midrule
        \multirow{3}{*}{FGI-EMIT} & Training & 19.0{\text{k}} & 0.5 & 99.9 & 99.8 & 99.8 & 94.7 & 2.4 & 64.2 \\
        & Validation & 8.8{\text{k}} & 0.3 & 99.6 & 99.6 & 99.6 & 89.7 & 2.2 & 64.1 \\
        & Test & 15.1{\text{k}} & 0.6 & 99.4 & 99.4 & 99.4 & 94.0 & 2.5 & 68.8 \\ \midrule
        \multirow{3}{*}{FOR-InstanceV2} & Training & 10.3{\text{k}} & 0.1 & 95.1 & 94.9 & 95.0 & 85.5 & 3.1 & 16.3 \\
        & Validation & 6.0{\text{k}} & 0.1 & 98.4 & 98.2 & 98.3 & 91.1 & 3.3 & 19.5 \\
        & Test & 4.3{\text{k}} & 0.2 & 98.4 & 98.2 & 98.3 & 92.3 & 4.0 & 30.5 \\ \midrule
        \multirow{3}{*}{ALS-XL-20-pts} & Training & 1.8{\text{k}} & 8.5 & 99.0 & 98.9 & 99.0 & 93.0 & 28.4 & 95.0 \\
        & Validation & 1.3{\text{k}} & 7.0 & 97.9 & 97.9 & 97.9 & 91.3 & 27.9 & 95.7 \\
        & Test & 1.4{\text{k}} & 8.0 & 98.8 & 98.8 & 98.8 & 92.3 & 28.8 & 95.1 \\
        \bottomrule
    \end{tabular}}
    \label{table:superpoint_generator_performance}
\end{table*}

The performance of the superpoint generator on FGI-EMIT, FOR-InstanceV2 and ALS-XL-20-pts is summarized in \Cref{table:superpoint_generator_performance}. Visual examples of the generated superpoints alongside the ground truth instance annotations are presented in \Cref{figure:superpoint_results}.

Overall, the Oracle metrics demonstrate strong performance, particularly for tree detection as measured by OPrecision, ORecall and OF$_1$-score. For FGI-EMIT and ALS-XL-20-pts, OF$_1$ values range between 97.9--99.8\%, indicating that nearly all trees across both datasets can be detected using superpoint representations when assigned based on point-wise majority voting. Although the OF$_1$ drops to 95.0\% for the FOR-InstanceV2 training set, the accuracy remains high when compared to the best reported FOR-InstanceV2 test set $\text{F}_1$ score to date at $85.5\%$ \citep{yue2026toward}.

Oracle Coverage scores exhibit slightly lower accuracy, ranging from 85.5\% to 94.7\%. This suggests that while the superpoints generally preserve the majority of tree structures, they do not strictly adhere to exact crown boundaries. Moreover, the OCov values appear to be positively correlated with the remaining Oracle metrics, such that lower OCov values typically align with lower overall accuracy. Nevertheless, similar to the Oracle detection metrics, the OCov values remain substantially above Coverages achieved in previous studies, which have typically been 15--30 percentage points (pp) lower \citep{xiang2025forestformer3d,ruoppa2026benchmarking}.

\begin{table*}[!t]
    \centering
    \caption{Comparison of SCATR-ITS and baseline methods on the FGI-EMIT and FOR-InstanceV2 test sets. Best performance for each dataset is in \textbf{bold}, and second-best is \underline{underlined}.} \bigskip
    \scriptsize{
    \begin{tabular}{ll*{5}{S[table-format=2.1]}*{3}{>{\raggedleft\arraybackslash}p{0.5cm}}}
        \toprule
        \multirow{2}{*}{\textbf{Dataset}} & \multirow{2}{*}{\textbf{Model}} & \multicolumn{1}{c}{\multirow{2}{*}{\textbf{Precision (\%)}}} & \multicolumn{1}{c}{\multirow{2}{*}{\textbf{Recall (\%)}}} & \multicolumn{1}{c}{\multirow{2}{*}{\textbf{F$_{\mathbf{1}}$ (\%)}}} & \multicolumn{1}{c}{\multirow{2}{*}{\textbf{Cov (\%)}}} & \multicolumn{1}{c}{\multirow{2}{*}{\textbf{AP$_{\mathbf{50}}$ (\%)}}} & \multicolumn{3}{c}{\textbf{Average time (s/plot)}} \\
        \cmidrule(lr){8-10}
        &&&&&&& \textbf{Tot.} & \textbf{Pre.} & \textbf{Seg.} \\ \midrule \midrule
        \multirow{5}{*}{FGI-EMIT} & YOLOv12 \citep{tian2025yolov12} & 78.3 & 43.0 & 55.5 & 37.9 & 40.4 & \underline{13} & 10 & \textbf{3} \\
        & SegmentAnyTree \citep{wielgosz2024segmentanytree} & 68.1 & 61.8 & \underline{64.8} & 59.6 & 49.7 & 242 & \underline{3} & 240 \\
        & TreeLearn \citep{henrich2024treelearn} & 63.8 & 60.3 & 62.0 & 58.3 & 33.4 & 100 & 40 & 60 \\
        & ForestFormer3D \citep{xiang2025forestformer3d} & \textbf{81.9} & \underline{66.3} & \textbf{73.3} & \underline{62.8} & \underline{63.1} & 68 & \textbf{1} & 67 \\
        & \textbf{SCATR-ITS (Ours)} & \underline{79.2} & \textbf{68.3} & \textbf{73.3} & \textbf{64.0} & \textbf{65.3} & \textbf{8} & 4 & \underline{4} \\ \midrule
        \multirow{5}{*}{FOR-InstanceV2} & YOLOv12 \citep{tian2025yolov12} & 78.2 & 33.7 & 47.1 & 31.2 & 32.4 & \underline{9} & 8 & \textbf{1} \\
        & SegmentAnyTree \citep{wielgosz2024segmentanytree} & 78.9 & 67.2 & 72.6 & 64.4 & 58.8 & 193 & \textbf{1} & 192 \\
        & TreeLearn \citep{henrich2024treelearn} & 82.0 & 63.3 & 71.4 & 61.6 & 49.8 & 45 & 14 & 31 \\
        & ForestFormer3D \citep{xiang2025forestformer3d} & \textbf{92.3} & \textbf{77.2} & \textbf{84.1} & \textbf{72.3} & \textbf{76.0} & 32 & \textbf{1} & 31 \\
        & \textbf{SCATR-ITS (Ours)} & \underline{91.4} & \underline{76.3} & \underline{83.2} & \underline{70.6} & \underline{73.8} & \textbf{7} & \underline{5} & \underline{2} \\
        \bottomrule
    \end{tabular}}
    \label{table:performance_comparison_fgi_emit_fiv2}
\end{table*}

\begin{table*}[!t]
    \centering
    \caption{Comparison of crown-category-level performance for SCATR-ITS and baseline methods on the FGI-EMIT and FOR-InstanceV2 test sets. Best performance for each dataset is in \textbf{bold}, and second-best is \underline{underlined}.} \bigskip
    \scriptsize{
    \begin{tabular}{ll*{4}{S[table-format=2.1]}}
        \toprule
        \textbf{Dataset} & \textbf{Model} & \textbf{Recall$_\text{A}$ (\%)} & \textbf{Recall$_\text{B}$ (\%)} & \textbf{Recall$_\text{C}$ (\%)} & \textbf{Recall$_\text{D}$ (\%)} \\ \midrule \midrule
        \multirow{5}{*}{FGI-EMIT} & YOLOv12 \citep{tian2025yolov12} & 80.9 & 34.2 & 7.0 & 0.0 \\
        & SegmentAnyTree \citep{wielgosz2024segmentanytree} & 90.7 & 38.4 & 42.2 & \underline{32.8} \\
        & TreeLearn \citep{henrich2024treelearn} & 84.8 & \textbf{65.8} & 36.7 & 19.0 \\
        & ForestFormer3D \citep{xiang2025forestformer3d} & \underline{94.1} & \underline{56.2} & \underline{43.0} & \underline{32.8} \\
        & \textbf{SCATR-ITS (Ours)} & \textbf{94.6} & \underline{56.2} & \textbf{47.7} & \textbf{36.2} \\ \midrule
        \multirow{5}{*}{FOR-InstanceV2} & YOLOv12 \citep{tian2025yolov12} & 76.2 & 47.9 & 15.6 & 0.3 \\
        & SegmentAnyTree \citep{wielgosz2024segmentanytree} & \textbf{96.0} & 73.7 & 63.8 & 35.8 \\
        & TreeLearn \citep{henrich2024treelearn} & 85.0 & 76.8 & 56.7 & 35.0 \\
        & ForestFormer3D \citep{xiang2025forestformer3d} & \underline{93.6} & \underline{81.8} & \textbf{75.6} & \textbf{57.6} \\
        & \textbf{SCATR-ITS (Ours)} & \textbf{96.0} & \textbf{84.2} & \underline{73.5} & \underline{51.3} \\ \bottomrule
    \end{tabular}}
    \label{table:performance_comparison_crown_categories}
\end{table*}

Qualitative visual inspection reveals that Oracle performance is lowest in deciduous and mixed forests, where crown boundaries are often ambiguous, as well as multi-layered, high-density forests with a high concentration of small understory trees. Moreover, dense TLS data presents additional challenges for the superpoint generator. However, such forest environments are inherently difficult to segment accurately, as established in prior studies \citep{xiang2024automated,ruoppa2026benchmarking,wielgosz2026segmentanytreev2}. These observations also explain why the lowest Oracle accuracy occurs on the FOR-InstanceV2 training set, which contains a large proportion of TLS plots depicting deciduous forests from the BlueCat region.

Although superpoint boundary metrics appear low, they align directly with our observations regarding OCov. This is illustrated particularly well by the FOR-InstanceV2 training set, where BR reaches only 16.3\% despite high Oracle detection metrics. This discrepancy suggests that the single-edge relaxation in the point cloud graph representation used for computing BR and BP is overly restrictive for high-density point clouds, where superpoint boundaries can shift by several graph nodes while still preserving the primary tree structure. Manual inspection confirms BR is especially low for dense TLS scans, where the single-edge tolerance provides minimal spatial leeway. Conversely, BP and BR are substantially higher on the sparse ALS-XL-20-pts dataset, where the same graph relaxation offers a far more generous spatial margin.

Based on the average superpoint count and reduction ratio, the generator yields particularly compact representations for the denser FGI-EMIT and FOR-InstanceV2 datasets, where the RR ranges between 0.1\% and 0.6\%. That is, the superpoints compress total point counts by up to four orders of magnitude. Even on the sparser ALS-XL-20-pts, the RR is 7\%--8.5\%, demonstrating a significant reduction in point count. Therefore, while superpoints are most effective at compressing high-density data, they remain computationally advantageous at densities as low as 20 points/m$^2$.

All performance metrics indicate that superpoints constructed using the proposed generator are both accurate and highly effective at compressing the point cloud. Consequently, they fulfill the key structural requirements for accelerating downstream ITS model inference.

\subsection{Performance of SCATR-ITS} \label{section:scatr_its_performance}

\begin{figure*}[!t]
\centering
\includegraphics[width=\textwidth]{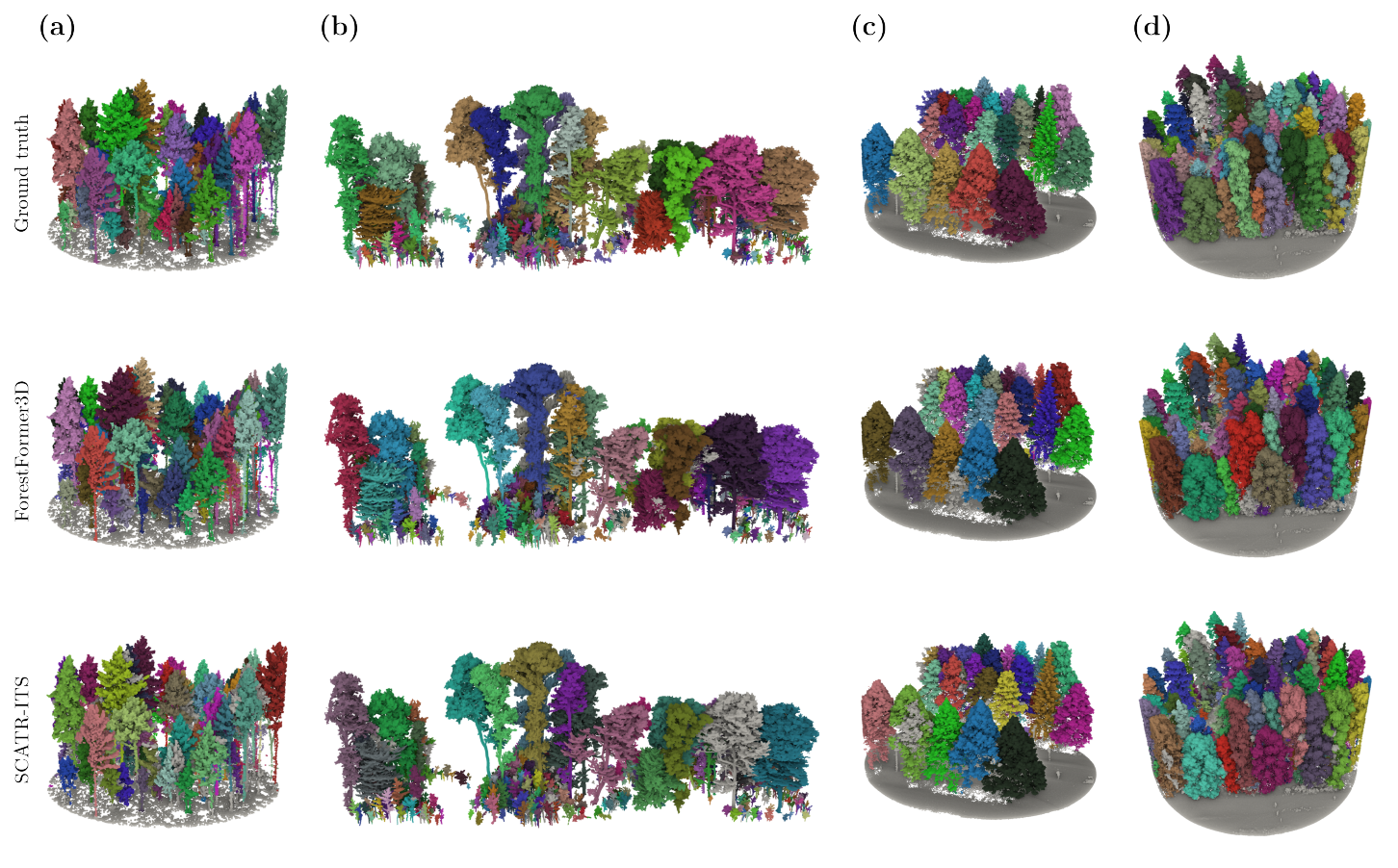}
\caption{Comparison of segmentation results produced by ForestFormer3D and SCATR-ITS on the FOR-InstanceV2 and FGI-EMIT test sets. Each predicted tree instance is shown in a distinct color, while non-tree points are shown in gray. (a) Plot 3 from the NIBIO2 region in FOR-InstanceV2. (b) BlueCat region plot from FOR-InstanceV2. (c) Plot 1008 from FGI-EMIT. (d) Plot 1018 from FGI-EMIT.}
\label{figure:fiv2_fgi_emit_results}
\end{figure*}

The segmentation accuracy of SCATR-ITS and the four baselines on the FGI-EMIT and FOR-InstanceV2 datasets is presented in \Cref{table:performance_comparison_fgi_emit_fiv2} with the corresponding crown-category-level performance shown in \Cref{table:performance_comparison_crown_categories}. Qualitative results produced by SCATR-ITS on the same datasets are visualized in \Cref{figure:fiv2_fgi_emit_results}.

The results demonstrate that SCATR-ITS achieves segmentation accuracy comparable to the state-of-the-art. On FGI-EMIT, our model yields an F$_1$-score of 73.3\%, matching ForestFormer3D exactly. On FOR-InstanceV2, the F$_1$ of SCATR-ITS is 83.2\%, only 0.9 pp below ForestFormer3D at 84.1\%. In terms of Precision, ForestFormer3D outperforms SCATR-ITS by a small margin on both datasets, indicating that our model has a slightly higher false positive rate. By contrast, SCATR-ITS achieves the higher Recall on FGI-EMIT, but is slightly outperformed by ForestFormer3D on FOR-InstanceV2. Similar trends are observed for Cov and AP. Overall, the differences in accuracy metrics between the two models across both datasets remain consistently within 3 percentage points. These patterns hold at the crown category level, although differences can be more pronounced. For example, SCATR-ITS outperforms ForestFormer3D by 4.7 pp in category C on FGI-EMIT, whereas ForestFormer3D yields a 6.3 pp higher Recall in category D on FOR-InstanceV2.

When compared to the other ITS baselines, SCATR-ITS is substantially more accurate across all metrics, outperforming the second best model, SegmentAnyTree, by 8.5 pp on FGI-EMIT and 10.6 pp on FOR-InstanceV2 in terms of F$_1$-score. This stark performance difference remains at the crown category level, with two notable exceptions. First, on FOR-InstanceV2, SegmentAnyTree achieves the same Recall as SCATR-ITS in the dominant category A at 96.0\%. Second, on FGI-EMIT, TreeLearn yields a significantly higher Recall than both SCATR-ITS and ForestFormer3D in category B, which consists of closely grouped trees of similar size. This phenomenon was previously observed and analyzed in detail by \citet{ruoppa2026benchmarking}.

The region-level accuracy metrics of SCATR-ITS and the baseline methods on FOR-InstanceV2 are provided in \Cref{appendix:performance_comparison_fiv2_region_level}. Since ALS-XL-20-pts primarily served as a training dataset for experiments on Finnish national laser scanning data, rather than a full benchmark, the corresponding performance metrics for SCATR-ITS, ForestFormer3D, and watershed are provided in \Cref{appendix:als_xl_20pts}. For a discussion on discrepancies in the segmentation accuracy of our ITS baselines and the corresponding performance reported in previous studies, refer to \Cref{appendix:performance_discrepancies}.

In addition to state-of-the-art segmentation accuracy, SCATR-ITS achieves the fastest average plot-wise processing time. When compared to the second-fastest 3D model, ForestFormer3D, SCATR-ITS is $8.5\times$ faster on FGI-EMIT and $4.6\times$ faster on FOR-InstanceV2. This indicates that while our proposed modifications to the underlying ForestFormer3D framework are highly effective at accelerating inference, the computational gains depend on data density and spatial extent. Similar to the accuracy comparisons, SCATR-ITS outperforms deep-clustering-based 3D baselines by a wide margin. For example, on FGI-EMIT, SCATR-ITS is $12.5\times$ than TreeLearn and $30\times$ faster than SegmentAnyTree.

The 2D YOLOv12 is the fastest model in terms of pure segmentation speed, outperforming SCATR-ITS on both datasets. By contrast, the preprocessing stage for YOLOv12, which includes both ground detection and height normalization, is quite computationally demanding, ultimately leading to a slower overall runtime than SCATR-ITS. Moreover, despite its rapid inference speed, YOLOv12 yields the lowest segmentation accuracy among all tested models. This poor performance stems from its 2D approach to ITS, which cannot capture 3D crown structure or segment suppressed understory trees. Consequently, SCATR-ITS exceeds the F$_1$-score of YOLOv12 by 17.8 pp on FGI-EMIT and 36.1 pp on FOR-InstanceV2. 

Phase-level runtimes indicate that the speed of SCATR-ITS is bottlenecked by preprocessing, which accounts for 50\% of its total runtime on FGI-EMIT and 70\% on FOR-InstanceV2. The issue is particularly pronounced for denser data modalities, such as TLS and MLS, where operating directly on raw, non-superpoint data becomes computationally expensive due to operations such as neighborhood graph construction.

\begin{figure*}[!t]
\centering
\includegraphics[width=\textwidth]{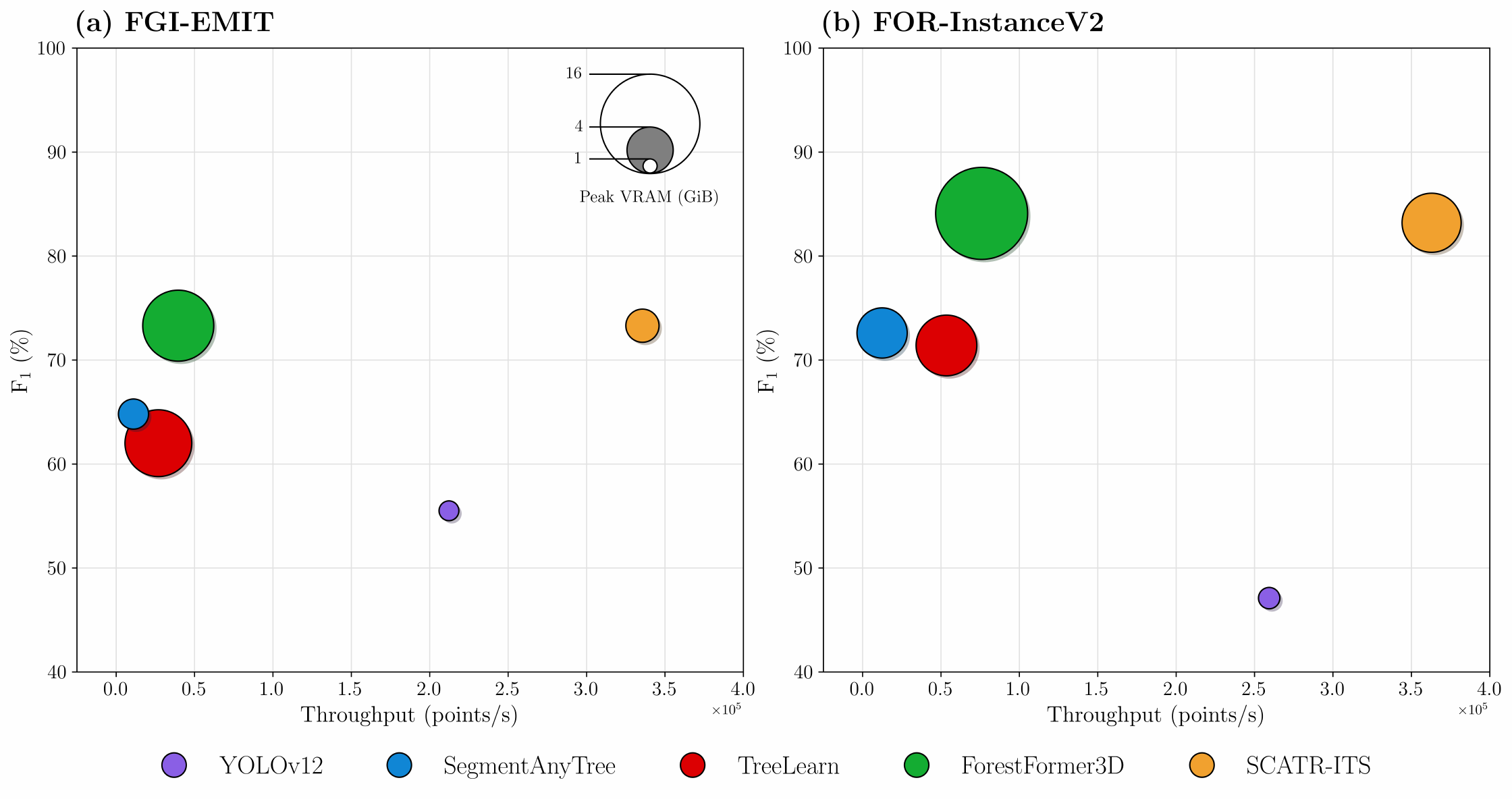}
\caption{Comparison of inference speed, segmentation accuracy, and peak GPU VRAM usage during inference for SCATR-ITS and performance baselines on the FGI-EMIT and FOR-InstanceV2 datasets. All data preprocessing is included in the throughput. SCATR-ITS achieves accuracy on par with the state of the art while providing significantly faster inference and lower VRAM usage.}
\label{figure:throughput_vram_f1_bubblechart}
\end{figure*}

Additional results describing the computational costs of SCATR-ITS and the four baseline methods are presented in \Cref{table:speed_vram_comparison}. A visual comparison combining segmentation accuracy, throughput, and VRAM footprint is provided in \Cref{figure:throughput_vram_f1_bubblechart}. We note that reported peak inference VRAM statistics exclude the preprocessing stage. This is primarily relevant for SCATR-ITS, as its superpoint generator runs on the GPU. In fact, given sufficiently dense input data, the VRAM consumption of the superpoint generator can surpass that of the segmentation stage. Reducing the input cylinder radius mitigates the issue of increased memory requirements for high point densities, albeit at the cost of preprocessing speed. Because the training procedure is inherently stochastic, reported peak training VRAM values should be interpreted as estimates.

\begin{table*}[!t]
    \centering
    \caption{Comparison of test set inference throughput and peak VRAM usage during both inference and training on the FGI-EMIT and FOR-InstanceV2 datasets. The listed peak VRAM usage statistics do not include the preprocessing stage. OOM indicates that the model exceeded the 48 GiB GPU memory available on the evaluation hardware. Best performance for each dataset is in \textbf{bold}, and second-best is \underline{underlined}.} \bigskip
    \scriptsize{
    \begin{tabular}{llS[table-format=2.1]*{2}{S[table-format=1.1]}}
        \toprule
        \textbf{Dataset} & \textbf{Model} & \parbox{1.7cm}{\centering\textbf{Throughput (points/s)}} & \parbox{1.8cm}{\centering\textbf{Inference VRAM (GiB)}} & \parbox{1.8cm}{\centering\textbf{Training VRAM (GiB)}} \\ \midrule \midrule
        \multirow{5}{*}{FGI-EMIT} & YOLOv12 \citep{tian2025yolov12} & \underline{212.3{\text{k}}} & \textbf{1.3} & 8.9 \\
        & SegmentAnyTree \citep{wielgosz2024segmentanytree} & 11.1{\text{k}} & \underline{2.1} & \textbf{6.7} \\
        & TreeLearn \citep{henrich2024treelearn} & 27.0{\text{k}} & 7.6 & 28.2 \\
        & ForestFormer3D \citep{xiang2025forestformer3d} & 39.6{\text{k}} & 8.5 & 41.3  \\
        & \textbf{SCATR-ITS (Ours)} & \bfseries 335.7{\text{k}} & 2.4 & \underline{7.4} \\ \midrule
        \multirow{5}{*}{FOR-InstanceV2} & YOLOv12 \citep{tian2025yolov12} & \underline{259.3{\text{k}}} & \textbf{1.4} & \textbf{8.2} \\
        & SegmentAnyTree \citep{wielgosz2024segmentanytree} & 12.5{\text{k}} & \underline{4.6} & \underline{8.6} \\
        & TreeLearn \citep{henrich2024treelearn} & 53.5{\text{k}} & 6.4 & 33.6 \\
        & ForestFormer3D \citep{xiang2025forestformer3d} & 76.0{\text{k}} & 13.8 & \text{OOM} \\
        & \textbf{SCATR-ITS (Ours)} & \textbf{362.9{\text{k}}} & 6.0 & 16.8 \\ \bottomrule
    \end{tabular}}
    \label{table:speed_vram_comparison}
\end{table*}

Model throughput values corroborate the average plot-level runtime results and offer a normalized measure of segmentation speed that allows direct comparisons between datasets. Notably, we observe that the throughput is consistently higher on FOR-InstanceV2 across all models, albeit marginally. This likely stems from a combination of the larger input radius and higher variance in point density on FOR-InstanceV2.

\begin{table*}[!t]
    \centering
    \caption{Ablation study of the proposed architectural components on the FGI-EMIT test set. SPD, SPP, SAP, DQ, and MaxS denote superpoint-level decoder, superpoint-level postprocessing, size-aware postprocessing, dynamic query count, and maximal input stride, respectively. Best performance is in \textbf{bold}, and second-best is \underline{underlined}.} \bigskip
    \scriptsize{
    \begin{tabular}{ccccc*{5}{S[table-format=2.1]}*{1}{S[table-format=1.1]}*{3}{>{\raggedleft\arraybackslash}p{0.5cm}}}
        \toprule
        \multirow{2}{*}{\textbf{SPD}} & \multirow{2}{*}{\textbf{SPP}} & \multirow{2}{*}{\textbf{SAP}} & \multirow{2}{*}{\textbf{DQ}} & \multirow{2}{*}{\textbf{MaxS}} & \multicolumn{1}{c}{\multirow{2}{*}{\textbf{Precision (\%)}}} & \multicolumn{1}{c}{\multirow{2}{*}{\textbf{Recall (\%)}}} & \multicolumn{1}{c}{\multirow{2}{*}{\textbf{F$_{\mathbf{1}}$ (\%)}}} & \multicolumn{1}{c}{\multirow{2}{*}{\textbf{Cov (\%)}}} & \multicolumn{1}{c}{\multirow{2}{*}{\textbf{AP$_{\mathbf{50}}$ (\%)}}} & \multicolumn{1}{c}{\multirow{2}{*}{\textbf{VRAM (GiB)}}} & \multicolumn{3}{c}{\textbf{Average time (s/plot)}} \\
        \cmidrule(lr){12-14}
        &&&&&&&&&&& \textbf{Tot.} & \textbf{Pre.} & \textbf{Seg.} \\ \midrule \midrule
        \multicolumn{5}{l}{ForestFormer3D \citep{xiang2025forestformer3d}} & \underline{81.9} & 66.3 & \underline{73.3} & 62.8 & 63.1 & 8.5 & 68 & \textbf{1} & 67 \\ \midrule
        &&&& \checkmark & \textbf{82.9} & 65.0 & 72.9 & 61.3 & 62.6 & 8.5 & 17 & \textbf{1} & 16  \\
        \checkmark &&&&& 78.7 & 67.2 & 72.5 & 64.4 & 64.0 & 4.2 & 16 & \underline{4} & 12  \\
        \checkmark & \checkmark &&&& 79.6 & 64.8 & 71.4 & 62.5 & 62.2 & \bfseries{\tablenum{2.3}} & \underline{15} & \underline{4} & \underline{11}  \\
        \checkmark & \checkmark & \checkmark &&& 79.3 & 67.8 & 73.1 & \underline{64.5} & 64.6 & \bfseries{\tablenum{2.3}} & \underline{15} & \underline{4} & \underline{11}  \\
        \checkmark & \checkmark & \checkmark & \checkmark && 78.6 & \textbf{70.4} & \textbf{74.3} & \textbf{65.7} & \textbf{66.2} & \underline{2.4} & \underline{15} & \underline{4} & \underline{11} \\
        \checkmark & \checkmark & \checkmark & \checkmark & \checkmark & 79.2 & \underline{68.3} & \underline{73.3} & 64.0 & \underline{65.3} & \underline{2.4} & \textbf{8} & \underline{4} & \textbf{4} \\
        \bottomrule
    \end{tabular}}
    \label{table:ablation_fgi_emit}
\end{table*}

\begin{table*}[!t]
    \centering
    \caption{Crown-category-level metrics for the ablation study on the FGI-EMIT test set. SPD, SPP, SAP, DQ, and MaxS denote superpoint-level decoder, superpoint-level postprocessing, size-aware postprocessing, dynamic query count, and maximal input stride, respectively. Best performance is in \textbf{bold}, and second-best is \underline{underlined}.} \bigskip
    \scriptsize{
    \begin{tabular}{ccccc*{4}{S[table-format=2.1]}}
        \toprule
        \textbf{SPD} & \textbf{SPP} & \textbf{SAP} & \textbf{DQ} & \textbf{MaxS} & \textbf{Recall$_\text{A}$ (\%)} & \textbf{Recall$_\text{B}$ (\%)} & \textbf{Recall$_\text{C}$ (\%)} & \textbf{Recall$_\text{D}$ (\%)} \\ \midrule \midrule
        \multicolumn{5}{l}{ForestFormer3D \citep{xiang2025forestformer3d}} & \underline{94.1} & \underline{56.2} & 43.0 & 32.8 \\ \midrule
        &&&& \checkmark  & 92.6 & 54.8 & 41.4 & 32.8 \\
        \checkmark &&&&& 93.1 & 53.4 & \underline{48.4} & 34.5 \\
        \checkmark & \checkmark &&&& 92.2 & 49.3 & 45.3 & 31.0 \\
        \checkmark & \checkmark & \checkmark &&& 93.6 & 54.8 & \underline{48.4} & \underline{36.2} \\
        \checkmark & \checkmark & \checkmark & \checkmark && \underline{94.1} & \textbf{57.5} & \textbf{53.1} & \textbf{41.4} \\
        \checkmark & \checkmark & \checkmark & \checkmark & \checkmark &  \textbf{94.6} & \underline{56.2} & 47.7 & \underline{36.2} \\
        \bottomrule
    \end{tabular}}
    \label{table:ablation_crown_ctg_fgi_emit}
\end{table*}

Peak inference VRAM statistics demonstrate that SCATR-ITS successfully reduces memory consumption compared to the ForestFormer3D baseline while maintaining comparable accuracy. Specifically, the peak VRAM footprint of SCATR-ITS is $3.5\times$ lower on FGI-EMIT and $2.3\times$ lower on FOR-InstanceV2. Conversely, while both SegmentAnyTree and YOLOv12 require less GPU memory than our model, they are substantially less accurate, and SegmentAnyTree is also by far the slowest model tested. Overall, these results highlight a generally favorable trade-off for SCATR-ITS, which offers a combination of high accuracy and moderate VRAM footprint during inference.

Because the inference VRAM consumption of ForestFormer3D is not particularly high to begin with, peak training VRAM is potentially more critical for practical deployment, especially since training memory requirements are generally substantially higher across all models. In this regard, SCATR-ITS is highly effective at reducing computational overhead. On FGI-EMIT, the peak training VRAM is only 7.4 GiB, $5.6\times$ lower than the 41.3 GiB required by ForestFormer3D. Similarly, on FOR-InstanceV2 SCATR-ITS caps at 16.8 GiB, whereas ForestFormer3D cannot be trained on our hardware as its peak VRAM exceeds the available 48 GiB. This demonstrates that the reduced VRAM footprint of SCATR-ITS enables training models with state-of-the-art performance on previously inadequate hardware. Although more lightweight baselines like SegmentAnyTree and YOLOv12 require less training memory, SCATR-ITS delivers a superior accuracy-to-resource ratio.

\subsection{Ablation study of the proposed modifications}

To assess the impact of our proposed modifications, specifically their ability to accelerate model inference and reduce VRAM footprint while retaining state-of-the-art ITS accuracy, we conducted an ablation study on the FGI-EMIT and FOR-InstanceV2 datasets. Using ForestFormer3D as the baseline, we examined five architectural modifications: superpoint-level instance decoding (SPD), superpoint-level postprocessing (SPP), size-aware postprocessing (SAP), the dynamic query count module (DQ), and the maximal input stride (MaxS). The results are presented in \Cref{table:ablation_fgi_emit,table:ablation_fiv2}, with the corresponding crown-category-level metrics listed in \Cref{table:ablation_crown_ctg_fgi_emit,table:ablation_crown_ctg_fiv2}.

Overall, the ablation study indicates that the proposed improvements achieve their intended purpose. Both the maximal input stride and superpoint-level decoding substantially accelerate inference, with the latter also reducing peak VRAM usage. However, due to the decreased redundancy in instance predictions and coarser input representations, both ablated models are less accurate than the ForestFormer3D baseline. Although superpoint-level postprocessing provides a modest speedup during segmentation, its impact on peak VRAM usage is much more pronounced, reducing memory footprint by 1.9 GiB on FGI-EMIT and 6.3 GiB on FOR-InstanceV2. The addition of size-awareness completely offsets the accuracy loss introduced by SPP, an effect particularly significant on FOR-InstanceV2 where it recovers a $6.4$ pp drop in F$_1$. Integrating the dynamic query count module with the default fixed-ratio input stride yields the highest accuracy on both datasets, outperforming ForestFormer3D by 1.0 and 0.5 pp in F$_1$ on FGI-EMIT and FOR-InstanceV2, respectively. DQ enables SCATR-ITS to effectively compensate for the accuracy trade-offs of MaxS and SPD, resulting in a model that is both computationally efficient and accurate. The consistent results across two distinct datasets further validate the efficacy of our model architecture.

Crucially, because SPP, SAP, and MaxS exclusively affect the inference phase, only two model variants required training for the ablation study: one featuring superpoint-level decoding, and a second additionally incorporating the dynamic query count module. All training hyperparameters for the ablated models remained identical to SCATR-ITS. Consistent with the full model, the instance score threshold was set to 0.3 for all superpoint-based configurations and the default query count of $K=300$ was used for models without DQ. Models with no SPP casted results back to the full point cloud immediately after the transformer decoder. The number of training epochs was determined as described in \Cref{appendix:scatr_its_training_details}. Variants without DQ were trained for 8,000 epochs on FGI-EMIT and 3,800 epochs on FOR-InstanceV2. Compared to the full model, this corresponds to a longer training duration on FOR-InstanceV2 and a shorter one on FGI-EMIT. This discrepancy likely arises because SCATR-ITS tends to utilize more queries on FGI-EMIT than the baseline, whereas its query count often decreases on FOR-InstanceV2. Consequently, the addition of DQ may simplify the learning objective on several stands of FOR-InstanceV2, accelerating convergence. This hypothesis is further supported by the peak VRAM footprint of SCATR-ITS, which increases on FGI-EMIT with DQ, but decreases on FOR-InstanceV2.

\begin{table*}[!t]
    \centering
    \caption{Ablation study of the proposed architectural components on the FOR-InstanceV2 test set. SPD, SPP, SAP, DQ, and MaxS denote superpoint-level decoder, superpoint-level postprocessing, size-aware postprocessing, dynamic query count, and maximal input stride, respectively. Best performance is in \textbf{bold}, and second-best is \underline{underlined}.} \bigskip
    \scriptsize{
    \begin{tabular}{ccccc*{5}{S[table-format=2.1]}*{1}{S[table-format=1.1]}*{3}{>{\raggedleft\arraybackslash}p{0.5cm}}}
        \toprule
        \multirow{2}{*}{\textbf{SPD}} & \multirow{2}{*}{\textbf{SPP}} & \multirow{2}{*}{\textbf{SAP}} & \multirow{2}{*}{\textbf{DQ}} & \multirow{2}{*}{\textbf{MaxS}} & \multicolumn{1}{c}{\multirow{2}{*}{\textbf{Precision (\%)}}} & \multicolumn{1}{c}{\multirow{2}{*}{\textbf{Recall (\%)}}} & \multicolumn{1}{c}{\multirow{2}{*}{\textbf{F$_{\mathbf{1}}$ (\%)}}} & \multicolumn{1}{c}{\multirow{2}{*}{\textbf{Cov (\%)}}} & \multicolumn{1}{c}{\multirow{2}{*}{\textbf{AP$_{\mathbf{50}}$ (\%)}}} & \multicolumn{1}{c}{\multirow{2}{*}{\textbf{VRAM (GiB)}}} & \multicolumn{3}{c}{\textbf{Average time (s/plot)}} \\
        \cmidrule(lr){12-14}
        &&&&&&&&&&& \textbf{Tot.} & \textbf{Pre.} & \textbf{Seg.}  \\ \midrule \midrule
        \multicolumn{5}{l}{ForestFormer3D \citep{xiang2025forestformer3d}} & \underline{92.3} & 77.2 & \underline{84.1} & 72.3 & \underline{76.0} & 13.8 & 32 & \textbf{1} & 31  \\ \midrule
        &&&& \checkmark & \textbf{92.6} & 75.7 & 83.3 & 70.7 & 74.8 & 13.8 & \underline{8} & \textbf{1} & 7  \\
        \checkmark &&&&& 87.7 & 79.4 & 83.3 & \underline{73.6} & \underline{76.0} & 12.4 & 10 & \underline{5} & 5  \\
        \checkmark & \checkmark &&&& 90.9 & 66.6 & 76.9 & 62.0 & 65.0 & \underline{6.1} & 10 & \underline{5} & \underline{4}  \\
        \checkmark &\checkmark & \checkmark &&& 86.7 & \underline{79.5} & 82.9 & \textbf{73.7} & 75.9 & \underline{6.1} & 10 & \underline{5} & \underline{4}  \\
        \checkmark &\checkmark & \checkmark & \checkmark && 90.2 & \textbf{79.6} & \textbf{84.6} & 73.5 & \textbf{76.6} & \textbf{6.0} & 9 & \underline{5} & \underline{4}  \\
        \checkmark &\checkmark & \checkmark & \checkmark & \checkmark & 91.4 & 76.3 & 83.2 & 70.6 & 73.8 & \textbf{6.0} & \textbf{7} & \underline{5} & \textbf{2}  \\
        \bottomrule
    \end{tabular}}
    \label{table:ablation_fiv2}
\end{table*}

\begin{table*}[!t]
    \centering
    \caption{Crown-category-level metrics for the ablation study on the FOR-InstanceV2 test set. SPD, SPP, SAP, DQ, and MaxS denote superpoint-level decoder, superpoint-level postprocessing, size-aware postprocessing, dynamic query count, and maximal input stride, respectively. Best performance is in \textbf{bold}, and second-best is \underline{underlined}.} \bigskip
    \scriptsize{
    \begin{tabular}{ccccc*{4}{S[table-format=2.1]}}
        \toprule
        \textbf{SPD} & \textbf{SPP} & \textbf{SAP} & \textbf{DQ} & \textbf{MaxS} & \textbf{Recall$_\text{A}$ (\%)} & \textbf{Recall$_\text{B}$ (\%)} & \textbf{Recall$_\text{C}$ (\%)} & \textbf{Recall$_\text{D}$ (\%)} \\ \midrule \midrule
        \multicolumn{5}{l}{ForestFormer3D \citep{xiang2025forestformer3d}} & 93.6 & 81.8 & 75.6 & \textbf{57.6} \\ \midrule
        &&&& \checkmark & 93.3 & 81.2 & 73.5 & 55.0 \\
        \checkmark &&&&&  \underline{96.8} & \underline{86.2} & 77.1 & \underline{57.1} \\
        \checkmark & \checkmark &&&& 90.4 & 78.1 & 62.5 & 35.3 \\
        \checkmark & \checkmark & \checkmark &&& 96.5 & \textbf{86.7} & \underline{77.3} & \underline{57.1} \\
        \checkmark & \checkmark & \checkmark & \checkmark && \textbf{97.1} & 85.8 & \textbf{78.8} & 56.1 \\
        \checkmark & \checkmark & \checkmark & \checkmark & \checkmark &  96.0 & 84.2 & 73.5 & 51.3 \\
        \bottomrule
    \end{tabular}}
    \label{table:ablation_crown_ctg_fiv2}
\end{table*}

\subsection{Generalization to out-of-dataset forest scenes} \label{section:generalization_lautx_wytham}

\begin{figure}[!t]
\centering
\includegraphics[width=0.48\textwidth]{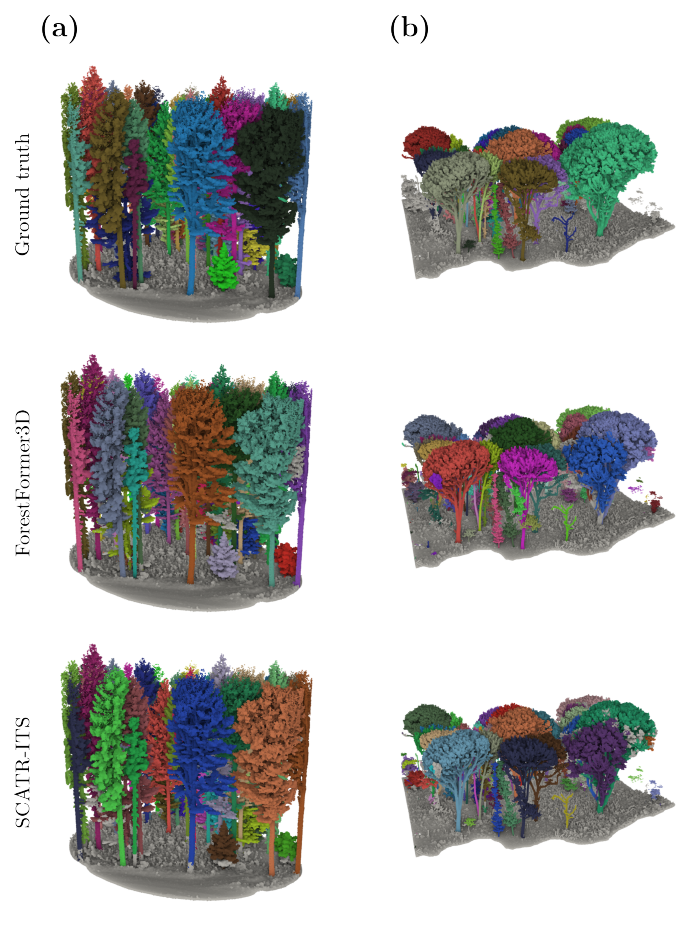}
\caption{Comparison of segmentation results produced by ForestFormer3D and SCATR-ITS on the LAUTx and Wytham Woods datasets. Each predicted tree instance is shown in a distinct color, while non-tree points are shown in gray. (a) Plot 12 from LAUTx. (b) Subsection of the full Wytham Woods plot.}
\label{figure:lautx_wytham_results}
\end{figure}

To assess the cross-dataset generalization of SCATR-ITS, we evaluated models trained on FOR-InstanceV2 directly on the LAUTx and Wytham Woods datasets without fine-tuning. A quantitative comparison against the ITS baselines is presented in \Cref{table:performance_comparison_lautx_wytham}, with visual examples of segmentation results shown in \Cref{figure:lautx_wytham_results}. Note that due to longer plot-level processing times, inference speed is reported in minutes per plot instead of seconds.

The results suggest that SCATR-ITS generalizes effectively to previously unseen data. On LAUTx, it achieves an F$_1$-score of 84\%, just 3 pp below the best-performing baseline, ForestFormer3D. On Wytham Woods, SCATR-ITS attains the highest F$_1$ among all evaluated methods at 62.8\%, outperforming the second-best approach, TreeLearn, by 1.9 pp. Consistent with prior studies \citep{xiang2025forestformer3d}, TreeLearn achieves the highest Recall across both datasets. However, its lower Precision leads to an F$_1$-score below that of SCATR-ITS.

The inference speed of SCATR-ITS also remains substantially higher than that of the other evaluated 3D DL models, although the advantage is less pronounced than on FGI-EMIT and FOR-InstanceV2, largely due to the extremely high point densities of LAUTx and Wytham Woods. Nevertheless, SCATR-ITS is approximately 3--5$\times$ faster than ForestFormer3D and up to $67\times$ faster than SegmentAnyTree. Notably, unlike on FGI-EMIT and FOR-InstanceV2, SCATR-ITS is not the fastest model overall in terms of average plot-level inference speed. Instead, YOLOv12 is generally 1.5--2.5$\times$ faster on our evaluation hardware, demonstrating that at under extremely high point densities, the processing efficiency introduced by superpoints is no longer sufficient to outperform 2D-based approaches. Simultaneously, however, YOLOv12 generalizes poorly to both LAUTx and Wytham Woods, achieving F$_1$-scores of only 28.7\% and 6.6\%, respectively, which renders it unsuitable for MLS and TLS data despite its high inference speed.

Similar to \Cref{section:scatr_its_performance}, see \Cref{appendix:performance_discrepancies} for a discussion on discrepancies between our ITS baseline metrics and those reported for the same models in prior work.

\begin{table*}[!t]
    \centering
    \caption{Comparison of SCATR-ITS and baseline methods on the LAUTx and Wytham Woods datasets using models trained on FOR-InstanceV2. Best performance for each dataset is in \textbf{bold}, and second-best is \underline{underlined}.} \bigskip
    \scriptsize{
    \begin{tabular}{ll*{5}{S[table-format=2.1]}*{3}{S[table-format=1.1,table-column-width=0.7cm]}}
        \toprule
        \multirow{2}{*}{\textbf{Dataset}} & \multirow{2}{*}{\textbf{Model}} & \multicolumn{1}{c}{\multirow{2}{*}{\textbf{Precision (\%)}}} & \multicolumn{1}{c}{\multirow{2}{*}{\textbf{Recall (\%)}}} & \multicolumn{1}{c}{\multirow{2}{*}{\textbf{F$_{\mathbf{1}}$ (\%)}}} & \multicolumn{1}{c}{\multirow{2}{*}{\textbf{Cov (\%)}}} & \multicolumn{1}{c}{\multirow{2}{*}{\textbf{AP$_{\mathbf{50}}$ (\%)}}} & \multicolumn{3}{c}{\textbf{Average time (min/plot)}} \\
        \cmidrule(lr){8-10}
        &&&&&&& \textbf{Tot.} & \textbf{Pre.} & \textbf{Seg.}  \\ \midrule \midrule
        \multirow{5}{*}{LAUTx} & YOLOv12 \citep{tian2025yolov12} & 34.5 & 24.5 & 28.7 & 32.0 & 14.7 & \textbf{0.4} & \textbf{0.2} & \textbf{0.2} \\
        & SegmentAnyTree \citep{wielgosz2024segmentanytree} & 75.7 & 75.3 & 75.5 & 69.2 & 61.8 & 67.8 & \underline{0.3} & 67.5 \\
        & TreeLearn \citep{henrich2024treelearn} & 77.0 & \textbf{89.9} & 82.9 & \textbf{83.8} & 52.5 & 8.0 & 2.6 & 5.4 \\
        & ForestFormer3D \citep{xiang2025forestformer3d} & \textbf{90.7} & \underline{83.7} & \textbf{87.0} & \underline{77.5} & \textbf{81.9} & 5.1 & \textbf{0.2} & 4.9 \\
        & \textbf{SCATR-ITS (Ours)} & \underline{89.5} & 79.2 & \underline{84.0} & 68.7 & \underline{75.4} & \underline{1.0} & 0.5 & \underline{0.5} \\ \midrule
        \multirow{5}{*}{Wytham Woods} & YOLOv12 \citep{tian2025yolov12} & 22.7 & 3.9 & 6.6 & 8.5 & 1.6 & \textbf{6.4} & 2.8 & \underline{3.6} \\
        & SegmentAnyTree \citep{wielgosz2024segmentanytree} & 48.0 & 44.4 & 46.1 & 46.4 & 28.2 & 332.1 & \underline{2.6} & 329.4 \\
        & TreeLearn \citep{henrich2024treelearn} & 56.0 & \textbf{66.7} & \underline{60.9} & \textbf{64.3} & 28.4 & 75.6 & 16.6 & 59.0 \\
        & ForestFormer3D \citep{xiang2025forestformer3d} & \underline{58.4} & 61.5 & 59.9 & 60.9 & \underline{44.6} & 29.0 & \textbf{1.2} & 27.9 \\
        & \textbf{SCATR-ITS (Ours)} & \textbf{62.9} & \underline{62.7} & \textbf{62.8} & \underline{61.1} & \textbf{51.3} & \underline{9.7} & 7.2 & \textbf{2.6} \\
        \bottomrule
    \end{tabular}}
    \label{table:performance_comparison_lautx_wytham}
\end{table*}

\subsection{Employing SCATR-ITS for large-area individual tree segmentation} \label{section:large_area_performance}

\begin{figure*}[!t]
\centering
\includegraphics[width=\textwidth]{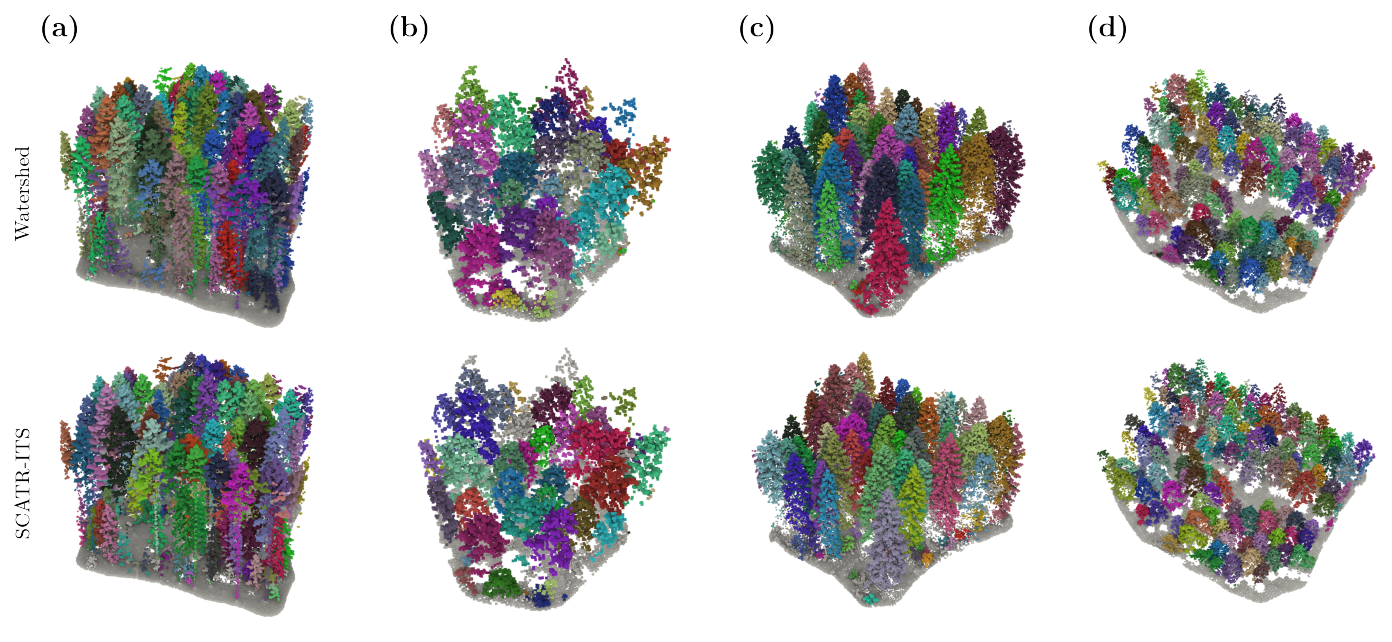}
\caption{Comparison of segmentation results produced by watershed and SCATR-ITS on the Finnish national laser scanning data (20 points/m$^2$). Each predicted tree instance is shown in a distinct color, while non-tree points are shown in gray. (a) Pine-dominated, multi-layered forest stand. (b) Deciduous-dominated stand consisting primarily of birch. (c) Spruce-dominated stand with several aspen scattered throughout. (d) Sparse mixed forest stand.}
\label{figure:padasjoki_20_pts_results}
\end{figure*}

\begin{figure}[!t]
\centering
\includegraphics[width=0.48\textwidth]{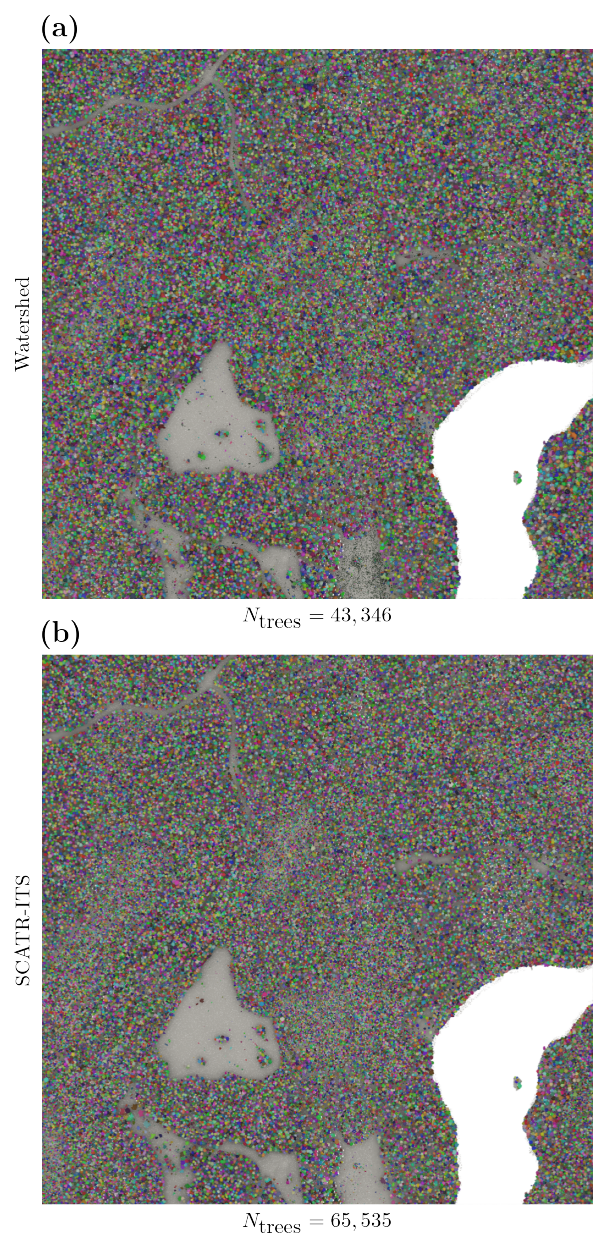}
\caption{Comparison of segmentation results produced by watershed and SCATR-ITS on the Finnish national laser scanning data (20 points/m$^2$) for a full 1 km $\times$ 1 km tile, shown from a bird's-eye view. Each predicted tree instance is shown in a distinct color, while non-tree points are shown in gray. $N_{\text{trees}}$ denotes the total number of individual tree segments identified by each method. (a) Segmentation results from watershed. (b) Segmentation results from SCATR-ITS.}
\label{figure:padasjoki_20_pts_bev}
\end{figure}

To demonstrate that SCATR-ITS is applicable to large-area individual tree segmentation and that its inference supports scaling processing to nationwide datasets, we applied the model trained on ALS-XL-20-pts to a 142 km$^2$ region of Finnish national laser scanning data (20 points/m$^2$). The watershed-based ITS algorithm of \citet{yu2011predicting} served as a performance baseline. Detection accuracy was evaluated on sub-regions containing field-collected reference data. In the absence of 3D ground truth, predicted segments were matched to field measurements using the method of \citet{yu2006change}. Quantitative results are presented in \Cref{table:performance_comparison_padasjoki_20_pts}, with visual examples shown in \Cref{figure:padasjoki_20_pts_results,figure:padasjoki_20_pts_bev}. All metrics were computed using distance thresholds $d_{\text{th}}=3$ m and $d_{\text{th}}=5$ m.

SCATR-ITS identifies a total of 8,201,435 individual trees, which is $\sim\!1.5\times$ more than the watershed baseline (5,565,347 trees). Furthermore, SCATR-ITS yields the higher detection accuracy of the two methods, attaining an F$_1$ of 70.6\% at $d_{\text{th}}=3$ m, 5.9 pp higher than watershed. The performance gap further increases to 7 pp when using $d_{\text{th}}=5$ m. We focus subsequent analysis on metrics computed under the stricter $d_{\text{th}}=3$ m threshold.

Although watershed achieves the higher Precision at 91.1\% compared to 78.8\% for SCATR-ITS, this is primarily driven by watershed's tendency toward undersegmentation. Because watershed relies on a 2D CHM, understory trees are frequently merged into nearby dominant trees. In contrast, SCATR-ITS also segments understory trees, which inevitably increases the false positive rate and reduces Precision.

In terms of Recall, SCATR-ITS outperforms watershed by 13.7 pp, detecting 63.9\% of all trees (versus 50.2\% for watershed). Furthermore, crown-category-level metrics demonstrate that SCATR-ITS consistently outperforms watershed across all forest strata. The performance difference is most pronounced for small understory trees in crown categories C and D, where SCATR-ITS detects 48.5\% ($+34$ pp) and 20.9\% ($+16.8$ pp) of trees, respectively. In addition, SCATR-ITS retains a clear advantage even for the large, isolated trees in category A, achieving 93.3\% Recall, 4.7 pp higher than watershed.

Watershed remains the computationally faster option, processing a 1 km $\times$ 1 km tile in an average of 5.6 minutes, approximately 2.5 times faster than the 13.8 minutes required by SCATR-ITS. Nevertheless, given its significantly higher detection accuracy and improved segment quality, SCATR-ITS provides a favorable trade-off for large-area ITS applications, despite its higher execution time.

\begin{table*}[!t]
    \centering
    \caption{Performance comparison of SCATR-ITS and Watershed on the Padasjoki production area national laser scanning data. Accuracy metrics are reported for the subsections with reference data coverage, whereas average runtimes are reported for the full 1 km $\times$ 1 km plots. Accuracy is evaluated at distance thresholds ($d_{\text{th}}$) of 3 m and 5 m.} \bigskip
    \scriptsize{
    \begin{tabular}{lc*{7}{S[table-format=2.1]}*{3}{S[table-format=1.1,table-column-width=0.6cm]}}
        \toprule
        \multirow{2}{*}{\textbf{Model}} & \multicolumn{1}{c}{\multirow{2}{*}{\textbf{d$_\text{th}$ (m)}}} & \multicolumn{3}{c}{\textbf{Overall metrics (\%)}} & \multicolumn{4}{c}{\textbf{Crown category-level metrics (\%)}} & \multicolumn{3}{c}{\textbf{Average time (min/plot)}} \\ \cmidrule(lr){3-5} \cmidrule(lr){6-9} \cmidrule(lr){10-12}
        && \textbf{Precision} & \textbf{Recall} & \textbf{F$_1$} & \textbf{Recall$_\text{A}$} & \textbf{Recall$_\text{B}$} & \textbf{Recall$_\text{C}$} & \textbf{Recall$_\text{D}$} & \textbf{Tot.} & \textbf{Pre.} & \textbf{Seg.}  \\ \midrule \midrule
        Watershed \citep{yu2011predicting} & 3 & 91.1 & 50.2 & 64.7 & 88.6 & 66.9 & 14.5 & 4.1 & 5.6 & 3.9 & 1.7 \\
        & 5 & 91.9 & 50.6 & 65.3 & 89.3 & 67.1 & 15.0 & 4.4 \\ \midrule
        \textbf{SCATR-ITS (Ours)} & 3 & 78.8 & 63.9 & 70.6 & 93.3 & 71.3 & 48.5 & 20.9 & 13.8 & 6.5 & 7.3 \\
        & 5 & 80.8 & 65.5 & 72.3 & 94.8 & 71.6 & 50.6 & 24.6 \\
        \bottomrule
    \end{tabular}}
    \label{table:performance_comparison_padasjoki_20_pts}
\end{table*}

\begin{figure*}[!t]
\centering
\includegraphics[width=\textwidth]{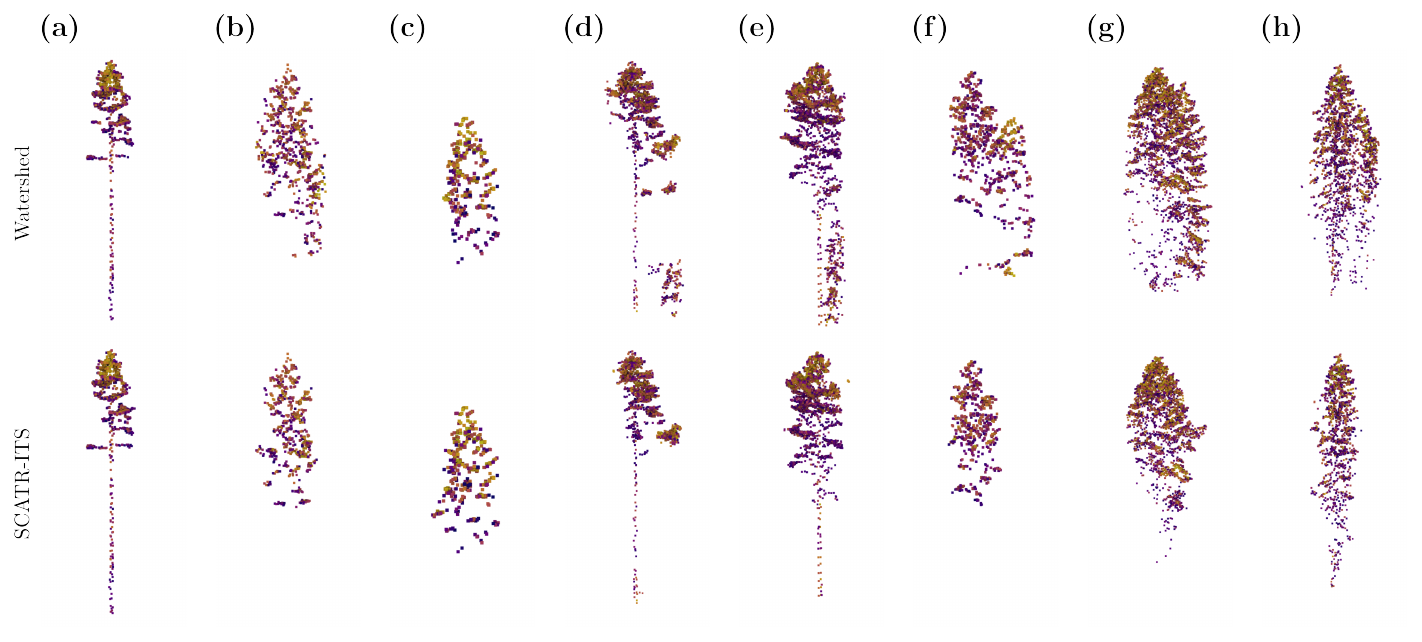}
\caption{Examples of individual tree segments produced by watershed and SCATR-ITS on the Finnish national laser scanning data (20 points/m$^2$). Both methods perform similarly in terms of segment quality when trees are sufficiently isolated and lack understory, whereas watershed degrades more than SCATR-ITS in dense or multi-layered stands. All visualized segments match field-collected reference trees ($d_{\text{th}}=3$ m). Colors represent intensity values, with higher values shown as brighter colors. (a) Pine, crown category A. (b) Birch, crown category A. (c) Spruce, crown category C. (d) Pine, crown category A. (e) Pine, crown category B. (f) Birch, crown category B. (g) Aspen, crown category A. The watershed segment merges an adjacent spruce. (h) Spruce, crown category B.}
\label{figure:padasjoki_20_pts_individual_segments}
\end{figure*}

Importantly, the employed evaluation strategy relies solely on tree locations and heights, completely disregarding 3D crown geometry. It follows that the reported detection metrics represent an optimistic upper bound of actual 3D segmentation performance. For watershed in particular, location-based evaluation may overestimate segmentation accuracy, as the segments can be poorly delineated despite correct positional alignment with ground truth. For example, in multi-layered forests, watershed segments frequently merge a large dominant tree with several suppressed understory trees. Conversely, SCATR-ITS operates directly in 3D, avoiding such 2D projection artifacts and yielding superior segment quality. A comparison of individual tree segments illustrating these qualitative differences is presented in \Cref{figure:padasjoki_20_pts_individual_segments}. We conjecture that the true performance gap between watershed and SCATR-ITS is more substantial than the detection metrics suggest. This premise aligns with the findings of \citet{ruoppa2026benchmarking}, where ForestFormer3D outperformed watershed by more than 10 pp in F$_1$-score on a 25 points/m$^2$ sparsified version of FGI-EMIT when evaluated against 3D ground truth.

Our watershed results are in line with \citet{hyyppa2024concepts}, who reported an F$_1$ of 59.2\% using the same algorithm and matching strategy on sparser Finnish national laser scanning data (5 points/m$^2$). Excluding ground classification, their average runtime of 1.5 minutes per 1 km $\times$ 1 km plot is directly comparable to our 1.7 minutes, further confirming the validity of our watershed baseline implementation for large-area ITS.

Due to the excessive computational cost of segmenting the full 142 km$^2$ area with ForestFormer3D, we instead estimated its runtime across 10 representative 1 km$^2$ tiles close to the median point count. ForestFormer3D averaged 4.28 hours per tile, making SCATR-ITS $\sim\!18.6\times$ faster on Finnish national laser scanning data. Overall, this experiment further demonstrates that while SCATR-ITS delivers accuracy on par with the state-of-the-art on benchmark datasets, it simultaneously reduces the runtime of 3D DL-based ITS to an operationally feasible scale for large-area ALS data.

Finally, we estimate that performing countrywide ITS for Finland (338,455 km$^2$, 20 points/m$^2$ data) using SCATR-ITS would require approximately 77,900 GPU-hours (8.89 GPU-years). Given that the data is processed in 1 km$^2$ tiles, the segmentation task is embarrassingly parallel. Distributing the workload across, for example, 128 GPUs would reduce the processing time to just $\sim\!25$ days of wall-clock time. Although substantial, this compute requirement demonstrates the feasibility of nation-scale 3D ITS, a task that remains prohibitively expensive using previous 3D DL models. Moreover, considering that Finnish national laser scanning operates in a nine-year acquisition cycle \citep{nls2026laserkeilaus}, and requires significant financial investments, the computational budget is reasonable in comparison.

\section{Discussion}

\subsection{Limitations}

Despite its efficiency, the proposed model has several limitations. First, while SCATR-ITS achieves instance segmentation accuracy comparable to ForestFormer3D, it omits the capability for fine-grained semantic segmentation since the employed superpoint representations do not strictly align with the wood--foliage semantic boundaries. Second, all segmentation errors occur at the superpoint level, making them larger in size than the localized errors of state-of-the-art point-level ITS models. Finally, because SCATR-ITS shares its core framework with ForestFormer3D, it inherits the same architectural limitations. Consequently, accurate segmentation of deciduous forests, sparse understory trees in high-density stands, and dense TLS scenes remains a challenge.

Although we demonstrate that SCATR-ITS achieves inference speeds that make national-scale application feasible, conventional ITS approaches like watershed remain demonstrably faster, even if substantially less accurate. Furthermore, because SCATR-ITS requires a GPU, whereas CPU hours are typically more cost-effective, there are also financial incentives to opt for conventional unsupervised methods in large-area processing. Finally, given that the runtime and VRAM usage of DL models depend on factors such as VRAM capacity and GPU architecture, the reported computational cost metrics are specific to the software and hardware employed in this study. While absolute inference speed and VRAM utilization may vary on other hardware, the relative model ranking is expected to remain unchanged.

While the proposed dynamic query count module demonstrably improves segmentation accuracy, it introduces several additional hyperparameters, specifically the query bin intervals and their corresponding query counts. Determining appropriate values for these parameters is a non-trivial task and may require empirical tuning. However, this limitation could potentially be circumvented by directly predicting the query count using, for example, density maps \citep{zhang2015cross,boominathan2016crowdnet}, although such learning objectives are inherently more challenging.

Lastly, existing instance-labeled forest point cloud data predominantly depict European boreal and temperate forests, as is the case for the datasets utilized in this study. Consequently, the reported performance may not directly generalize to structurally distinct forest environments, such as subtropical forests or rainforests.

\subsection{Prospects for future research}

The computational efficiency and lower VRAM requirements of SCATR-ITS unlock several promising avenues for future research. The superpoint-level processing frees capacity for the integration of more computationally complex feature encoders and decoders. For example, similar to \citet{wielgosz2026segmentanytreev2} and \citet{yue2026toward}, the sparse-convolution-based backbone could be replaced with a modern point cloud transformer encoder, such as LitePT \citep{yue2026litept}. Furthermore, the reduced VRAM requirements enable the development of more sophisticated approaches for incorporating non-geometric features. Rather than direct concatenation with $xyz$-coordinates, which degrades the performance of models like ForestFormer3D \citep{ruoppa2026benchmarking}, a dedicated spectral encoder could allow the model to effectively exploit reflectance information to improve ITS accuracy, similar to the unsupervised case \citep{dai2018new,huo2020individual}.

Further improving the inference speed of SCATR-ITS presents another important direction for future research. The low VRAM footprint during inference leaves room for potential parallelization: although the processing pipeline for overlapping input cylinders is inherently sequential, the workload could be partitioned into several disjoint subsets for parallel execution. However, the extent of practical speedups remains bounded by the GPU worker utilization rate. On the other hand, because the proposed maximal stride scales directly with the input radius, increasing the input size can accelerate inference. While the low VRAM usage makes larger inputs feasible, two issues limit practical application. First, existing ITS training data predominantly consist of small forest stands, constraining the spatial extent of model inputs. This limitation could be mitigated by manually labeling large continuous forest areas or by generating synthetic training data. Second, preliminary experiments indicate that when the input radius is increased, the default six-layer decoder fails to fully capture the extended spatial context and model performance degrades. Consequently, a deeper instance decoder may be essential. Whether the added computational cost of a deeper decoder would offset the efficiency gains of a larger input radius remains an open question.

Finally, developing a superpoint generator capable of constructing representations that respect semantic boundaries of forest point clouds is of high interest, as it would extend superpoint-based processing to full panoptic segmentation.

\section{Conclusions}

In this study, we proposed SCATR-ITS, a fast and scalable transformer-based model for individual tree segmentation from LiDAR point clouds. Built upon the ForestFormer3D framework, SCATR-ITS introduces several modifications designed to accelerate inference and reduce GPU VRAM usage while preserving accuracy. These include superpoint-level instance decoding, superpoint-level postprocessing, size-aware postprocessing, a dynamic query count module, and a maximal input stride formulation. Superpoint representations are generated using learned instance-discriminative embeddings combined with a parallelized greedy graph clustering algorithm.

SCATR-ITS achieved segmentation accuracy on par with the state of the art set by ForestFormer3D on both the FGI-EMIT and FOR-InstanceV2 datasets with an up to $8.5\times$ higher inference speed. Furthermore, the inference was approximately $30\times$ faster than SegmentAnyTree and even surpassed the 2D YOLOv12 when preprocessing was included, all while reducing GPU VRAM footprint by 2--3$\times$ during inference and $>\!5\times$ during training. An ablation study confirmed that the proposed architectural components successfully increased segmentation speed while retaining accuracy. Experiments on the LAUTx and Wytham Woods datasets further demonstrated that SCATR-ITS generalizes effectively to out-of-dataset forest scenes, maintaining an accuracy comparable to the state of the art, with a more moderate 3--5$\times$ speedup on the dense MLS and TLS point clouds.

Finally, we demonstrated the large-area segmentation capabilities of SCATR-ITS by processing a 142 km$^2$ region of Finnish national laser scanning data (20 points/m$^2$). The model achieved an average inference speed of 13.8 minutes/km$^2$, producing a significantly higher accuracy and superior segment quality compared to the unsupervised watershed baseline, which was approximately $2.5\times$ faster. Crucially, SCATR-ITS was an estimated $\sim\!18.6\times$ faster than ForestFormer3D on the same dataset.

Overall, our experiments show that SCATR-ITS delivers state-of-the-art accuracy at speeds where operational-scale adoption becomes viable, paving the way toward countrywide deployment. Furthermore, the model's significantly lower computational resource usage unlocks various prospects for improving its performance in future work.

\section*{CRediT authorship contribution statement}

\textbf{Lassi Ruoppa:} Writing - Original Draft, Conceptualization, Methodology, Software, Investigation, Data Curation, Visualization, Supervision. \textbf{Verneri Seppänen:} Writing - Review \& Editing, Software. \textbf{Antero Kukko:} Writing - Review \& Editing, Resources, Funding Acquisition. \textbf{Juha Hyyppä:} Writing - Review \& Editing, Resources, Supervision, Funding Acquisition.

\section*{Acknowledgments}

The study has received funding from the Research Council of Finland (RCF) through the following grants: ``High-performance computing allowing high-accuracy country-level individual tree carbon sink and biodiversity mapping'' (RCF 359203), ``High-performance computing for assessing boreal forest resources and sinks'' (RCF 380410), and ``Enhanced Wood Tracing Systems for Sustainable Forestry and Improved Wood Utilization'' (RCF 373290). The work was done under RCF Research Flagship ``Forest-Human-Machine Interplay -- Building Resilience, Redefining Value Networks and Enabling Meaningful Experiences'' (RCF 359175, 374285) using RCF research infrastructure ``Measuring Spatiotemporal Changes in Forest Ecosystem'' (RCF 346382). The work was additionally supported by the Ministry of Agriculture and Forestry grant ``Future Forest Information System at Individual Tree Level 2.0'' (VA-MMM-2024-25-1). The authors thank the Finnish Forest Centre for conducting the field measurements in the Padasjoki area and Tarmo Hietala for contributing to the code used for hyperparameter optimization of the watershed individual tree segmentation algorithm.

\section*{Data availability}

The individual tree segmentation datasets used in this study are publicly available: FGI-EMIT (\url{https://doi.org/10.5281/zenodo.19351234}), FOR-InstanceV2 (\url{https://doi.org/10.5281/zenodo.16742708}), AerialTrees (\url{https://github.com/A8366233/AerialTrees}), EvoMS (\url{https://doi.org/10.5281/zenodo.15913427}), and the extended LAUTx and Wytham Woods (\url{https://doi.org/10.25625/QUTUWU}).

The Finnish national laser scanning data (20 points/m$^2$, 2025 test flights) are not publicly available and cannot be redistributed by the authors due to restrictions on the use and distribution of the data. Comparable data are expected to become available through the National land survey of Finland (\url{https://www.maanmittauslaitos.fi/laserkeilaus-ja-ilmakuvaus}, in Finnish). Access will be subject to the provider's terms and conditions.

The source code and pretrained weights for the SCATR-ITS model introduced in this study are available in the following GitHub repository: \url{https://github.com/ruoppa/scatr_its}, which also contains the script used for preprocessing the FOR-InstanceV2 dataset into the format used in our experiments. The source code of each deep-learning-based performance baseline is publicly available: YOLOv12 (\url{https://github.com/sunsmarterjie/yolov12}), SegmentAnyTree (\url{https://github.com/SmartForest-no/SegmentAnyTree}), TreeLearn (\url{https://github.com/ecker-lab/treelearn}), and ForestFormer3D (\url{https://github.com/SmartForest-no/ForestFormer3D}). The watershed algorithm implementation employed in this study is private but can be obtained from the authors upon reasonable request.

The Python scripts used for segmentation accuracy evaluation and for sparsifying the point clouds in the ALS-XL-20-pts dataset are available at \url{https://github.com/ruoppa/fgi_emit}.

\scriptsize{\bibliography{library}}

\clearpage
\normalsize
\begin{appendices}
\renewcommand{\thesection}{\Alph{section}}
\crefalias{section}{appendix}
\crefalias{subsection}{appendix}
\crefalias{subsubsection}{appendix}
\crefname{appendix}{appendix}{appendices}
\Crefname{appendix}{Appendix}{Appendices}
\counterwithin{figure}{section}
\counterwithin{table}{section}

\section{Details of the ALS-XL-20-pts dataset} \label{appendix:als_xl_20pts}

\subsection{Training, validation, and test sets}

The ALS-XL-20-pts dataset combines all publicly available instance-annotated ALS and ULS data into a single composite dataset. The individual point clouds are all randomly downsampled to an approximate density of 20 points/m$^2$. Any instances with fewer than 5 points post-downsampling are removed, as such small and sparse objects are practically undetectable. Specifically, ALS-XL-20-pts consists of data from FOR-InstanceV2 (CULS, NIBIO, RMIT, SCION, TUWIEN, and Yuchen), FGI-EMIT, EvoMS, and the publicly available regions of AerialTrees (Gaofeng and Mulan). The exact data splits are detailed in \Cref{table:als_xl_20_pts_data_split}. The official training, validation, and test splits were adopted for FGI-EMIT and FOR-InstanceV2 data, while EvoMS was assigned entirely to the training set due to its limited size. In the absence of predefined data splits for AerialTrees, its plots were randomly allocated into training, validation, and test sets using a 70\%/10\%/20\% ratio.

\begin{table*}[!t]
    \centering
    \caption{Overview of training, validation, and test splits for ALS-XL-20-pts. For each split, we list the IDs of the corresponding forest plots from the original datasets and the final number of tree instances following downsampling and the removal of instances with fewer than 5 points.} \bigskip
    \scriptsize{
    \begin{tabular}{
            ll
            *{1}{S[
                table-format=5.0,
                group-separator={,},
                group-minimum-digits=4
            ]}
            L{9cm}
        }
        \toprule
        \textbf{Data split} & \textbf{Original dataset} & \multicolumn{1}{P{1cm}}{\textbf{Number of trees}} & \textbf{Plots} \\ \midrule \midrule
        Training & FOR-InstanceV2 & 2551 & CULS 1, NIBIO 3, NIBIO 4, NIBIO 6, NIBIO 7, NIBIO 10, NIBIO 11, NIBIO 12, NIBIO 21, RMIT train, SCION 39, SCION 87, TUWIEN train, NIBIO2 2, NIBIO2 8, NIBIO2 9, NIBIO2 12, NIBIO2 14, NIBIO2 16, NIBIO2 20, NIBIO2 21, NIBIO2 22, NIBIO2 23, NIBIO2 24, NIBIO2 25, NIBIO2 26, NIBIO2 28, NIBIO2 31, NIBIO2 33, NIBIO2 38, NIBIO2 39, NIBIO2 40, NIBIO2 42, NIBIO2 43, NIBIO2 45, NIBIO2 46, NIBIO2 50, NIBIO2 51, NIBIO2 55, NIBIO2 56, NIBIO2 57, NIBIO2 61, Yuchen train \\
        & FGI-EMIT & 801 & 1001, 1003, 1005, 1009, 1010, 1013, 1020, 1023, 1024, 1027 \\
        & EvoMS & 355 & 1, 2 \\
        & AerialTrees & 1169 & Gaofeng 1, Gaofeng 2, Gaofeng 5, Gaofeng 6, Gaofeng 7, Gaofeng 9, Gaofeng 10, Gaofeng 11, Gaofeng 12, Gaofeng 15, Gaofeng 17, Gaofeng 18, Gaofeng 19, Gaofeng 20, Gaofeng 21, Gaofeng 22, Mulan 3, Mulan 6, Mulan 7, Mulan 9, Mulan 10, Mulan 11, Mulan 13, Mulan 14, Mulan 16, Mulan 18, Mulan 19, Mulan 20, Mulan 21, Mulan 22, Mulan 24, Mulan 25, Mulan 28, Mulan 29, Mulan 30, Mulan 31, Mulan 32, Mulan 33, Mulan 34, Mulan 35 \\
        & \textbf{Total} & \bfseries{\tablenum{4876}} & \\ \midrule
        Validation & FOR-InstanceV2 & 611 & CULS 3, NIBIO 2, NIBIO 8, NIBIO 9, NIBIO 13, NIBIO 16, NIBIO 19, SCION 35, NIBIO2 13, NIBIO2 41, NIBIO2 44, NIBIO2 47, NIBIO2 54, NIBIO2 59, Yuchen train \\
        & FGI-EMIT & 233 & 1019, 1022, 1031 \\
        & AerialTrees & 226 & Gaofeng 14, Gaofeng 16, Mulan 1, Mulan 4, Mulan 8, Mulan 15 \\
        & \textbf{Total} & \bfseries{\tablenum{1070}} & \\ \midrule
        Test & FOR-InstanceV2 & 1158 & CULS 2, NIBIO 1, NIBIO 5, NIBIO 17, NIBIO 18, NIBIO 22, NIBIO 23, RMIT test, SCION 31, SCION 61, TUWIEN test, NIBIO2 1, NIBIO2 3, NIBIO2 6, NIBIO2 10, NIBIO2 15, NIBIO2 27, NIBIO2 32, NIBIO2 34, NIBIO2 35, NIBIO2 48, NIBIO2 49, NIBIO2 52, NIBIO2 53, NIBIO2 58, NIBIO2 60,  Yuchen test \\
        & FGI-EMIT & 427 & 1002, 1004, 1008, 1012, 1018, 1028 \\
        & AerialTrees & 284 & Gaofeng 3, Gaofeng 4, Gaofeng 8, Gaofeng 13, Gaofeng 23, Mulan 2, Mulan 5, Mulan 12, Mulan 17, Mulan 23, Mulan 26, Mulan 27 \\
        & \textbf{Total} & \bfseries{\tablenum{1869}} & \\
        \bottomrule
    \end{tabular}}
    \label{table:als_xl_20_pts_data_split}
\end{table*}

\subsection{Segmentation performance on ALS-XL-20-pts}

\Cref{table:performance_comparison_als_xl_20_pts} summarizes the segmentation performance of SCATR-ITS alongside the two baselines, watershed and ForestFormer3D, on ALS-XL-20-pts. Consistent with the findings on other datasets presented in \Cref{section:scatr_its_performance,section:generalization_lautx_wytham}, SCATR-ITS performs on par with ForestFormer3D on, achieving an F$_1$-score of 79.7\% ($-0.5$ pp). Simultaneously, SCATR-ITS is approximately $11\times$ faster, processing each plot in an average of 0.35 s. Conversely, Although watershed is $1.75\times$ faster than SCATR-ITS, its accuracy is substantially lower, yielding an F$_1$-score of 54.9\% ($-24.8$ pp). Notably, this performance gap is much starker than that observed on the Finnish national laser scanning data in \Cref{section:large_area_performance}. This further supports our previous hypothesis that watershed's atypically high accuracy on the national data stems from the position-based evaluation, which completely disregards 3D crown structure. Additionally, whereas the evaluation on the Finnish national laser scanning data relied on field-collected tree positions, performance on ALS-XL-20-pts is evaluated only using instances visible in the 3D point clouds. Consequently, the national laser scanning reference likely includes several trees that are invisible in the LiDAR data, which may explain the comparatively higher Recall values observed on ALS-XL-20-pts.

\begin{table*}[!t]
    \centering
    \caption{Comparison of SCATR-ITS, ForestFormer3D and Watershed on the ALS-XL-20-pts test set. Best performance is in \textbf{bold}, and second-best is \underline{underlined}.} \bigskip
    \scriptsize{
    \begin{tabular}{l*{5}{S[table-format=2.1]}*{3}{S[table-format=1.2]}}
        \toprule
        \multirow{2}{*}{\textbf{Model}} & \multicolumn{1}{c}{\multirow{2}{*}{\textbf{Precision (\%)}}} & \multicolumn{1}{c}{\multirow{2}{*}{\textbf{Recall (\%)}}} & \multicolumn{1}{c}{\multirow{2}{*}{\textbf{F$_{\mathbf{1}}$ (\%)}}} & \multicolumn{1}{c}{\multirow{2}{*}{\textbf{Cov (\%)}}} & \multicolumn{1}{c}{\multirow{2}{*}{\textbf{AP$_{\mathbf{50}}$ (\%)}}} & \multicolumn{3}{c}{\textbf{Average time (s/plot)}} \\
        \cmidrule(lr){7-9}
        &&&&&& \textbf{Tot.} & \textbf{Pre.} & \textbf{Seg.}  \\ \midrule \midrule
        Watershed \citep{yu2011predicting} & 73.6 & 43.8 & 54.9 & 41.3 & \text{--} & \textbf{0.20} & 0.16 & \textbf{0.04} \\ 
        ForestFormer3D \citep{xiang2025forestformer3d} & \textbf{88.1} & \textbf{73.7} & \textbf{80.2} & \textbf{68.2} & \textbf{72.3} & 3.83 & \textbf{0.01} & 3.82 \\
        \textbf{SCATR-ITS (Ours)} & \underline{87.1} & \underline{73.5} & \underline{79.7} & \underline{67.4} & \underline{71.3} & \underline{0.35} & \underline{0.13} & \underline{0.21} \\
        \bottomrule
    \end{tabular}}
    \label{table:performance_comparison_als_xl_20_pts}
\end{table*}

\section{SCATR-ITS training details} \label{appendix:scatr_its_training_details}

\subsection{Loss functions}

Both SCATR-ITS and the superpoint generator combine multiple loss functions to optimize their respective prediction heads, with some predictions contributing to several loss terms. In this section, we first present the mathematical formulations of the individual losses (\Cref{appendix:bce_loss,appendix:mse_loss,appendix:lovasz_hinge_loss,appendix:dice_loss,appendix:instance_disc_loss}), followed by the full loss functions for the superpoint generator (\Cref{appendix:sp_generator_loss}) and SCATR-ITS (\Cref{appendix:scatr_its_loss}).

\subsubsection{BCE loss} \label{appendix:bce_loss}

Binary cross-entropy is one of the most widely employed loss functions for binary classification tasks. In SCATR-ITS, BCE is used to supervise both semantic and instance segmentation, as well as the dynamic query count module. Given logits $\mathbf{z}\in\real^N$ and binary ground truth labels $\mathbf{y}\in\{0,1\}^N$, the BCE loss is defined as:
\begin{equation}
    \mathcal{L}_{\text{BCE}}\left(\mathbf{z},\mathbf{y}\right)=-\frac{1}{N}\sum_{i=1}^N \left[y_i\log\sigma(z_i)+(1-y_i)\log(1-\sigma(z_i))\right],
\end{equation}
where $z_i$ and $y_i$ denote the $i$th elements of $\mathbf{z}$ and $\mathbf{y}$, respectively.

\subsubsection{MSE loss} \label{appendix:mse_loss}

Similar to the BCE loss, mean squared error is a widely used loss function in machine learning and deep learning. Unlike BCE, MSE is most commonly employed for regression tasks. In SCATR-ITS, the MSE loss is used to supervise instance confidence scores. Given predicted and ground truth values, denoted by $\hat{\mathbf{y}}\in\real^N$ and $\mathbf{y}\in\real^N$, respectively, the MSE loss is computed as follows: 
\begin{equation}
    \mathcal{L}_{\text{MSE}}\left(\hat{\mathbf{y}},\mathbf{y}\right)=\frac{1}{N}\sum_{i=1}^N \left(\hat{y}_i-y_i\right)^2.
\end{equation}

\subsubsection{Lovász hinge loss} \label{appendix:lovasz_hinge_loss}

The Lovász hinge loss proposed by \citet{berman2018lovasz} provides an alternative to BCE for binary segmentation tasks. Before defining the full loss function, we introduce the necessary auxiliary definitions and notation. Let $\mathbf{z}\in\real^N$ denote the unnormalized logits and $\mathbf{y}\in\{0,1\}^N$ the corresponding binary ground truth labels. We define the index set of the positive class, denoted by $\mathbfcal{Y}$, as:
\begin{equation}
    \mathbfcal{Y}=\Big\{i\in\big\{1,\ldots,N\big\}\;|\;y_i=1\Big\}.
\end{equation}
That is, $\mathbfcal{Y}$ contains index $i$ if and only if the ground truth label $y_i$ belongs to the positive class.

We express the Jaccard loss in the binary case as a set function parametrized by $\mathbf{y}$, formally defined as $\mathcal{L}_{\text{Jaccard},\mathbf{y}}:\mathbf{2}^{\{1,\ldots,N\}}\mapsto\real$:
\begin{equation} \label{equation:jaccard_misprediction}
    \mathcal{L}_{\text{Jaccard},\mathbf{y}}(\hat{\mathbfcal{Y}})=\frac{|\hat{\mathbfcal{Y}}|}{|\hat{\mathbfcal{Y}}\cup \mathbfcal{Y}|}.
\end{equation}
Under the standard definition of the Jaccard index, $\hat{\mathbfcal{Y}}$ denotes the index set of mispredictions. That is, the set of indices for which the predicted binary label $\mathbf{\hat{y}}$ differs from the ground truth:
\begin{equation}
    \hat{\mathbfcal{Y}}=\Big\{i\in\{1,\ldots,N\}\;|\;\hat{y}_i\neq y_i\Big\}.
\end{equation}
\Cref{equation:jaccard_misprediction} is then equivalent to one minus the IoU between the predicted and ground truth positive labels.

Let the $i$th element of a margin vector $\boldsymbol{\ell}\in\real_0^N$ be:
\begin{equation}
    \ell_i=\max\Big(1-z_i(2y_i -1), 0\Big),
\end{equation}
that is, the element-wise hinge loss applied to the logits $\mathbf{z}$. Furthermore, let $\boldsymbol{\pi}$ denote a permutation such that:
\begin{equation}
    \ell_{\pi_1}\geq \ell_{\pi_2}\geq\cdots\geq \ell_{\pi_N}.
\end{equation}
We then define the corresponding nested index sets as:
\begin{equation}
    \mathbfcal{O}_i:=\big\{\pi_1,\ldots,\pi_i\big\},\quad \mathbfcal{O}_0:=\emptyset.
\end{equation}

The Lovász extension of a set function $g:\{0,1\}^N\rightarrow\real$ is given by \citep{berman2018lovasz}:
\begin{equation}
    g^L(\mathbf{\boldsymbol{\ell}})=\sum_{i=1}^N \ell_{\pi_i}\Big(g(\mathbfcal{O}_i)-g(\mathbfcal{O}_{i-1})\Big).
\end{equation}
The Lovász hinge loss is then obtained by evaluating the Lovász extension of the Jaccard loss set function at the margin vector $\boldsymbol{\ell}$:
\begin{equation}
    \mathcal{L}_{\text{Lovász}}(\mathbf{z},\mathbf{y})=\mathcal{L}_{\text{Jaccard},\mathbf{y}}^L(\boldsymbol{\ell}).
\end{equation}

Under the standard formulation, the Jaccard loss set function defined in \Cref{equation:jaccard_misprediction} is submodular \citep{yu2020lovasz}. Crucially, the Lovász extension of a set function $g(\cdot)$ is convex if and only if $g(\cdot)$ is submodular \citep{lovasz1983submodular}. Thus, submodularity of the Jaccard loss set function guarantees that its Lovász extension is convex with respect to the margin vector $\boldsymbol{\ell}$. The resulting Lovász hinge loss provides a convex surrogate for optimizing the IoU of the positive class.

\subsubsection{Dice loss} \label{appendix:dice_loss}

The Dice loss is typically used for semantic segmentation and tasks involving significant class imbalance. Notably, it is well suited for scenarios in which the target occupies only a small fraction of the full scene, rendering it suitable for training binary masks to extract individual trees. Given a vector of unnormalized logits $\mathbf{z}\in\real^N$ and the corresponding binary ground truth labels $\mathbf{y}\in\{0,1\}^N$, the Dice loss is defined as:
\begin{equation}
    \mathcal{L}_{\text{Dice}}\left(\mathbf{z},\mathbf{y}\right)=1-\frac{1+2\sum_{i=1}^N \sigma(z_i)y_i}{1+\sum_{i=1}^N \sigma(z_i)+\sum_{i=1}^N y_i}.
\end{equation}

\subsubsection{Instance-discriminative loss} \label{appendix:instance_disc_loss}

The instance-discriminative loss, first proposed by \citet{debrabandere2017semantic} and adapted for point cloud data by \citet{wang2019associatively}, is designed for learning feature representations that are compact within each instance while remaining well separated across distinct instances. Let $\mathbf{F}\in\real^{N\times D}$ denote the $D$-dimensional feature representations of the $N$ points, and let $\mathbf{Y}_{\text{inst}}\in\{0,1\}^{C\times N}$ denote the binary instance masks for the $C$ instances. Furthermore, let $\boldsymbol{\mu}_c\in\real^D$ denote the mean feature embedding of instance $c\in\{1,\ldots,C\}$:
\begin{equation}
    \boldsymbol{\mu}_c=\frac{1}{N_c}\sum_{i\in\mathbf{Y}_{\text{inst},c}}\mathbf{f}_i,
\end{equation}
where $\mathbf{f}_i\in\real^D$ is the feature vector corresponding to the $i$th point.

The instance-discriminative loss consists of three terms. The first enforces intra-instance compactness by pulling feature representations within each instance toward their corresponding mean (\Cref{equation:loss_instdisc_var}). The second pushes the means of distinct instances apart (\Cref{equation:loss_instdisc_dist}). The third applies regularization by penalizing large mean embeddings, ensuring they remain close to the origin (\Cref{equation:loss_instdisc_reg}).
\begin{align}
    &\mathcal{L}_{\text{Var}}\left(\mathbf{F},\mathbf{Y}_{\text{inst}}\right)=\frac{1}{C}\sum_{c=1}^C\frac{1}{N_c}\sum_{i\in\mathbf{Y}_{\text{inst},c}}\Big[\max\big(||\mathbf{f}_i-\boldsymbol{\mu}_c||_1-\delta_v,0\big)\Big]^2, \label{equation:loss_instdisc_var} \\
    &\mathcal{L}_{\text{Dist}}\left(\mathbf{F},\mathbf{Y}_{\text{inst}}\right)=\frac{1}{C(C-1)}\sum_{c_1\neq c_2}\Big[\max\big(2\delta_d-||\boldsymbol{\mu}_{c_1}-\boldsymbol{\mu}_{c_2}||_1,0\big)\Big]^2, \label{equation:loss_instdisc_dist} \\
    &\mathcal{L}_{\text{Reg}}\left(\mathbf{Y}_{\text{inst}}\right)=\frac{1}{C}\sum_{c=1}^C||\boldsymbol{\mu}_c||_1,\label{equation:loss_instdisc_reg}
\end{align}
where $||\cdot||_1$ is the L$_1$-norm and $\delta_v,\delta_d\in\real_0$ are margin parameters controlling the desired intra-instance compactness and inter-instance separation, respectively. The full instance-discriminative loss function is defined as:
\begin{align}
    \mathcal{L}_{\text{InstDisc}}\left(\mathbf{F},\mathbf{Y}_{\text{inst}}\right)&=\mathcal{L}_{\text{Var}}\left(\mathbf{F},\mathbf{Y}_{\text{inst}}\right)\nonumber\\
    &\quad+\mathcal{L}_{\text{Dist}}\left(\mathbf{F},\mathbf{Y}_{\text{inst}}\right)\nonumber\\
    &\quad+w_r\mathcal{L}_{\text{Reg}}\left(\mathbf{Y}_{\text{inst}}\right),
\end{align}
where $w_r\in\real_0$ is the regularization strength. Following \citet{debrabandere2017semantic,wang2019associatively,xiang2025forestformer3d}, we set $\delta_v=0.5$, $\delta_d=1.5$, and $w_r=0.001$.

\subsubsection{Full superpoint generator loss function} \label{appendix:sp_generator_loss}

Let $\mathbf{F}_1\in\real^{N\times 5}$ denote the 5D instance-discriminative features, $\mathbf{F}_2\in\real^N$ the foreground-background logits, $\mathbf{Y}_{\text{inst}}\in\{0,1\}^{C\times N}$ the ground truth instance masks, and $\mathbf{Y}_{\text{sem}}\in\{0,1\}^{N}$ the ground truth semantic mask. The superpoint generator uses the same backbone and MLP branches as ForestFormer3D. Therefore, we adopt a similar loss function. Specifically, we use the same instance-discriminative loss to train the 5D branch, and the Lovász hinge loss to train the foreground-background branch. The total loss for a single forest scene is:
\begin{equation}
    \mathcal{L}=
    \mathcal{L}_{\text{InstDisc}}\left(\mathbf{F}_1,\mathbf{Y}_{\text{inst}}\right)+
    \mathcal{L}_{\text{Lovász}}\left(\mathbf{F}_2,\mathbf{Y}_{\text{sem}}\right).
\end{equation}
As noted in \Cref{section:superpoint_training}, we replace the original BCE loss for the foreground-background branch with the Lovász hinge loss, since experiments indicate it provides a slight improvement in superpoint quality.

\subsubsection{Full SCATR-ITS loss function} \label{appendix:scatr_its_loss}

Let $\mathbf{F}_1\in\real^{N\times 5}$ denote the 5D instance-discriminative features, $\mathbf{F}_2\in\real^N$ the foreground-background logits, $\mathbf{Y}_{\text{inst}}\in\{0,1\}^{C\times N}$ the ground truth instance masks, and $\mathbf{Y}_{\text{sem}}\in\{0,1\}^{N}$ the ground truth semantic mask. Let $\mathbf{Z}_Q\in\real^4$ denote the ordinal query bin logits and $\mathbf{Y}_Q\in\{0,1\}^4$ the corresponding binary query bin labels computed from the scene-level instance count $C$. Furthermore, let $\mathbf{Z}^{\mathbfcal{S}}_{\text{inst}}\in\real^{K\times M}$ denote the predicted superpoint-level instance mask logits, where $M$ and $K$ are the numbers of superpoints and predicted instances, respectively, and let $\mathbf{S}_{\text{inst}}\in\real^K$ denote the corresponding predicted confidence scores. Similarly, let $\mathbf{Z}^{\mathbfcal{S}}_{\text{sem}}\in\real^M$ denote the predicted semantic mask logits for the foreground class. Finally, let $\mathbf{Y}^{\mathbfcal{S}}_{\text{inst}}\in\{0,1\}^{C\times M}$ denote the ground truth instance masks and $\mathbf{Y}^{\mathbfcal{S}}_{\text{sem}}\in\{0,1\}^M$ the ground truth foreground mask, both aggregated at the superpoint level by majority voting.

The loss function used to train SCATR-ITS is essentially identical to that of ForestFormer3D, although SCATR-ITS computes the instance mask, semantic mask and score losses at the superpoint level, rather than the voxel level like ForestFormer3D. The foreground-background and instance-discriminative branches are trained using BCE loss and instance-discriminative loss, respectively. BCE loss is also applied to the ordinal query bin predictions from the dynamic query count module and the superpoint-level semantic predictions produced by the transformer decoder. For the $K$ instance predictions, a combination of BCE loss and Dice loss is used for the superpoint-level instance mask logits, while MSE loss is applied to the predicted confidence scores.

Let $\mathbfcal{M}\in\{1,\ldots,K\}\times\{1,\ldots,C\}$ denote the set of prediction--ground truth index pairs obtained through one-to-many matching (see \Cref{section:scatr_its_training}). The full instance mask losses are computed by averaging the losses over the matched pairs of predicted mask logits $\mathbf{Z}^{\mathbfcal{S}}_{\text{inst},i}$ and ground truth masks $\mathbf{Y}^{\mathbfcal{S}}_{\text{inst},j}$, where $(i,j)\in\mathbfcal{M}$. The IoU between the predicted and ground truth superpoint-level mask is employed as the score loss target. The full loss function for a single forest scene is:
\begin{align}
    \mathcal{L}&=\mathcal{L}_{\text{InstDisc}}\left(\mathbf{F}_1,\mathbf{Y}_{\text{inst}}\right)\nonumber\\
    &\quad+\mathcal{L}_{\text{BCE}}\left(\mathbf{F}_2,\mathbf{Y}_{\text{sem}}\right)\nonumber\\
    &\quad+\mathcal{L}_{\text{BCE}}\left(\mathbf{Z}_Q,\mathbf{Y}_{Q}\right)\nonumber\\
    &\quad+w_{\text{sem}}\mathcal{L}_{\text{BCE}}\left(\mathbf{Z}^{\mathbfcal{S}}_{\text{sem}},\mathbf{Y}^{\mathbfcal{S}}_{\text{sem}}\right)\nonumber\\
    &\quad+\frac{w_{\text{inst},1}}{|\mathbfcal{M}|}\sum_{(i,j)\in\mathbfcal{M}}\mathcal{L}_{\text{BCE}}\left(\mathbf{Z}^{\mathbfcal{S}}_{\text{inst},i},\mathbf{Y}^{\mathbfcal{S}}_{\text{inst},j}\right)\nonumber\\
    &\quad+\frac{w_{\text{inst},2}}{|\mathbfcal{M}|}\sum_{(i,j)\in\mathbfcal{M}}\mathcal{L}_{\text{Dice}}\left(\mathbf{Z}^{\mathbfcal{S}}_{\text{inst},i},\mathbf{Y}^{\mathbfcal{S}}_{\text{inst},j}\right)\nonumber\\
    &\quad+\frac{w_{\text{score}}}{|\mathbfcal{M}|}\sum_{(i,j)\in\mathbfcal{M}}\mathcal{L}_{\text{MSE}}\left(\mathbf{S}_{\text{inst},i},\text{IoU}\left[\sigma\left(\mathbf{Z}^{\mathbfcal{S}}_{\text{inst},i}\right),\mathbf{Y}^{\mathbfcal{S}}_{\text{inst},j}\right]\right).
\end{align}
Following \citet{xiang2025forestformer3d}, the instance mask and score losses are also applied to the intermediate outputs from each of the five preceding transformer decoder layers. Unmatched instance predictions are assigned all-zero targets for the instance mask and confidence score losses. Following \citet{xiang2025forestformer3d}, we set the loss weights to $w_{\text{sem}}=0.2$, $w_{\text{inst},1}=1.0$, $w_{\text{inst},2}=1.0$, and $w_{\text{score}}=0.5$.

We also experimented with superpoint-size-aware loss functions for the instance decoder, equivalent to the size-aware postprocessing described in \Cref{section:superpoint_level_postprocessing}. As an example, the superpoint-size-aware BCE loss is defined as:
\begin{equation}
    \mathcal{L}_{\text{BCE}}=-\frac{1}{\sum_{i=1}^M m_i}\sum_{i=1}^M m_i \left[y_i\log\sigma(z_i)+(1-y_i)\log(1-\sigma(z_i))\right],
\end{equation}
where $M$ is the number of superpoints and $m_i$ is the size of the $i$th superpoint. However, our experiments indicate that, within the SCATR-ITS framework, superpoint-size weighting has effects ranging from negligible to detrimental. Consequently, we do not apply size-based weighting in the final model loss function. Incorporating superpoint size as an additional input feature may provide an alternative approach to accounting for superpoint size during training, but we leave this question for future work.

\subsection{Other training hyperparameters}

\subsubsection{Superpoint generator}

As mentioned in \cref{section:superpoint_generator}, the lack of instance decoder significantly reduces the computational overhead and simplifies the learning objective, permitting the use of relatively large radii for the cylindrical inputs of the superpoint generator. Consequently, because a larger radius directly accelerates the superpoint generation pipeline, we chose the radius corresponding to each training dataset relative to its maximum plot size, 32 m for FGI-EMIT and ALS-XL-20-pts, and 46 m for FOR-InstanceV2. Besides input radius, other model hyperparameters, including batch size, optimizer, and the dimension of extracted feature vectors, remained equivalent to those used by ForestFormer3D. 

Following standard conventions in deep learning, training convergence was monitored using the training and validation losses and superpoint quality metrics (see \Cref{section:superpoint_quality_metrics}). Consequently, the number of training epochs required for convergence was determined to be 1,000, 800, and 700, for FGI-EMIT, FOR-InstanceV2 and ALS-XL-20-pts datasets, respectively. 

\subsubsection{SCATR-ITS}

To isolate the effects of our proposed improvements, model hyperparameters other than the number of training epochs were kept identical to ForestFormer3D on both FGI-EMIT and FOR-InstanceV2. Default parameter values were also used for ALS-XL-20-pts, except for adjustments to the input extent made possible by the substantially sparser data. SCATR-ITS was trained using an AdamW optimizer with an initial learning rate of 0.0001, $\beta=(0.9,0.999)$, a weight decay of 0.05, and a polynomial learning rate scheduler. The batch size and voxel size were set to 2 and 0.2 m, respectively. Data augmentation during training consisted of random horizontal flipping, random rotation around the $z$-axis, and random scaling by a factor in the range $[0.8, 1.2]$.

Following \citet{ruoppa2026benchmarking} and \citet{xiang2025forestformer3d}, the input radii were set to 12 m and 16 m for FGI-EMIT and FOR-InstanceV2, respectively. For ALS-XL-20-pts, the input radius was increased to 24 m to accelerate inference, taking advantage of the lower number of understory trees due to data sparsity as well as the reduced VRAM requirements.

SCATR-ITS was trained for 8,500 epochs on FGI-EMIT, 3,600 epochs on FOR-InstanceV2, and 3,500 epochs on ALS-XL-20-pts. In all cases, following \citet{xiang2025forestformer3d}, the first 1,000 epochs served as a warmup, during which the instance decoder remained inactive. Similar to the superpoint generator, convergence was monitored using the training and validation losses, and the corresponding segmentation accuracy metrics.

\section{Performance baseline training details} \label{appendix:baseline_training_details}

This appendix outlines the training details for the individual tree segmentation baselines trained from scratch for this study. Existing pretrained model weights were used where available. Refer to the original publications for the corresponding training setups and hyperparameter values.

For FGI-EMIT, only YOLOv12 was retrained due to improvements in DTM construction. For SegmentAnyTree, TreeLearn, and ForestFormer3D we utilized existing pretrained weights from the authors' previous work \citep{ruoppa2026benchmarking}. On FOR-InstanceV2, official pretrained model weights were used for ForestFormer3D \citep{xiang2025forinstancev2}, whereas the other three models were trained from scratch using setups and hyperparameter values adopted from the original publications. Because ALS-XL-20-pts is newly introduced in this work, both ForestFormer3D and watershed were trained from scratch. Where hyperparameter values are not explicitly listed, default values were used.

\subsection{YOLOv12}

Prior to training, the 3D point clouds are transformed into 2D images by projecting them onto the $xy$-plane. Image generation follows the procedure described in \citep{ruoppa2026benchmarking}, which is based on the approach of \citet{straker2023instance}. First, ground points are identified using a cloth simulation filter \citep{zhang2016easy} with a cloth resolution of 0.2 m and a ground distance threshold of 0.6 m, both of which were tuned empirically. A digital terrain model is then constructed using the ground points, and the $z$-coordinates are normalized by subtracting the ground elevation. The point clouds are rasterized on a 0.1 m grid, assigning the corresponding maximum $z$-coordinate value to each cell. These values are subsequently $z$-score normalized, and scaled to the interval $[0,1]$. Finally, pseudo-color images are generated from the raster using the \emph{inferno} colormap. For details on transforming the ground truth instance labels into 2D image format, we refer to \citet{ruoppa2026benchmarking}.

Following \citet{ruoppa2026benchmarking}, YOLOv12 was trained on FGI-EMIT and FOR-InstanceV2 using the default input image size ($640\times640$) and a batch size of 4. An AdamW optimizer with an initial learning rate of 0.001667, a weight decay of 0.0005, and a cosine learning rate scheduler was used for training. The number of epochs was set to 1,000, with early stopping triggered after 100 consecutive epochs with no observed improvement in validation accuracy. Models were initialized using publicly available weights pretrained on the MSCOCO 2017 dataset \citep{lin2014microsoft}.

\subsection{SegmentAnyTree}

SegmentAnyTree was trained on FOR-InstanceV2 using the default setup presented in the original paper \citep{wielgosz2024segmentanytree}, which was also utilized previously for FGI-EMIT. The input cylinder radius was set to 8 m and the batch size to 4. An Adam optimizer with a learning rate of 0.001, $\beta=(0.9,0.999)$, no weight decay, and an exponential learning rate scheduler was used for training. Based on training and validation loss curves and accuracy trends, the number of epochs was set to 160, with the first 30 epochs serving as a warmup during which the clustering stage was omitted. As in the original paper, we employed a data augmentation strategy in which point clouds downsampled to 1,000, 500, 100, 75, 50, 25, and 10 points/m$^2$ are incorporated into the training data.

\subsection{TreeLearn}

We adopt the exact setup described by \citet{xiang2025forestformer3d} for training TreeLearn on FOR-InstanceV2. A total of 2,500 samples with a tile size of 30 m were generated for training the model, and the batch size was set to 4. The training employed an AdamW optimizer with an initial learning rate of 0.001, a weight decay of 0.001, $\beta=(0.9,0.999)$, and a cosine learning rate scheduler. The model was trained for a total of 1,200 epochs.

In contrast to \citet{xiang2025forestformer3d}, we included the Yuchen region in the training, validation, and test sets, even though TreeLearn struggles with such sparse data. This inclusion ensured that segmentation performance remained fully comparable across all evaluated ITS methods, rather than being specific to a subset of the FOR-InstanceV2 dataset. Because we observed no significant performance drop for TreeLearn compared to the metrics in \citet{xiang2025forestformer3d}, we conclude that this training configuration is appropriate.

\subsection{ForestFormer3D}

ForestFormer3D was trained on ALS-XL-20-pts following mostly the default setup presented by \citet{xiang2025forestformer3d}. Specifically, the training employed an AdamW optimizer with an initial learning rate of 0.0001, $\beta=(0.9,0.999)$, a weight decay of 0.05, and a polynomial learning rate scheduler. Moreover, the batch size was set to 2 and the voxel size to 0.2 m. The instance score threshold was set to the default value of 0.4. Data augmentation during training consisted of random horizontal flipping, random rotation around the $z$-axis, and random scaling by a factor in the range $[0.8, 1.2]$.

Similar to SCATR-ITS, the input radius was increased to 24 m to accelerate inference, a modification enabled by the substantially sparser data. The number of queries was increased from the default 300 to 900, as preliminary experiments indicated that tripling the query count when doubling the radius yielded optimal segmentation accuracy (see \Cref{section:dynamic_query_count}). Based on a combination of training and validation loss curves along with accuracy trends, the number of epochs was set to 2,600.

\subsection{Watershed}

\citet{ruoppa2026benchmarking} observed that random search yielded hyperparameter values for the watershed ITS algorithm that resulted in slightly higher segmentation accuracy than those obtained via Bayesian optimization. Consequently, we optimized the hyperparameters of watershed on ALS-XL-20-pts using random search. The parameter search space consisted of the rasterization resolution, the standard deviation of the Gaussian filter ($\sigma$), and the window sizes for both the Gaussian and maximum filters. The optimal parameter values, tested ranges, and step sizes are listed in \Cref{table:watershed_optimal_params}.

\begin{table}[!t]
    \centering
    \caption{Optimal hyperparameter values and tested ranges of the watershed individual tree segmentation algorithm. GF and MF denote Gaussian filter and maximum filter, respectively} \bigskip
    \scriptsize{
    \begin{tabular}{lR{1cm}R{1cm}R{1.2cm}r}
        \toprule
        \textbf{Hyperparameter} & \textbf{Default value} & \textbf{Optimal value} & \textbf{Tested range} & \textbf{Step size} \\ \midrule \midrule
        Resolution & 0.5 & 0.6 & [0.05,1] & 0.05 \\
        $\sigma$ & 0.7 & 0.7 & [0.1,6] & 0.1 \\
        Window size (GF) & 5 & 39 & [3,41] & 2 \\
        Window size (MF) & 5 & 3 & [3,41] & 2 \\
        \bottomrule
    \end{tabular}
    }
    \label{table:watershed_optimal_params}
\end{table}

\section{Region-level accuracy metrics on the FOR-InstanceV2 test split} \label{appendix:performance_comparison_fiv2_region_level}

The region-level segmentation accuracy metrics for SCATR-ITS and the four baselines on FOR-InstanceV2 are presented in \Cref{table:performance_comparison_fiv2_regions}. ForestFormer3D generally yields the highest segmentation accuracy at the region level, with SCATR-ITS achieving comparable, albeit slightly lower, performance. In less complex regions, such as CULS and SCION, SegmentAnyTree and TreeLearn also deliver competitive results. However, ForestFormer3D and SCATR-ITS outperform them by a substantial margin in more challenging scenes consisting of dense and multi-layered forests.

BlueCat is the only region where SCATR-ITS surpasses the performance of ForestFormer3D, achieving an F$_1$-score of 69.1\%, 4 pp higher than ForestFormer3D. Moreover, TreeLearn achieves the highest relative recall in BlueCat, the only region where this occurs. These findings align with the results observed on the Wytham Woods dataset (see \Cref{section:generalization_lautx_wytham}), further suggesting that SCATR-ITS may be more effective than ForestFormer3D for segmenting TLS data captured in deciduous temperate forests. Conversely, in the TUWIEN region, ForestFormer3D achieves a significantly higher F$_1$ than SCATR-ITS, outperforming it by 9.2 pp, although SCATR-ITS remains second best. This indicates that the superpoint representations utilized by our model may lack sufficient precision for accurate segmentation in temperate deciduous forests at sparser point densities, where crown boundaries become more difficult to distinguish.

SegmentAnyTree produces the highest F$_1$-score in the Yuchen region by a narrow margin of 0.1 pp, marking the only instance where neither ForestFormer3D nor SCATR-ITS achieves the best segmentation accuracy across the entire FOR-InstanceV2 dataset. This likely stems from Yuchen's point density being substantially lower than the majority of data in FOR-InstanceV2, which may hinder the models' ability to learn accurate ITS for the sparser point clouds. SegmentAnyTree mitigates this issue through the explicit use of point cloud sparsification as a data augmentation strategy, which may explain its comparatively better performance in the region.

\begin{table*}[!t]
    \centering
    \caption{Comparison of region-level performance for SCATR-ITS and baseline methods on the FOR-InstanceV2 test set. Best performance for each region is in \textbf{bold}, and second-best is \underline{underlined} (excluding zero-value metrics).} \bigskip
    \scriptsize{
    \begin{tabular}{ll*{9}{S[table-format=2.1]}}
        \toprule
        \multirow{2}{*}{\textbf{Region}} & \multirow{2}{*}{\textbf{Model}} & \multicolumn{5}{c}{\textbf{Overall metrics (\%)}} & \multicolumn{4}{c}{\textbf{Crown category-level metrics (\%)}} \\ \cmidrule(lr){3-7} \cmidrule(lr){8-11}
        && \textbf{Precision} & \textbf{Recall} & \textbf{F$_1$} & \textbf{Cov} & \textbf{AP$_{\mathbf{50}}$} & \textbf{Recall$_\text{A}$} & \textbf{Recall$_\text{B}$} & \textbf{Recall$_\text{C}$} & \textbf{Recall$_\text{D}$} \\ \midrule \midrule
        \multirow{5}{*}{CULS} & YOLOv12 \citep{tian2025yolov12} & \underline{95.2} & \bfseries\tablenum{100.0} & \underline{97.6} & 90.4 & \bfseries\tablenum{100.0} & \bfseries\tablenum{100.0} & \bfseries\tablenum{100.0} & \text{--} & \text{--} \\
        & SegmentAnyTree \citep{wielgosz2024segmentanytree} & 80.0 & \bfseries\tablenum{100.0} & 88.9 & 98.2 & \underline{98.3} & \bfseries\tablenum{100.0} & \bfseries\tablenum{100.0} & \text{--} & \text{--} \\
        & TreeLearn \citep{henrich2024treelearn} & \bfseries\tablenum{100.0} & \bfseries\tablenum{100.0} & \bfseries\tablenum{100.0} & \textbf{99.5} & \bfseries\tablenum{100.0} & \bfseries\tablenum{100.0} & \bfseries\tablenum{100.0} & \text{--} & \text{--} \\
        & ForestFormer3D \citep{xiang2025forestformer3d} & \bfseries\tablenum{100.0} & \bfseries\tablenum{100.0} & \bfseries\tablenum{100.0} & \underline{99.3} & \bfseries\tablenum{100.0} & \bfseries\tablenum{100.0} & \bfseries\tablenum{100.0} & \text{--} & \text{--} \\
        & \textbf{SCATR-ITS (Ours)} & \bfseries\tablenum{100.0} & \bfseries\tablenum{100.0} & \bfseries\tablenum{100.0} & \underline{99.3} & \bfseries\tablenum{100.0} & \bfseries\tablenum{100.0} & \bfseries\tablenum{100.0} & \text{--} & \text{--} \\ \midrule
        \multirow{5}{*}{NIBIO} & YOLOv12 \citep{tian2025yolov12} & 77.3 & 63.4 & 69.6 & 57.4 & 60.7 & 89.1 & 92.1 & 39.7 & 7.1 \\
        & SegmentAnyTree \citep{wielgosz2024segmentanytree} & 86.8 & 90.1 & 88.4 & 81.7 & 83.1 & \bfseries\tablenum{100.0} & 94.7 & \underline{90.5} & 42.9 \\
        & TreeLearn \citep{henrich2024treelearn} & 91.4 & 91.9 & 91.6 & 85.0 & 77.0 & \underline{95.7} & \bfseries\tablenum{100.0} & \underline{90.5} & \textbf{64.3} \\
        & ForestFormer3D \citep{xiang2025forestformer3d} & \underline{96.8} & \textbf{95.0} & \textbf{95.9} & \textbf{88.7} & \textbf{95.0} & \bfseries\tablenum{100.0} & \bfseries\tablenum{100.0} & \textbf{95.2} & \textbf{64.3} \\
        & \textbf{SCATR-ITS (Ours)} & \textbf{97.4} & \underline{93.2} & \underline{95.2} & \underline{85.4} & \underline{92.2} & \bfseries\tablenum{100.0} & \underline{97.4} & \textbf{95.2} & \underline{50.0} \\ \midrule
        \multirow{5}{*}{RMIT} & YOLOv12 \citep{tian2025yolov12} & 64.7 & 17.2 & 27.2 & 17.0 & 13.0 & 40.0 & 26.3 & 0.0 & 0.0 \\
        & SegmentAnyTree \citep{wielgosz2024segmentanytree} & 76.6 & 76.6 & 76.6 & 68.9 & 57.8 & \bfseries\tablenum{100.0} & 78.9 & \underline{76.2} & 33.3 \\
        & TreeLearn \citep{henrich2024treelearn} & \underline{81.6} & 62.5 & 70.8 & 62.0 & 54.0 & \underline{93.3} & 73.7 & 42.9 & 33.3 \\
        & ForestFormer3D \citep{xiang2025forestformer3d} & \textbf{85.5} & \textbf{82.8} & \textbf{84.1} & \textbf{74.6} & \textbf{77.5} & 86.7 & \bfseries\tablenum{100.0} & \textbf{81.0} & \underline{44.4} \\
        & \textbf{SCATR-ITS (Ours)} & 80.6 & \textbf{78.1} & \underline{79.4} & \underline{70.4} & \underline{73.5} & 80.0 & \underline{94.7} & 71.4 & \textbf{55.6} \\ \midrule
        \multirow{5}{*}{SCION} & YOLOv12 \citep{tian2025yolov12} & 74.5 & 88.4 & 80.9 & 74.6 & 84.7 & 92.3 & \bfseries\tablenum{100.0} & 50.0 & 0.0 \\
        & SegmentAnyTree \citep{wielgosz2024segmentanytree} & \underline{83.3} & \underline{93.0} & \underline{87.9} & 87.1 & 74.2 & \underline{96.2} & \bfseries\tablenum{100.0} & \underline{75.0} & 0.0 \\
        & TreeLearn \citep{henrich2024treelearn} & \textbf{97.7} & \textbf{97.7} & \textbf{97.7} & \textbf{92.2} & \underline{94.7} & \bfseries\tablenum{100.0} & \bfseries\tablenum{100.0} & \underline{75.0} & \bfseries\tablenum{100.0} \\
        & ForestFormer3D \citep{xiang2025forestformer3d} & \textbf{97.7} & \textbf{97.7} & \textbf{97.7} & 90.6 & \textbf{97.7} & \bfseries\tablenum{100.0} & \bfseries\tablenum{100.0} & \bfseries\tablenum{100.0} & 0.0 \\
        & \textbf{SCATR-ITS (Ours)} & \textbf{97.7} & \textbf{97.7} & \textbf{97.7} & \underline{91.4} & \textbf{97.7} & \bfseries\tablenum{100.0} & \bfseries\tablenum{100.0} & \bfseries\tablenum{100.0} & 0.0 \\ \midrule
        \multirow{5}{*}{TUWIEN} & YOLOv12 \citep{tian2025yolov12} & 25.0 & 20.0 & 22.2 & 25.9 & 5.1 & 30.0 & 30.0 & 9.1 & 0.0 \\
        & SegmentAnyTree \citep{wielgosz2024segmentanytree} & 51.2 & 60.0 & 55.3 & 53.3 & 45.8 & \underline{80.0} & 70.0 & 45.5 & \textbf{25.0} \\
        & TreeLearn \citep{henrich2024treelearn} & 69.7 & \underline{65.7} & 67.6 & \underline{61.6} & 36.9 & \underline{80.0} & \underline{80.0} & \underline{54.5} & \textbf{25.0} \\
        & ForestFormer3D \citep{xiang2025forestformer3d} & \textbf{92.9} & \textbf{74.3} & \textbf{82.5} & \textbf{62.0} & \textbf{73.4} & \textbf{90.0} & \textbf{90.0} & \textbf{63.6} & \textbf{25.0} \\
        & \textbf{SCATR-ITS (Ours)} & \underline{88.0} & 62.9 & \underline{73.3} & 56.2 & \underline{62.3} & \textbf{90.0} & 70.0 & 45.5 & \textbf{25.0} \\ \midrule
        \multirow{5}{*}{NIBIO2} & YOLOv12 \citep{tian2025yolov12} & 87.5 & 45.0 & 59.5 & 39.4 & 43.8 & 85.0 & 81.7 & 16.1 & 0.7 \\
        & SegmentAnyTree \citep{wielgosz2024segmentanytree} & 84.7 & 79.8 & 82.2 & 74.0 & 72.7 & \underline{98.4} & \underline{91.6} & 70.5 & 58.0 \\
        & TreeLearn \citep{henrich2024treelearn} & 86.4 & 60.8 & 71.4 & 59.2 & 53.6 & 86.5 & 75.2 & 49.3 & 30.7 \\
        & ForestFormer3D \citep{xiang2025forestformer3d} & \textbf{95.5} & \textbf{86.5} & \textbf{90.8} & \textbf{79.3} & \textbf{86.2} & \textbf{99.5} & \textbf{95.1} & \textbf{80.8} & \textbf{69.3} \\
        & \textbf{SCATR-ITS (Ours)} & \underline{94.2} & \underline{82.9} & \underline{88.2} & \underline{75.3} & \underline{81.3} & \underline{98.4} & \underline{91.6} & \underline{76.4} & \underline{64.0} \\ \midrule
        \multirow{5}{*}{NIBIO\_MLS} & YOLOv12 \citep{tian2025yolov12} & 70.0 & 60.9 & 65.1 & 55.2 & 49.2 & \underline{90.9} & 50.0 & \underline{66.7} & 0.0 \\
        & SegmentAnyTree \citep{wielgosz2024segmentanytree} & \underline{80.8} & \underline{91.3} & 85.7 & \underline{83.5} & 72.9 & \underline{90.9} & \bfseries\tablenum{100.0} & \bfseries\tablenum{100.0} & \underline{80.0} \\
        & TreeLearn \citep{henrich2024treelearn} & 45.5 & 43.5 & 44.4 & 47.1 & 14.8 & 27.3 & \underline{75.0} & \underline{66.7} & 40.0 \\
        & ForestFormer3D \citep{xiang2025forestformer3d} & \bfseries\tablenum{100.0} & \bfseries\tablenum{100.0} & \bfseries\tablenum{100.0} & \textbf{94.4} & \bfseries\tablenum{100.0} & \bfseries\tablenum{100.0} & \bfseries\tablenum{100.0} & \bfseries\tablenum{100.0} & \bfseries\tablenum{100.0} \\
        & \textbf{SCATR-ITS (Ours)} & \bfseries\tablenum{100.0} & \underline{91.3} & \underline{95.5} & 83.2 & \underline{91.3} & \bfseries\tablenum{100.0} & \bfseries\tablenum{100.0} & \bfseries\tablenum{100.0} & 60.0 \\ \midrule
        \multirow{5}{*}{BlueCat} & YOLOv12 \citep{tian2025yolov12} & 38.0 & 5.0 & 8.9 & 8.3 & 3.2 & 36.8 & 7.1 & 0.7 & 0.0 \\
        & SegmentAnyTree \citep{wielgosz2024segmentanytree} & 63.4 & 35.2 & 45.3 & 40.2 & 25.4 & 81.6 & 45.2 & 35.1 & 17.8 \\
        & TreeLearn \citep{henrich2024treelearn} & 72.8 & \textbf{57.9} & 64.5 & \textbf{57.5} & 37.7 & \textbf{97.4} & \textbf{72.6} & \textbf{60.4} & \underline{36.0} \\
        & ForestFormer3D \citep{xiang2025forestformer3d} & \underline{84.5} & 52.9 & \underline{65.1} & 53.0 & \underline{49.6} & \underline{92.1} & 68.5 & 49.3 & 34.5 \\
        & \textbf{SCATR-ITS (Ours)} & \textbf{86.1} & \underline{57.7} & \textbf{69.1} & \underline{55.8} & \textbf{54.8} & 89.5 & \underline{71.4} & \underline{56.0} & \textbf{41.1} \\ \midrule
        \multirow{5}{*}{Yuchen} & YOLOv12 \citep{tian2025yolov12} & 54.5 & 25.0 & 34.3 & 28.6 & 21.4 & 35.3 & 0.0 & 0.0 & \text{--} \\
        & SegmentAnyTree \citep{wielgosz2024segmentanytree} & 70.4 & \textbf{79.2} & \textbf{74.5} & \textbf{75.9} & 53.3 & \textbf{94.1} & 0.0 & \textbf{60.0} & \text{--} \\
        & TreeLearn \citep{henrich2024treelearn} & 50.0 & 4.2 & 7.7 & 8.3 & 1.0 & 5.9 & 0.0 & 0.0 & \text{--} \\
        & ForestFormer3D \citep{xiang2025forestformer3d} & \underline{84.2} & \underline{66.7} & \underline{74.4} & \textbf{67.4} & \textbf{57.4} & \underline{76.5} & 0.0 & \textbf{60.0} & \text{--} \\
        & \textbf{SCATR-ITS (Ours)} & \textbf{88.2} & 62.5 & 73.2 & 67.0 & \underline{56.4} & \underline{76.5} & 0.0 & \underline{40.0} & \text{--} \\ \bottomrule
    \end{tabular}}
    \label{table:performance_comparison_fiv2_regions}
\end{table*}

\section{On the discrepancies in reported ITS baseline performance compared to prior studies} \label{appendix:performance_discrepancies}

When evaluating the baseline methods on public individual tree segmentation datasets, we observed that some accuracy metrics differ from those reported in previous studies, despite using the exact same model code. In this appendix, we analyze potential causes for these performance discrepancies across the four datasets used in this study: FGI-EMIT, FOR-InstanceV2, LAUTx, and Wytham Woods.

\subsection{FGI-EMIT}

Slight differences occur between our segmentation accuracy for YOLOv12 and ForestFormer3D and the corresponding benchmark values reported by \citet{ruoppa2026benchmarking}. However, both discrepancies have straightforward explanations. First, YOLOv12 yields a higher overall accuracy than that reported by \citet{ruoppa2026benchmarking}, with the F$_1$-score increasing from 52.0\% to 55.5\%. This stems from improvements implemented to the DTM construction phase during preprocessing, which now produces more accurate height normalization in forest plots with steep elevation changes and consequently improves overall model performance by a small margin.

Second, \citet{ruoppa2026benchmarking} performed a second iteration of inference with ForestFormer3D on one of the denser plots in FGI-EMIT to maximize accuracy. By contrast, we use only a single inference iteration across all plots to ensure the reported inference runtimes remain strictly comparable. Consequently, marginal differences exist in the accuracy metrics of ForestFormer3D between our study and the benchmark of \citet{ruoppa2026benchmarking} (for example, a 0.2 pp difference in Recall).

Finally, the reported average plot-level runtimes differ from those presented in \citet{ruoppa2026benchmarking}. For all models except ForestFormer3D, this stems from including all preprocessing steps in our reported runtimes. For ForestFormer3D, where our reported average runtime is nearly $2.5\times$ faster, the speedup results from using an optimized internal reimplementation developed at FGI rather than the official source code. This reimplementation yields identical results to the original model, with speedups arising purely from data-processing optimizations. Because SCATR-ITS is also built upon this optimized version, we use it for benchmarking to ensure a fair runtime comparison.

\subsection{FOR-InstanceV2}

Our results for TreeLearn and ForestFormer3D on FOR-InstanceV2 differ from those reported by \citet{xiang2025forestformer3d}. Most notably, a systematic discrepancy occurs in the Coverage metric, where previously published values have consistently been higher. For example, on the FOR-InstanceV2 test set, we observe a Coverage of 72.3\% with ForestFormer3D using the publicly available pretrained model weights. By contrast, \citet{xiang2025forestformer3d} reported a Coverage of 81.2\%, 8.9 pp higher than ours, with the other metrics remaining comparable. This difference likely results from the employed Coverage formulation: an alternative definition weights each $\text{maxIoU}(\cdot)$ by the number of points in the corresponding ground truth tree \citep{wielgosz2026segmentanytreev2}. However, since point-count-weighting can disproportionately emphasize dense data modalities like TLS, we adopt the unweighted Coverage instead.

Discrepancies also occur in the detection metrics of both models, specifically Precision, Recall, and F$_1$-score. For ForestFormer3D, the difference is marginal ($+1.3$ pp in F$_1$-score) and likely stems from differing evaluation protocols and possibly model stochasticity. For TreeLearn, however, we observe a more substantial performance difference: our F$_1$-score on FOR-InstanceV2 is 71.4\%, whereas \citet{xiang2025forestformer3d} reported an F$_1$ of only 50.6\% on the same dataset. This occurs despite employing the exact TreeLearn training setup described by \citet{xiang2025forestformer3d}, with the only difference being our inclusion of the Yuchen region in the training and test sets. Because such a minor dataset difference is unlikely to cause this performance gap, we conjecture that \citet{xiang2025forestformer3d} may have opted not to employ geometric verticality, which is an optional input feature in TreeLearn. However, our preliminary experiments showed that TreeLearn typically performs significantly better when verticality is included. Consequently, since verticality can be directly computed from raw $xyz$-coordinates at the cost of additional preprocessing time, we include it across all TreeLearn experiments.

\subsection{LAUTx and Wytham Woods}

For LAUTx and Wytham Woods, the performance discrepancies between this study and previous work are often significantly larger than those observed for the other two datasets. For example, on Wytham Woods, we achieve an F$_1$-score of 59.9\% using ForestFormer3D, whereas \citet{xiang2025forestformer3d} reported 75.0\% with the same model and dataset, a difference of 15.1 pp. This variance is largely attributable to the less rigorous benchmarking practices for datasets typically reserved for assessing ITS model generalization. As a result, many of the previously reported results on LAUTx and Wytham Woods have been achieved under highly different experimental setups and are not strictly comparable.

Aside from differences in metric calculation, which are a potential source for minor discrepancies in model accuracy, the observed performance gaps are most likely primarily driven by two data-related factors. First, previous works did not necessarily train the baseline ITS methods on FOR-InstanceV2. For example, \citet{henrich2024treelearn} trained TreeLearn on a dataset composed of MLS data captured in temperate forests. Consequently, better performance on the relatively similar Wytham Woods, compared to a model trained on FOR-InstanceV2, is to be expected. Moreover, \citet{xiang2025forestformer3d} compiled metrics reported in prior literature, whereas we retrain all baselines from scratch on FOR-InstanceV2 prior to evaluating on LAUTx and Wytham Woods. This ensures that all models share equivalent prior knowledge, rendering the reported results strictly comparable. Although prior work demonstrates that certain baselines can achieve higher accuracy on LAUTx and Wytham Woods when trained on datasets other than FOR-InstanceV2, this holds true for any model. Ultimately, strictly comparable and reproducible results achieved under an identical experimental setup provide a more robust foundation for evaluating model performance.

Second, dataset versioning presents another data-related issue affecting both LAUTx and Wytham Woods. For both datasets, an original \citep{tockner2022automatic,tockner2022lautx,calders2022laser,calders2022terrestrial} and an extended \citep{henrich2024extended} version exist. The original datasets contain only the individual tree segments, whereas the extended versions include the surrounding environment, such as ground and understory vegetation. Consequently, the extended versions are inherently more challenging to segment. Qualitative results presented by \citet{xiang2025forestformer3d} (Figure 11) lack ground and understory points, strongly suggesting ForestFormer3D was evaluated on the original datasets, which likely explains the previously reported higher accuracies. Similarly, dataset source links in \citet{wielgosz2024segmentanytree} point to the use of non-extended versions for evaluating SegmentAnyTree.

\end{appendices}

\end{document}